\documentclass{article}

\usepackage{microtype}
\usepackage{graphicx}
\usepackage{subfigure}
\usepackage{booktabs} 
\usepackage{algorithm}
\usepackage{algorithmic}
\usepackage{hyperref}

\usepackage{thm-restate}

\newcommand{\alglinelabel}{%
  \addtocounter{ALC@line}{-1}
  \refstepcounter{ALC@line}
  \label
}

\usepackage[accepted]{icml2025}

\usepackage{amsmath}
\usepackage{amssymb}
\usepackage{mathtools}
\usepackage{amsthm}

\usepackage{placeins}

\usepackage[capitalize,noabbrev]{cleveref}

\theoremstyle{plain}
\newtheorem{theorem}{Theorem}[section]

\theoremstyle{definition}
\newtheorem{definition}[theorem]{Definition}

\theoremstyle{remark}

\usepackage{multicol}

\usepackage[textsize=tiny]{todonotes}

\usepackage{amsmath,amsfonts,bm}
\usepackage{algorithm,algorithmic}

\def\1{\bm{1}}

\def\vmu{{\bm{\mu}}}
\def\vtheta{{\bm{\theta}}}
\def\vphi{{\bm{\varphi}}}
\def\vpi{{\bm{\pi}}}

\def\va{{\bm{a}}}

\def\vo{{\bm{o}}}

\def\vr{{\bm{r}}}
\def\vs{{\bm{s}}}

\def\vw{{\bm{w}}}
\def\vx{{\bm{x}}}

\def\vz{{\bm{z}}}
\def\vtheta{{\bm{\theta}}}

\DeclareMathAlphabet{\mathsfit}{\encodingdefault}{\sfdefault}{m}{sl}
\SetMathAlphabet{\mathsfit}{bold}{\encodingdefault}{\sfdefault}{bx}{n}
\usepackage{amssymb}

\newcommand{\innerproduct}[2]{\left< #1, #2 \right>}

\usepackage{xcolor}
\definecolor{mydarkblue}{rgb}{0,0.08,0.45}
\definecolor{mygreen}{rgb}{0.032, 0.6392, 0.2039}
\definecolor{mypurple}{HTML}{B266FF}

\def\E{{\mathbb E}}

\def\X{{\mathcal X}}
\def\S{{\mathcal S}}
\def\A{{\mathcal A}}
\def\Z{{\mathcal Z}}

\def\R{{\mathbb R}}

\def\vpi{{\boldsymbol \pi}}
\def\vrho{{\boldsymbol \rho}}

\def\X{{\mathcal X}}
\def\B{{\mathcal B}}
\def\O{{\mathcal O}}
\def\K{{\mathcal K}}

\usepackage[font=small]{caption}
\usepackage{sidecap}
\usepackage{booktabs}       
\usepackage{amsfonts}       
\usepackage{nicefrac}       
\usepackage{microtype}      
\usepackage{xcolor}         

\usepackage{afterpage}

\usepackage[inline, shortlabels]{enumitem}  

\newcommand{\column}[2]{%
  \begin{tabular}[b]{@{}c@{}}#1\\#2\end{tabular}%
}

\definecolor{DarkGreen}{rgb}{0.1,0.5,0.1}
\definecolor{DarkRed}{rgb}{0.5,0.1,0.1}
\definecolor{DarkBlue}{rgb}{0.1,0.1,0.5}
\definecolor{DarkYellow}{rgb}{.79,.79,0}
\usepackage{hyperref}       
\hypersetup{
    unicode=false,          
    pdftoolbar=true,        
    pdfmenubar=true,        
    pdffitwindow=false,      
    pdfnewwindow=true,      
    colorlinks=true,       
    linkcolor=DarkBlue,          
    citecolor=DarkGreen,        
    filecolor=DarkRed,      
    urlcolor=DarkBlue,          
    pdftitle={},
    pdfauthor={},
}
\usepackage{url}

\newcommand\scalemath[2]{\scalebox{#1}{\mbox{\ensuremath{\displaystyle #2}}}}

\icmltitlerunning{Addressing Rotational Learning Dynamics in Multi-Agent Reinforcement Learning}

\begin{document}
\twocolumn[
\icmltitle{Addressing Rotational Learning Dynamics in \\Multi-Agent Reinforcement Learning}



\icmlsetsymbol{equal}{*}

\begin{icmlauthorlist}
\icmlauthor{Baraah A. M. Sidahmed}{bbb}
\icmlauthor{Tatjana Chavdarova}{ttt}
\end{icmlauthorlist}

\icmlaffiliation{bbb}{CISPA Helmholtz Center for Information Security \& Universität des Saarlandes.}
\icmlaffiliation{ttt}{Politecnico di Milano  \& University of California, Berkeley}

\icmlcorrespondingauthor{Baraah A. M. Sidahmed}{baraah.adil@cispa.de}
\icmlcorrespondingauthor{Tatjana Chavdarova}{tatjana.chavdarova@berkeley.edu}

\icmlkeywords{Machine Learning, ICML}
\vskip 0.3in
]



\printAffiliationsAndNotice{}  

\begin{abstract} 
Multi-agent reinforcement learning (MARL) has emerged as a powerful paradigm for solving complex problems through agents' cooperation and competition, finding widespread applications across domains. Despite its success, MARL faces a reproducibility crisis.
We show that, in part, this issue is related to the rotational optimization dynamics arising from competing agents' objectives, and require methods beyond standard optimization algorithms.
We reframe MARL approaches using Variational Inequalities (VIs), offering a unified framework to address such issues. Leveraging optimization techniques designed for VIs, we propose a general approach for integrating gradient-based VI methods capable of handling rotational dynamics into existing MARL algorithms. Empirical results demonstrate significant performance improvements across benchmarks.
In zero-sum games, \textit{Rock--paper--scissors} and \textit{Matching pennies}, VI methods achieve better convergence to equilibrium strategies, and in the \textit{Multi-Agent Particle Environment: Predator-prey}, they also enhance team coordination. These results underscore the transformative potential of advanced optimization techniques in MARL.

\end{abstract}

\section{Introduction}\label{sec:intro}

Multi-Agent Reinforcement Learning (MARL) extends Reinforcement Learning (RL) to environments where multiple agents interact, cooperate, or compete to achieve their goals. 
Indeed, in many real-world scenarios, decision-making is inherently multi-agent in nature, requiring adaptive strategies to navigate dynamic interactions. Additionally, competition among agents often drives more efficient and robust learning.
As a result, MARL has been applied to various domains such as autonomous driving, robotic coordination, financial markets, and multi-player games, demonstrating its potential to tackle complex problems~\citep[for instance,][]{Omidshafiei2017DeepDM,vinyals2017starcraft,drones_players,zhou2021game,bertsekas2021rollout}.

Unfortunately, deploying and advancing MARL research presents significant challenges.
The iterative training process in data-driven MARL is notoriously unstable, often struggling to converge. Performance is highly sensitive to minor factors, such as the choice of the initial random seed, which complicates benchmarking. 
While similar issues exist in single-agent actor-critic RL—an instance of a two-player game~\citep{wang2022pesky,eimer2023hyperparameters}—they are even more severe in MARL, leading to what is now termed the \emph{MARL reproducibility crisis}~\citep{bettini2024marl}.
For instance, \citet{gorsane2022standardised} report significant performance variability across different seeds in popular MARL benchmarks such as the \emph{StarCraft} multi-agent challenge~\citep{samvelyan2019starcraft}.
Additionally, gradient-based optimization in MARL struggles with exploring the joint policy space~\citep{10.5555/3618408.3619214, christianos2021shared}, often resulting in suboptimal solutions.  Some MARL structures also exhibit inherent cycling effects~\citep{zheng2021stackelberg}, further exacerbating the problem of convergence.
Notably, both actor-critic RL and MARL go beyond standard minimization, operating as two- or multi-player games. The introduction of competitive learning objectives and interaction terms leads to distinct learning dynamics, where standard gradient descent (GD) methods fail to converge even in simple cases~\citep{korpelevich1976extragradient}.

\let\thefootnote\relax\footnotetext{\url{https://github.com/badil96/VI-marl}.}

In mathematics and numerical optimization, several frameworks allow for modeling equilibrium-finding problems. These relate to, most prominently, the \emph{Variational Inequality}~\citep[VIs,][]{stampacchia1964formes,facchinei2003finite} framework; refer to Section~\ref{sec:prelim} for definition.
Informally, GD's failure on simple VI instances is due to the ``rotational component'' of the associated vector field in such problems~\citep{mescheder2018convergence,balduzzi2018mechanics}. For instance, the GD method for the $\min_{\vz_1\in \R^{d_1}} \max_{\vz_2\in \R^{d_2}} \  \  \vz_1 \cdot \vz_2$ game, rotates around the equilibrium $(0,0)$ for infinitesimally small learning rates, and diverges away from it for practical choices of its value.
As a result, GD---and all its adaptive step size variations, such as \emph{Adam}~\citep{kingma2014adam}---have no hope of converging for a large class of equilibrium-seeking problems. 

Inspired by the training difficulties in generative adversarial networks~\citep[GANs,][]{goodfellow2014generative}, significant recent progress has been made in understanding how to solve VIs. 
This includes both developing theoretical guarantees~\citep[e.g.,][]{golowich2020last,daskalakis2020complexity,gorbunov2022extra} and the development of efficient methods for large-scale optimization~\citep{diakonikolas2020halpern,chavdarova2021lamm}; refer to Appendix~\ref{app:related_works} for an extended discussion.

MARL problems are modeled with \emph{stochastic games}~\citep{littman1994}; refer to Section~\ref{sec:prelim}.
Three main MARL learning paradigms are commonly used:
\begin{itemize}[leftmargin=*,itemsep=0em,,topsep=0em]
    \item \emph{value-based learning}---focuses on estimating so-called \emph{value functions} (e.g., $Q$-learning, Deep \emph{Q-Networks}~\citep{mnih2015humanlevel}) to learn action-values first and infer a policy based on it,
    \item \emph{policy-based learning}---directly optimizes the \emph{policy} (e.g., \emph{REINFORCE~\citep{10.1007/BF00992696}}) by adjusting action probabilities without explicitly learning the value functions, and
    \item \emph{actor-critic methods}---combines value-based and policy-based approaches where an actor selects actions, and a critic evaluates them.
\end{itemize}
Furthermore, MARL can be broadly categorized into \emph{centralized} and \emph{independent} learning approaches. In centralized MARL, a global critic or shared value function leverages information from all agents to guide learning, improving coordination~\citep{Sunehag2017ValueDecompositionNF,lowe2017multi,Yu2021TheSE}. In contrast, independent MARL treats each agent as a separate learner, promoting scalability but introducing non-stationarity as agents continuously adapt to each other’s evolving policies~\citep{matignon2012independent,foerster2017}. In this work, we focus on \emph{Centralized Training Decentralized Execution} (CTDE) approaches, specifically the ones with centralized critics.
 Several of the new algorithms in MARL belong to this category such as \citep[MADDPG, ][]{lowe2017multi}, \citep[MATD3,][]{Ackermann2019ReducingOB}, \citep[MAPPO, ][]{Yu2021TheSE}, and \citep[COMA, ][]{foerster2018counterfactual}.

In summary, this paper explores the following:
\begin{center}
    \emph{Do rotational learning dynamics significantly contribute to training instabilities in centralized MARL? \\
    Can performance be improved using variational inequality (VI) methods?}
\end{center}
To address this question, we primarily focus on the CTDE actor-critic MARL learning paradigm, and build VI approaches leveraging a (combination of) \emph{nested-Lookahead-VI}~\citep{chavdarova2021lamm} and \emph{Extragradient}~\citep{korpelevich1976extragradient} methods for iteratively solving variational inequalities (VIs).
Our main contributions are as follows:
\begin{itemize}[leftmargin=*,itemsep=0em,,topsep=0em]
    \item We introduce a VI perspective for multi-agent reinforcement learning (MARL).
    \item We propose two general approaches for integrating variational inequality (VI) methods with actor-critic MARL. The primary focus of this work is \textit{LA-MARL} (Algorithm~\ref{alg:nested_lookahead_marl}), a computationally efficient approach particularly well-suited for large-scale MARL optimization where agents are represented by neural networks. While we present the algorithm in the actor-critic framework, the underlying approach is applicable beyond this setting, as we elaborate.
    \item We evaluate our proposed methods against standard optimization techniques in two zero-sum games---\emph{Rock--paper--scissors} and \emph{Matching pennies}---and in two benchmarks from the \emph{Multi-Agent Particle Environments}~\citep[MPE,][]{lowe2017multi}.
    \item We also illustrate the limitations of using the agents' \emph{rewards} as a performance metric in MARL and provide empirical insights and an alternative.
\end{itemize}

\section{Related Works}\label{sec:related}

In the following, we discuss related works that study the optimization in centralized MARL. Our approach primarily builds on two key areas, VIs, and MARL, which we review in Appendix~\ref{app:related_works}. The necessary VI/MARL background is presented in Section~\ref{sec:prelim}. Works focused on optimization in independent MARL are also discussed in Appendix~\ref{app:related_works}.

\noindent\textbf{Limit behavior.}
\citet{mazumdar2020} study the continuous-time limit behavior. Using dynamical systems and stochastic approximation tools, they show how gradient-based algorithms can lead to oscillatory behaviors and non-Nash attractors of gradient descent that have no meaning for the game.

\noindent\textbf{Convergence.}
Several works rely on two-player zero-sum Markov games to study the regret of an agent relative to a perfect adversary. For instance, \citet{bai2020} introduces self-play algorithms for online learning---the \emph{Value Iteration with Upper/Lower Confidence Bound} (VI-ULCB) and an explore-then-exploit algorithm---and show the respective regret bounds.  
In addition to the online setting, \citet{xie20markov} also consider the offline setting where they propose using Coarse Correlated Equilibria (CCE) instead of Nash Equilibrium (NE) and derive concentration bounds for CCEs.

For the classical \emph{linear quadratic regulator} (LQR) problem~\citep{kalmanLQR}, single-agent policy gradient methods are known to exhibit global convergence~\citep{fazelLQR}. 
The LQR problem extends to the multi-agent setting through \emph{general-sum linear quadratic (LQ) games}, where multiple agents jointly control a (high-dimensional) linear state process. 
Unlike the zero-sum case~\citep{bu2019globalconvpolicygradient,zhang2021derivativefree}, policy gradient faces significant challenges for general-sum LQ games with $n$ players.
\citet{mazumdar2019policygrad} highlight a negative result 
that contrasts sharply with the corresponding result in the single-agent setting.
In particular, policy gradient methods fail to guarantee even local convergence in the deterministic setting~\citep{mazumdar2019policygrad}, and additional techniques are required to guarantee convergence~\citep{hambly2023linQuadGames}.
Beyond LQ games, policy gradient struggles with multi-agent settings more broadly.
\citet{ma2021polymatrix} address gradient descent limitations in multi-agent scenarios, introducing an alternative gradient-based algorithm for finding equilibria in \emph{polymatrix games}~\citep{janovskaja1968polymatrix}, which model multi-agent interactions with pairwise competition.

Convergence in MARL is challenging due to complex interactions and non-stationarity among agents. While multi-agent actor-critic methods are widely used~\citep{bettini2024benchmarl}, their optimization and convergence properties remain underexplored, making this an open problem.

\section{Preliminaries}\label{sec:prelim}

\noindent\textbf{Notation.}
We denote 
\begin{enumerate*}[series = tobecont, itemjoin = \enspace, label=(\roman*),font=\itshape]
\item vectors with small bold letters,
\item sets with curly capital letters,
\item real-valued functions with small letters, and
\item operators $\Z\mapsto\Z$ with capital letters, e.g., $F\,.$ 
\end{enumerate*}
The notation $[n]$ denotes the set $\{1, \dots, n\}$.
In the following, let $\mathcal{Z}$ be a convex and compact set in the Euclidean space, 
equipped with the inner product $\innerproduct{\cdot}{\cdot}$. 
We adopt standard MARL notation to describe the setting, as we discuss next.

\paragraph{MARL problem formulation.}
\emph{Markov games}~\citep[MGs, also known as \emph{stochastic games},][]{shapley1953,littman1994} generalize \emph{Markov Decision Processes}~\citep[MDPs][]{mdp_book1994} to a multi-agent setting. 
An MG is defined by the tuple:
\begin{equation}\label{eq:mg_tuple} \tag{MG}
\big( n, \S, \{\A_i\}_{i=1}^n, p, \{ r_i\}_{i=1}^n, \gamma \big) \,,
\end{equation}
where $n$ agents interact within an environment characterized by a common state space $\S$. 
Each agent $i \in [n]$ receives observation $\vo_i \in \mathcal{O}$ of the current state $\vs \in \S$ of the environment. 
In the most general case, agent $i$'s observation $\vo_i = f(\vs)$, where $f\colon \S\to \mathcal{O}_i $.
For instance, $f$ can be an identity or coordinate-selection map with $\mathcal{O}_i\subseteq\S$, or $f$ can be a nontrivial mapping.
Based on its policy $\pi_i\colon \O_i \to \A_i$, each agent $i \in [n]$ selects an action $a_i \in \A_i$, where $\A_i$ is its finite action set.
The joint actions of all agents are represented as $\va \triangleq(a_1, \dots, a_n)$, and the joint action space as $\A \triangleq \A_1\times \dots \times \A_n\,.$

The environment transitions to a new state $\vs'\in\S$ according to a \emph{transition function} $p \colon \S \times \A \to \Delta(\S)\,,$ where $\Delta(\S)$ is the space of probability distributions over $\S$ (non-negative $|\S|$-dimensional vector summing to $1$). The function $p$ specifies the probability distribution of the next state $\vs'$, given the current state $\vs$ and the joint action $\va$.

Each agent $i \in [n]$ receives a reward $r_i$,  where the reward function $r_i\colon \S \times \A \times \S \to \R $ depends on the current state, the joint action, and the resulting next state. 
The importance of future rewards is governed by the \emph{discount factor} $\gamma\in[0,1)$.

\ref{eq:mg_tuple}s generalize both MDPs and \emph{repeated games}~\citep{aumann1995repeated} by introducing non-stationary dynamics, where agents learn their policies jointly and adaptively.
Each agent $i\in[n]$ aims to maximize its expected cumulative reward (return), defined as: 
\begin{equation*} \label{eq:general_marl}\tag{MA-Return}
v_i^{\pi_i, \vpi_{-i}} 
({\vs}) {=} \E \Big[
\sum\limits_{t=0}^{\infty} \gamma^t r_i(\vs_t,  \va_t, \vs_{t+1}) | \vs_0 {\sim} \vrho, \va_t {\sim} \vpi(s_t)
\Big]  
    \,,
\end{equation*}
where  $\vpi\triangleq(\pi_1, \dots , \pi_n)$ represents the joint policy of all agents, 
$\vpi_{-i}$ denotes the policies of all agents except agent $i$, 
 and $\vrho$ is the initial state distribution.
The interaction among agents introduces challenges such as non-stationarity (due to evolving policies) and complex reward interdependencies, leading to a distinct optimization landscape. Solutions often aim to find \emph{Nash equilibria}, where no agent can improve its return by unilaterally altering its policy.

\paragraph{MADDPG.}
\emph{Multi-agent deep deterministic policy gradient \citep[MADDPG, ][]{lowe2017multi}}, 
extends \emph{Deep deterministic policy gradient}~\citep[DDPG, ][]{Lillicrap2015ContinuousCW} to multi-agent setting, leveraging a centralized training with decentralized execution paradigm.
Each agent $i\in [n]$ has: \vspace{-.3em}
\begin{itemize}[leftmargin=*,noitemsep,topsep=0pt]
    \item \emph{critic} network $Q_i\colon\O_1\times \dots \times\O_n\times\A \to \R$, parametrized by $\vw_i \in\R^{d_i^Q}$: acts as a centralized action-value function, evaluating the expected return of joint actions $\va$ in state $\vs$ and
    \item \emph{actor} network $\mu_i\colon \O_i \to \Delta(\A_i)$, parametrized by $\vtheta_i \in \R^{d_i^\mu}$: represents the agent's policy, mapping agents' observation of states $s$ to a probability distribution over actions $a_i$.
\end{itemize}
For stability during training, MADDPG employs \emph{target} networks, which are delayed versions of the critic and actor networks:
\begin{itemize}[leftmargin=*,noitemsep,topsep=0pt]
    \item \emph{target critic}: $\bar{Q}_i$, parametrized by $\bar\vw_i \in\R^{d_i^Q} \,,$ and
    \item \emph{target-actor} $\bar\mu_i$, parametrized by $\bar\vtheta_i \,.$
\end{itemize}

The parameters of the target networks are updated using a soft update mechanism:
\begin{align}
\bar\vw_i \leftarrow \tau\vw_i + (1 - \tau)\bar\vw_i \,,\label{eq:target_critic}\tag{Target-Critic}\\
\bar\vtheta_i \leftarrow \tau\vtheta_i + (1 - \tau)\bar\vtheta_i \,,
\label{eq:target_actor}\tag{Target-Actor}
\end{align}
where $\tau \in (0,1]$ is a hyperparameter controlling the update rate of the target networks. 

\paragraph{MATD3.}
\emph{Multi-agent TD3 \citep{Ackermann2019ReducingOB}} improves upon MADDPG by introducing two key modifications: 
\begin{itemize}[leftmargin=*,noitemsep,topsep=0pt]
    \item Dual critics: each agent $i\in[n]$ has two critic networks, $Q_{i,1}$ and $Q_{i,2}$. When calculating the target for $Q$-learning, the smaller of the two values is used, mitigating overestimation bias. 
    \item Delayed actor updates: Actor policies and target networks are updated less frequently, typically after every $c$ critic updates, to improve training stability.
\end{itemize}
Additionally, random noise is added to the target actor's outputs during training to introduce exploration and avoid deterministic policies getting stuck in suboptimal regions. These enhancements make MATD3 more robust in multi-agent settings compared to MADDPG.

\paragraph{Variational Inequality}~\citep{stampacchia1964formes,facchinei2003finite}.
Variational Inequalities (VIs) extend beyond standard minimization problems to encompass a broad range of equilibrium-seeking problems. The connection to such general problems can be understood from the optimality condition for convex functions: a point $\vz^\star$ is an optimal solution if and only if $\langle \vz - \vz^\star, \nabla f(\vz^\star) \rangle \geq 0, \forall\vz \in \text{dom} f\,.$
In the framework of VIs, this condition is generalized by replacing the gradient field $\nabla f$ with a more general vector field $F$, allowing for the modeling of a wider class of problems.
Formally, the VI goal is to find an equilibrium  $\vz^\star$ from the domain of continuous strategies $\mathcal{Z}$, such that:
\begin{equation} \label{eq:vi} \tag{VI}
	\langle \vz-\vz^\star, F(\vz^\star) \rangle \geq 0, \quad \forall \vz \in \mathcal{Z} \,,
\end{equation}
where $F\colon \mathcal{Z}\to \R^d$, referred to as the \emph{operator}, is continuous, and $\mathcal{Z}$ is a subset of the Euclidean $d$-dimensional space $\R^d$.
VIs are thus characterized by the tuple $(F,\mathcal{Z})$, denoted herein as VI($F$, $\mathcal{Z}$).
For a more comprehensive introduction to VIs, including examples and applications, see Appendix~\ref{app:vi_intro}.

\begin{algorithm}[tbh]
    \begin{algorithmic}[1]
        \STATE {\bfseries procedure} {\scshape NestedLookAhead}{\bfseries:}
        \begin{ALC@g}
            \STATE {\bfseries Input:}
                $\#$agents $n$,
                episode counter  $e$,
                actor and critic weights and snapshots: $\{(\vtheta_i, \vtheta_{i}^{(1)}, \dots, \vtheta_{i}^{(l)})\}_{i=1}^n$ and $\{(\vw_i, \vw_{i}^{(1)}, \dots,  \vw_{i}^{(l)})\}_{i=1}^n$,
                LA hyperparameters: levels $l$, $(k^{(1)}$, \dots, $k^{(l)})$
                and $(\alpha_\vtheta, \alpha_\vw)$.
            \FORALL{$j \in [l]$}
            \IF{ $e \% k^{(j)} == 0$}
                \FORALL{agent $i \in [n]$}
                        \STATE $\vw_i \leftarrow \vw_{i}^{(j)} + \alpha_{\vw} (\vw_i - \vw_{i}^{(j)})$
                        \hfill\emph{\color{gray}LA $j^{\text{th}}$ level}\label{line:la-update-a1}
                        \STATE $\vtheta_i \leftarrow \vtheta_{i}^{(j)} + \alpha_{\vtheta} (\vtheta_i - \vtheta_{i}^{(j)})$ \label{line:la-update-c1}
                        \STATE $(\vtheta_{i}^{(1)}, \dots , \vtheta_{i}^{(j)}, \vw_{i}^{(1)}, \dots,  \vw_{i}^{(j)}) \leftarrow (\{\vtheta_i\}_{\times j}, \{\vw_i\}_{\times j} )$
                        \hfill \emph{\color{gray}Update copies up to $j^{\text{th}}$} 
                        \label{line:la-save-1}
                    \ENDFOR
            \ENDIF
            \ENDFOR
        \end{ALC@g}
        \STATE {\bfseries end procedure}
    \end{algorithmic}
    \caption{Pseudocode for LA-VI, called from Algorithm~\ref{alg:nested_lookahead_marl}. Updates the parameters in-place.}
   \label{alg:lookahead}
\end{algorithm}

\paragraph{VI methods.}
The \textit{gradient descent} method straightforwardly extends for the \ref{eq:vi} problem as follows:
\begin{equation} \tag{GD}\label{eq:gd}
    \vz_{t+1} = \vz_t - \eta F(\vz_t) \,,
\end{equation}
where $t$ denotes the iteration count, and $\eta \in (0,1)$ the step size or learning rate.

\textit{Extragradient}~\citep{korpelevich1976extragradient} is a modification of \ref{eq:gd}, which uses a ``prediction'' step to obtain an extrapolated point $\vz_{t+\frac{1}{2}}$ 
 using~\ref{eq:gd}: $\vz_{t+\frac{1}{2}} \!=\! 
 \vz_t - \eta F(\vz_t) 
 $, and the gradients at the \textit{extrapolated}  point are then applied to the \textit{current} iterate $\vz_t$ as follows:
\begin{equation} \tag{EG} \label{eq:extragradient}
\begin{aligned}     
\vz_{t+1}  \!=\! 
\vz_t - \eta F \Big(  
\vz_{t+\frac{1}{2}} 
\Big)  
\,.
\end{aligned}
\end{equation}
Unlike gradient descent, \ref{eq:extragradient} converges in some simple game instances, such as in games linear in both players~\citep{korpelevich1976extragradient}.

The \emph{nested-Lookahead-VI} (LA) algorithm for \ref{eq:vi} problems~\citep[Alg. 3,][]{chavdarova2021lamm}, is a general wrapper of a ``base'' optimizer $B\colon\R^n\to \R^n$ where, after every $k$ iterations with $B$, $\vz_{t+1} = B(\vz_t)$ an averaging step is performed as follows:
\begin{equation}
\tag{LA}
\vz_{t+k} \leftarrow \vz_t + \alpha (\vz_{t+k} -  \vz_t), \quad \alpha \in [0,1]  
\,. \label{eq:lookahead_mm}
\end{equation}
For this purpose a copy (snapshot) of the iterate after the averaging step is stored for the next LA update. See Appendix~\ref{app:vi_methods} for an alternative view.

This averaging can be applied recursively across multiple levels $l$, when using \ref{eq:lookahead_mm} as base optimizer, typically with $l\in [1,3]$. In Algorithm~\ref{alg:lookahead}, the parameter $k^{(j)}$ at level $j\in[l]$ is defined as the multiple of $k^{(j-1)}$ from the previous level $j-1$, specifically $k^{(j)}= c_j \cdot k^{(j-1)}$.
For $l=1, k=2$, LA has connections to \ref{eq:extragradient}~\citep{chavdarova2021hrdes}, however for higher values of $k$ and $l$ the resulting operator exhibits stronger contraction~\citep{chavdarova2021lamm,ha2022la}, which effectively addresses rotational learning dynamics.

\section{VI Perspective \& Optimization}
\label{sec:approach}

In this section, we introduce the operators for multi-agent general policy-based learning, actor-critic methods, and the specific operator corresponding to MADDPG. Following this, we present a broader class of algorithms that incorporate a designated MARL operator and integrate it with LA and/or EG.

\subsection{MARL Operators}\label{sec:vi_perspective_general}

\vspace{.3em}
\paragraph{General MARL.}
Policy-based learning directly solves the \eqref{eq:general_marl} problem, where agents optimize their policy parameters directly to maximize their return.
The operator $F_{\text{MAR}}$, where \emph{MAR} stands for \textit{multi-agent-return}, with $\mathcal{Z}\equiv \A$, corresponds to:
\begin{equation}\tag{$F_{\text{MAR}}$}\label{eq:F_ma_max-return}
\begin{aligned} 
& \qquad 
F_{\text{MAR}} \Big(  \begin{bmatrix} \vdots \\
\vpi_i\\
\vdots\\
\end{bmatrix}   \Big) {\triangleq}
\begin{bmatrix} \vdots \\
\nabla_{\vpi_i} v_i^{\pi_i, \vpi_{-i}} \\ 
 \vdots \\
\end{bmatrix} \\
& \equiv
\begin{bmatrix} \vdots \\
\nabla_{\vpi_i} \Big(
    \E \Big[ \sum\limits_{t=0}^{\infty}
\gamma^t r_i(\vs_t,  \va_t, \vs_{t+1}) | \vs_0 {\sim} \vrho, \va_t {\sim} \vpi(s_t)
\Big] 
\Big)\\ 
 \vdots \\
\end{bmatrix} \,.
\\
\end{aligned}
\end{equation}
\vspace{-.3em}

\paragraph{Actor-critic MARL.}
We denote by $\vx$ the full state information from which the agents observations $\vo_i$ are derived.
As above, consider a centralized critic network, denoted as the $Q$--network: $\mathbf{Q}^\vmu_i(\vx_t,\va_t ; \vw_i)$ and an associated policy network $\vmu_i(\vo_i;\vtheta_i)$ for each agent $i \in [n]\,.$ 
Given a batch of experiences $\B=\{(\vx^j, \va^j, \vr^j, \vx'^j )\}_{j=1}^{|\B|}\,,$ drawn from a replay buffer $\mathcal{D}$, the objective is to find an equilibrium by solving the \eqref{eq:vi} with the operator defined as:

\begin{equation}
\tag{$F_{\text{MAAC}}$}\label{eq:F_marl}
F_{\text{MAAC}} \Big(\begin{bmatrix} \vdots \\
\vw_i\\
\vtheta_i\\
\vdots\\
\end{bmatrix}   \Big) {\equiv}
\begin{bmatrix} \vdots \\
\nabla_{\vw_i} \Big( { \frac{1}{|\B|}\sum\limits_{j=1}^{|\B|}{ \ell_i^\vw (\cdot ;\vw_i)}} \Big)\\ 
\nabla_{\vtheta_i} \Big( { \frac{1}{|\B|}\sum\limits_{j=1}^{|\B|}{ \ell_i^\vtheta (\cdot;\vtheta_i)}} \Big)\\
 \vdots \\
\end{bmatrix} \,,
\end{equation}
\vspace{.3em} 
where the parameter space is $\mathcal{Z}\equiv \R^d$, with $d=\sum_{i=1}^n (d_i^{Q} + d_i^{\mu})$; and \emph{MAAC} stands for \textit{multi-agent-actor-critic}.
Even in the single-agent case ($n=1$), an inherent game-like interaction exists between the actor and the critic: the update of $\vw_i$ depends on $\vtheta_i$, and vice versa. This interplay is fundamental to the optimization dynamics in multi-agent actor-critic frameworks.

\paragraph{MADDPG.}
As an illustrative example, we fully present the terms in \eqref{eq:F_marl} for MADDPG,
deferring the other algorithms to Appendix~\ref{app:vi_presp}.
The critic loss function is defined as:
\begin{equation}\tag{$\ell^{\vw}_{\text{MADDPG}}$}\label{eq:cr_maddpg}
\begin{aligned}
\ell_i^\vw (\cdot ;\vw_i) & = \left( y_i
- \mathbf{Q}^\vmu_i(\vx^j,\va^j ; \vw_i )\right)^2,\\
y_i & = r_i^j + \gamma\mathbf{Q}_i^{\bar\vmu}\left.(\vx'^j,\va'; \vw_i')  \right|_{\va' = \bar\vmu(\vo'^{j)}} \,. \\
\end{aligned}
\end{equation} 
The actor loss function is given by:
\begin{equation} \tag{$\ell^{\vtheta}_{\text{MADDPG}}$}\label{eq:ac_maddpg}
\scalemath{0.9}{
    \ell_i^\vtheta (\cdot;\vtheta_i) {=} \vmu_i(\vo^j_i; \vtheta_i)\nabla_{a_i}\left. \mathbf{Q}^\vmu_i(\vx^j,a_1^j,\dots,a_i,\dots,a_n^j; \vw_i)\right|_{a_i=\vmu_i(\vo^j_i)}} \,.
\end{equation}
This formulation captures the interplay between the actor and critic networks in MADDPG, where the critic updates its parameters to minimize the Bellman error, while the actor updates its policy by maximizing the $Q$-value.

\subsection{Proposed Methods}

\begin{algorithm}[h]
\begin{algorithmic}[1]
   \STATE {\bfseries Input:}
        Environment $\mathcal{E}$, 
        operator $F$,
        number of agents $n$, 
        number of episodes $t$, 
        action spaces $\{\A_i\}_{i=1}^n$, 
        number of
        random steps $t_{\text{rand}}$, 
        learning interval $t_{\text{learn}}$, 
        actor networks $\{\vmu_i\}_{i=1}^n$ with $\vtheta \equiv \{\vtheta_i\}_{i=1}^n$, 
        critic networks $\{{Q}_i\}_{i=1}^n$ with $\vw \equiv \{\vw_i\}_{i=1}^n$, 
        target actor networks $\{\bar\vmu_i\}_{i=1}^n$ with $\bar\vtheta \equiv \{\bar\vtheta_i\}_{i=1}^n$, 
        target critic networks $\{\bar{Q}_i\}_{i=1}^n$ with $\bar\vw \equiv \{\bar\vw_i\}_{i=1}^n$, 
        learning rates $\eta_\vtheta, \eta_\vw$, 
        base optimizer $B$, 
        discount factor $\gamma$, 
        lookahead hyper-parameters $\mathcal{L}\equiv (l, \{k^{(j)}\}_{j=1}^l, \alpha_\vtheta, \alpha_\vw )$,
        soft update parameter $\tau$.
        
   \STATE {\bfseries Initialize:}
    \begin{ALC@g}
   \STATE Replay buffer $\mathcal{D} \leftarrow \varnothing$
   \STATE Initialize LA parameters: $\vphi \leftarrow \{ \vtheta \}_{\times l}, \{\vw \}_{\times l} $  
    \end{ALC@g}
   \FORALL{episode $e = 1$ to $t$}
       \STATE Sample initial state $\vx$ from $\mathcal{E}$ (with  $\vo \equiv f(\vx)$)
       \STATE $\text{step} \leftarrow 1$
       
       \REPEAT
           \IF{$\text{step} \leq t_{\text{rand}}$}
               \STATE Randomly select actions for each agent $i$
           \ELSE
               \STATE Select actions using policy for each agent $i$
           \ENDIF
           
           \STATE Execute $\va$, observe rewards and state $(\mathbf{r}, {\vx'})$
           \STATE Store $(\vx, \va, \mathbf{r}, {\vx'})$ in replay buffer $\mathcal{D}$
           \STATE $\vx \leftarrow {\vx'}$
           
           \IF{$\text{step} \% t_{\text{learn}} == 0$}
               \STATE Sample a batch $\mathcal{B}$ from $\mathcal{D}$
               \STATE Use  $\mathcal{B}$ and update to solve VI($F$, $\R^d$)  using $B$
                \STATE $\bar\vtheta \leftarrow \tau \vtheta + (1 - \tau)\bar\vtheta$ \hfill \emph{\color{gray}Update target actor}
                \STATE $\bar\vw \leftarrow \tau \vw + (1 - \tau)\bar\vw$ \hfill \emph{\color{gray}Update target critic}             
           \ENDIF
           
           \STATE $\text{step} \leftarrow \text{step} + 1$
       \UNTIL{environment terminates}
       
       \STATE {\scshape NestedLookAhead}($n, e,
        \vphi,
        \mathcal{L}
        $)
   \ENDFOR
   
   \STATE {\bfseries Output:} $\vtheta^{(l-1)}$, $\vw^{(l-1)}$
\end{algorithmic}
   \caption{LA--MARL Pseudocode. 
   }
   \label{alg:nested_lookahead_marl}
\end{algorithm}

To solve the~\ref{eq:vi} problem with an operator corresponding to the MARL algorithm---for instance~\eqref{eq:F_ma_max-return} or~\eqref{eq:F_marl}---we propose the \emph{LA-MARL}, and \emph{EG-MARL} methods, described in detail in this section.

\paragraph{LA-MARL, Algorithm~\ref{alg:nested_lookahead_marl}.} 
LA-MARL periodically saves snapshots of all agents' networks (both actor and critic) and averages them with the current networks during training. It operates with a base optimizer (e.g., Adam~\citep{kingma2014adam}, which is gradient descent with per-parameter adaptive step sizes) and applies a lookahead averaging step every $k$ intervals. Specifically, the current network parameters ($\vtheta, \vw$)  are updated using their saved snapshots ($\vtheta^{(j)}, \vw^{(j)}$) through 
$\alpha$-averaging (Algorithm~\ref{alg:lookahead}).

The algorithm allows for multiple nested lookahead levels, where higher levels update their corresponding parameters less frequently. All agents apply lookahead updates simultaneously at each step, ensuring consistency across both actor and critic parameters. Extended versions of LA-MARL tailored for MADDPG and MATD3, with more detailed notations, can be found in the appendix (Algorithms~\ref{alg:nested_lookahead_maddpg_extended} and \ref{alg:nested_lookahead_matd3_extended}).

\textit{Generalization and Adaptability.}
While Algorithm~\ref{alg:nested_lookahead_marl} focuses on off-policy actor-critic methods, it serves as a general framework and can be adapted to other MARL learning paradigms:
\begin{itemize}[leftmargin=*,itemsep=0em,,topsep=0em]
\item Policy Gradient Methods: It can be instantiated with a specific operator---of the form Eq.~\eqref{eq:F_ma_max-return}---by setting one set of parameters ($\vw$ or $\vtheta$) to $\empty$; or
\item On-Policy Learning: modifications include removing the use of target networks.
\end{itemize}
However, the lookahead method (Eq.~\ref{eq:lookahead_mm}) must be applied in the joint strategy space of all players. This is crucial because, in multi-agent reinforcement learning (MARL), the adversarial nature of agents' objectives introduces a rotational component in the associated vector field. The averaging steps help mitigate this effect. Particularly, to ensure correct updates, no agent should use parameters that have already been averaged within the same iteration.

\paragraph{(LA-)EG-MARL.}
For \textbf{EG-MARL}, the \eqref{eq:extragradient} update rule is used for both the actor and critic networks and for all agents; refer to  Algorithm~\ref{alg:extragradient} for full desrtiption.
Moreover, Algorithm~\ref{alg:nested_lookahead_marl} can also  use \ref{eq:extragradient} as the base optimizer---represented by $B$---herein denoted as \textbf{LA-EG-MARL}.

\paragraph{Remark on the convergence.} 
 
Assuming the above VI is \emph{monotone} (see Appendix~\ref{app:vi_methods} for definition)---that is, that agents’ reward functions satisfy certain structure---the above methods have convergence guarantees, that is EG-MARL~\citep{korpelevich1976extragradient,gorbunov2022extra}, LA-MARL~\citep{pethick2023stable}, and LA-EG-MARL~\citep[][Thm. 3]{chavdarova2021lamm}  have convergence guarantees. 
Contrary to these, the standard gradient descent method does not converge for this problem~\citep {korpelevich1976extragradient}.

\section{Experiments}\label{sec:experiments}

\subsection{Setup}

We use the open-source \textit{PyTorch} implementation of MADDPG~\citep{lowe2017multi}\footnote{Available at \url{https://github.com/Git-123-Hub/maddpg-pettingzoo-pytorch/tree/master}.} and extend it to MATD3 using the same hyperparameters specified in the original papers; detailed in Appendix \ref{app:implementation}.
Our experiments cover two zero-sum games—Rock--paper--scissors and Matching Pennies—along with two Multi-Agent Particle Environments (MPE) \citep{lowe2017multi}. We use game implementations from \emph{PettingZoo}~\citep{terry2021pettingzoo} and train for $60$k episodes, with $5$--$10$ random seeds.

\begin{figure*}[htp]
  \centering
  \subfigure[Rock--paper--scissors]{\includegraphics[width=.47\linewidth]{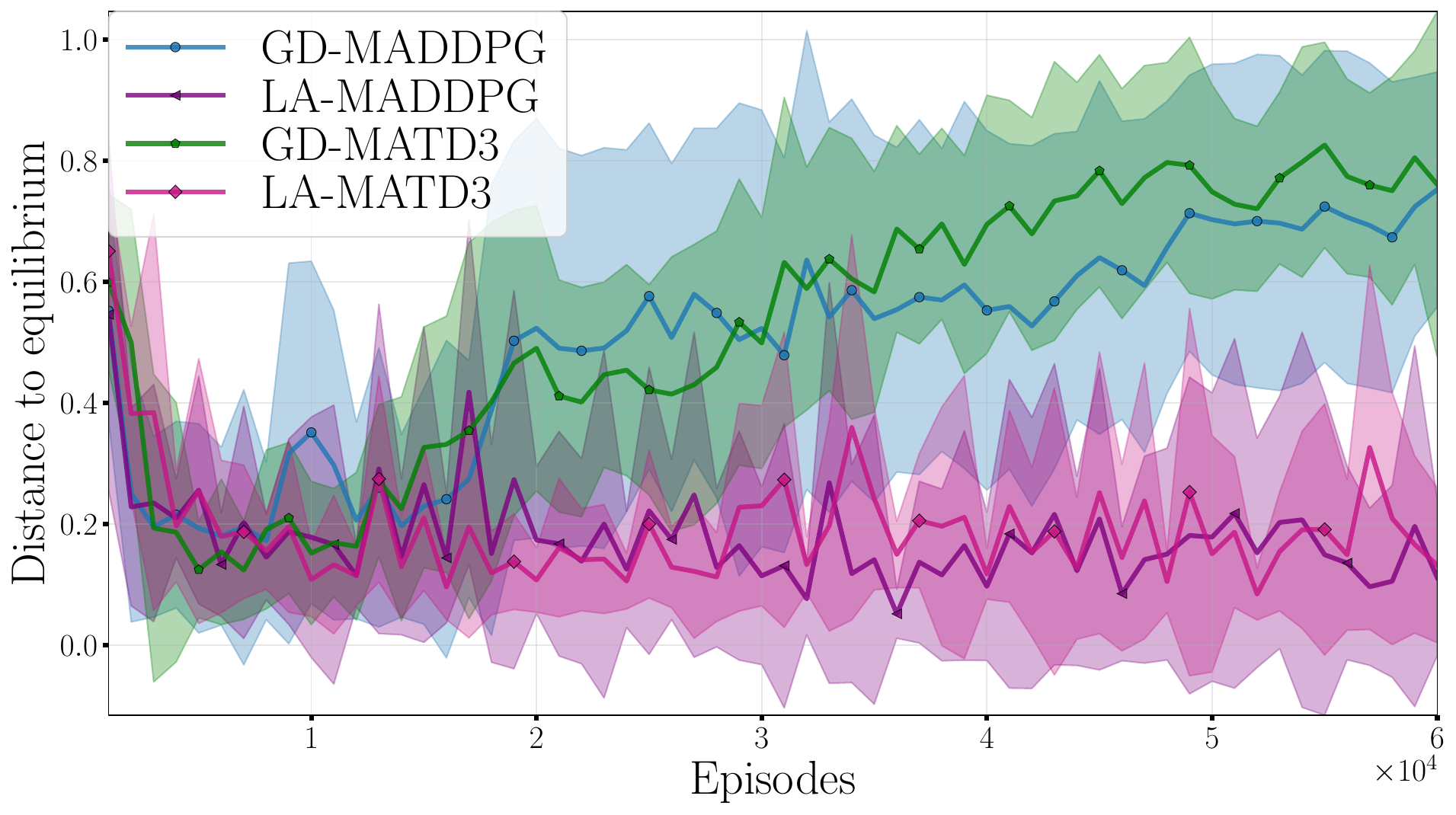}\label{fig:adamvslavseg_rps}}
  \subfigure[Matching pennies]{\includegraphics[width=.47\linewidth]{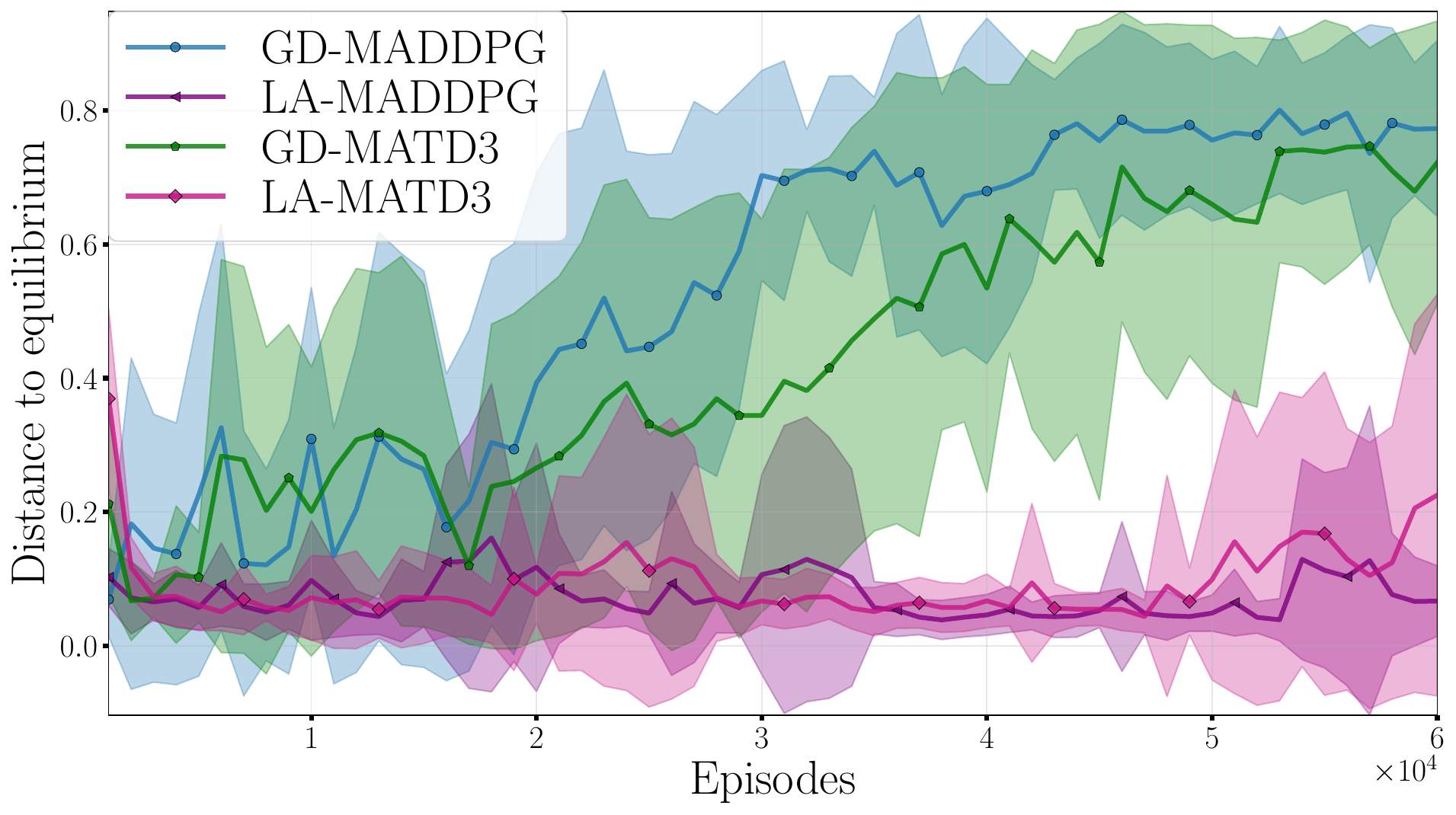}\label{fig:adamvslavseg_mp}}
  \caption{  \textbf{Comparison between \emph{GD-(MADDPG/MATD3)} and \emph{LA-(MADDPG/MATD3)}, on Rock--paper--scissors and Matching pennies.}  
$x$-axis: training episodes. 
$y$-axis: total distance of agents' policies to the equilibrium policy; averaged over $10$ seeds.
}
\end{figure*}

\noindent\textbf{Rock--paper--scissors (RPS) and Matching pennies (MP).}  
Widely studied games in multi-agent settings due to their analytically computable mixed Nash equilibria (MNE), which allows for a precise performance measure, and insights into their inherent cyclical behavior~\citep {Zhou_2015,Wang_2014,srinivasan2018}.
In RPS, two players ($n=2$) choose among three actions ($m=3$) per step, with a MNE of $\vpi^\star_{\text{RPS}}(\frac{1}{3}, \frac{1}{3}, \frac{1}{3})$. Players observe their opponent’s previous action before selecting their own, earning $+1$ for a win, $0$ for a tie, and $-1$ for a loss, over $t=25$ steps per episode.
Similarly, MP is a two-player, two-action ($m=2$) game where one player (\emph{Even}) wins if both actions match, while the other (\emph{Odd}) wins if they differ. The game has a MNE, of $\vpi^\star_{\text{MP}} =(\frac{1}{2}, \frac{1}{2})$, and runs also for $t=25$ steps.
We compute the squared norm of the learned policy probabilities relative to the equilibrium for both games.

\noindent\textbf{MPE: Predator-Prey and Physical Deception.}
We evaluate two environments from the Multi-Agent Particle Environments (MPE) benchmark \citep{lowe2017multi}. \emph{Predator-Prey}, has $n$ \emph{good} agents, $m$ \emph{adversaries}, and $l$ landmarks, where good agents are faster but penalized for being caught or going out of bounds, while adversaries collaborate to capture them. We use $n=1$, $m=2$, and $l=2$. 
In \emph{Physical deception}, we have $n$ good agents, one adversary, and $n$ landmarks, with one landmark designated as the \emph{target}. Good agents aim to get close to the target landmark while misleading the adversary, which must infer the target’s location. 
Unlike Predator-Prey, this environment does not involve direct competition for the adversary--its reward depends solely on its own policy.
In our experiments, we set $n=2$. 

\noindent\textbf{Methods.}
We evaluate our methods against the baseline, which is the MARL algorithms (MADDPG or MATD3) with Adam~\citep{kingma2014adam} as the optimizer $B$. Throughout the rest of this section, we will refer to baseline as GD-MARL (GD).
When referring to LA-based methods, we will indicate the $k$ values for each lookahead level in brackets. For instance, LA ($10,1000$) denotes a two-level lookahead where $k^{(1)} = 10$ and $k^{(2)} = 1000$. We denote with \textit{EG} the \ref{eq:extragradient} method, and refer to it analogously. 
Further details on hyperparameters are provided in Appendix~\ref{app:implementation}.

\subsection{Results}\label{sec:results}

\begin{table}[t]
\centering
    \begin{tabular}{@{}cccc@{}}
        \textbf{GD vs. GD} & \textbf{GD vs. LA}  & \textbf{LA vs. LA} & \textbf{LA vs. GD}  \\
        \hline
        $2.99 \pm 1.73$  & $2.14 \pm .91$ $\downarrow$  & $5.44 \pm 1.27$  & $7.41 \pm 1.75$ $\uparrow$             \\
    \end{tabular}
  \caption{  \textbf{Competition between agents trained with different algorithms.}
  Means and standard deviations (over $5$ seeds) of adversary reward in MPE: Physical deception, on $100$ test environments.
  When GD's opponent is switched to LA, its reward decreases, and vice versa. See Section \ref{sec:results}.
  }
  \vspace{-1em}
  \label{table:pp}
\end{table}

\noindent\textbf{$2$-player games: RPS \& MP.} Figures \ref{fig:adamvslavseg_rps} and \ref{fig:adamvslavseg_mp} illustrate the average distance between learned and equilibrium policies. GD-MARL eventually \emph{diverge}, whereas LA-MARL consistently converges to a near-optimal policy, outperforming the baseline. Both MARL algorithms perform similarly, though MATD3 exhibits lower variance across seeds than MADDPG.

For LA-based methods, experiments with different $k$ values indicate that smaller $k$-values for the innermost LA-averaging yield better performance (see Appendix \ref{app:hyperparam-selection}). The results consistently show performance improvements.

\begin{table}
\centering
    \begin{tabular}{cc}
        \textbf{Method} & \textbf{Adversary Win Rate} \\
        \hline
        Baseline        & $0.45 \pm .16$                \\
        LA-MADDPG        & $0.53\pm .11$              \\
        EG-MADDPG        & $0.56 \pm .27$              \\
        LA-EG-MADDPG       & $\mathbf{0.51} \pm \mathbf{.14}$              \\
    \end{tabular}
  \caption{ \textit{\textbf{Equilibrium reached?}}
  Average adversary win rate for MPE: Physical deception on $100$ test environments.
  The \textit{win rate} is the fraction of times the adversary was closer to the target at the end of episode. 
  \textit{Closer to $0.5$ is better.} Refer to Section~\ref{sec:results}.
  }
  \vspace{-1em}
  \label{table:phydec}
\end{table}

\begin{figure}[!thb]
    \centering
    \includegraphics[width=\columnwidth]{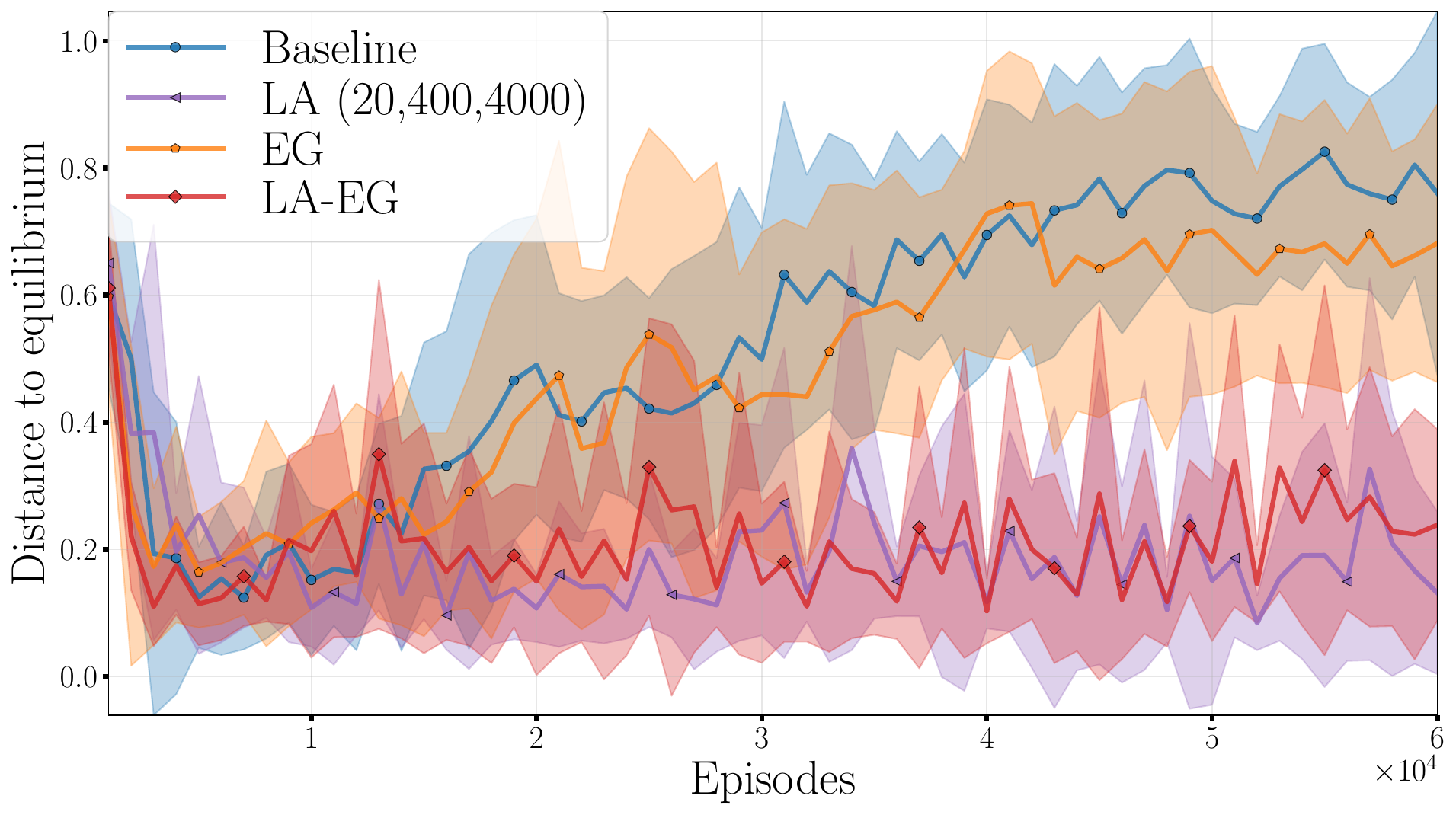}
    \caption{\textbf{Comparing the \emph{GD}, \emph{ LA}, \emph{ EG}, and \emph{LA-EG} optimization methods on the Rock--paper--scissors game.} 
$x$-axis: training episodes. 
$y$-axis: squared norm of the learned
policy probabilities relative to the equilibrium.
}
    \label{fig:rps_vi}
\end{figure}

\begin{figure*}[!htb]
  \centering
  \begin{subfigure}[Rewards]{\centering\includegraphics[width=.56\linewidth]{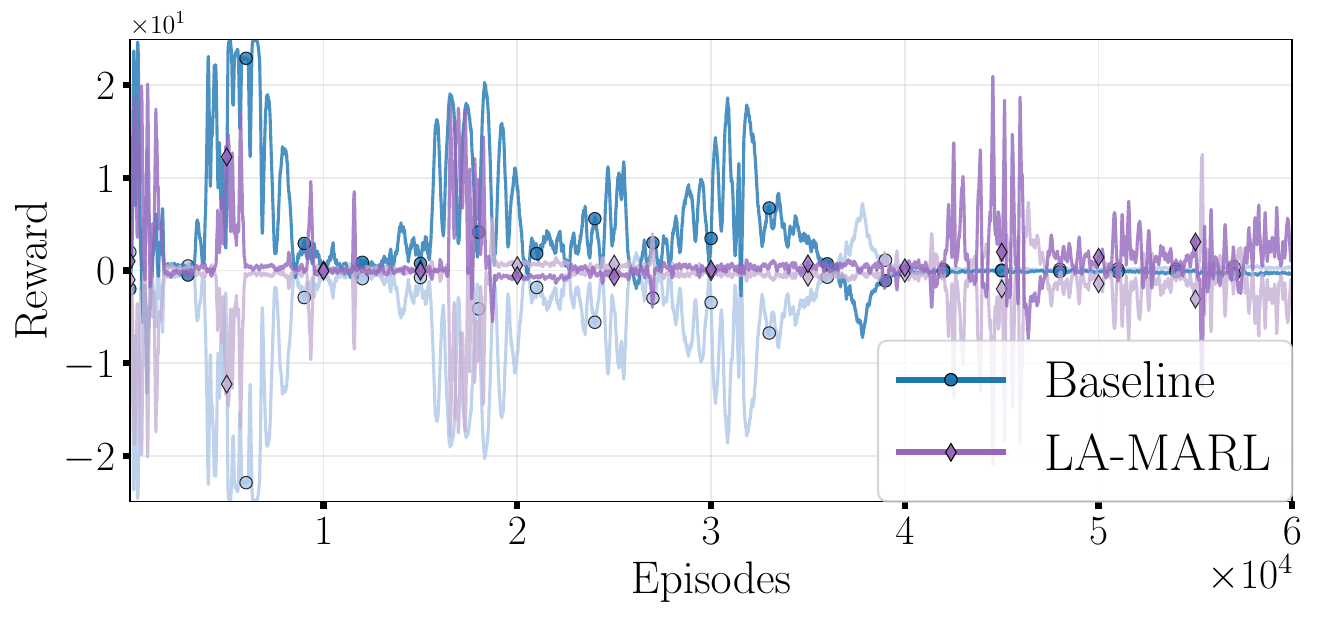}\label{fig:rewards}}
  \end{subfigure}
  \begin{subfigure}[Sampled actions]{
  \centering\includegraphics[width=.37 \linewidth]{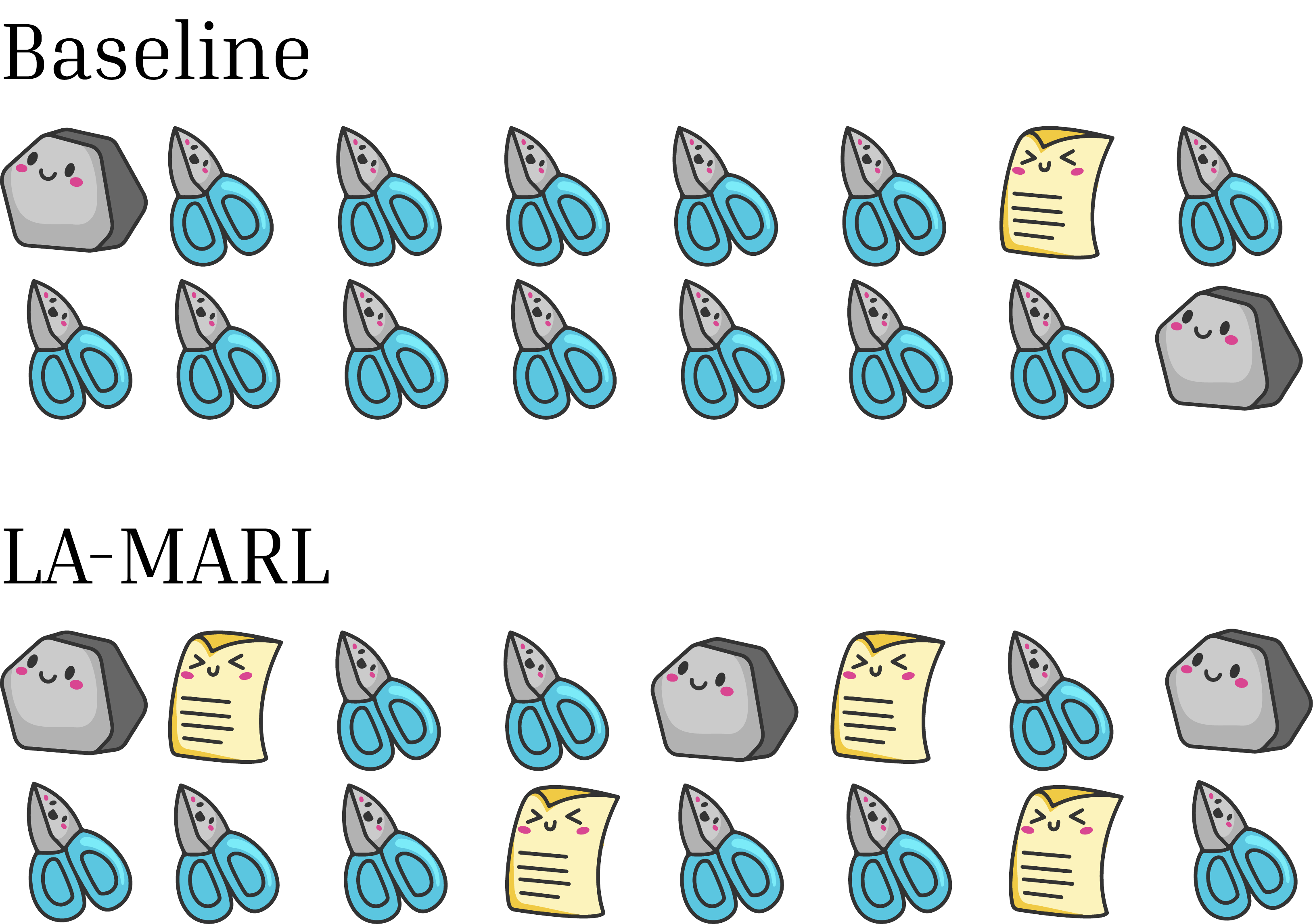}\label{fig:actions}
  }
  \end{subfigure}
 \caption{ 
 \textbf{Rewards (left) vs. sampled actions from learned policies (right), of (LA-)MADDPG in the Rock--paper--scissors game.}
 The baseline has saturating rewards (in the last part), however, that is not indicative of the agents' performances.
 Refer to Section~\ref{sec:results} for a discussion, and Figure~\ref{fig:snapshots_rps_more_details} for more detailed plots and larger action samples.
 }\label{fig:snapshots_rps}
\end{figure*}

\noindent\textbf{MPE: Predator-prey.} 
Table \ref{table:pp} presents the average rewards of the adversaries competing against the good agents. We train $5$--different seeds for GD-MATD3 (baseline) and LA-MATD3, then evaluate them against each other. The results show that LA-trained agents consistently outperform GD-trained agents, both in individual performance and in direct competition.

\noindent\textbf{MPE: Physical deception.}
Table \ref{table:phydec} presents the mean and standard deviation of the adversary's win rate, measuring how often it successfully reaches the target. In this setting, equilibrium is achieved when both teams win with equal probability across multiple instances. To ensure robust evaluation, we tested each method across $100$ environments per seed.
Given the \emph{cooperative} nature of the game, the baseline performs relatively well, with EG-MADDPG achieving similar performance. However, both LA-MADDPG and LA-EG-MADDPG outperform their respective base optimizers (MADDPG and EG-MADDPG), demonstrating improved stability and effectiveness.

\noindent\textbf{Summary.} Overall, our results indicate the following:
\begin{enumerate*}[series = tobecont, label=(\roman*),font=\itshape]
\item VI-based methods consistently outperform their respective baselines, by effectively handling the rotational dynamics.
\item LA-VI outperforms the other methods.
\end{enumerate*}

\noindent\textbf{Comparison among VI methods \& insights from GANs.}
The widely used \eqref{eq:extragradient} for solving VIs is known to converge for monotone VIs. However, in our experiments, EG performs only slightly better than the baseline because it introduces only a minor local adjustment compared to gradient descent (GD). This aligns with expectations: while EG occasionally outperforms GD, its improvements are often marginal.
In contrast, Lookahead, introduces a significantly stronger contraction, improving both stability and convergence. As the number of nested levels increases, performance gains become more pronounced—particularly in preventing the last iterate from diverging. However, too many nested levels can lead to overly conservative or slow updates. Based on our experiments, three levels of nested LA yielded the best balance between stability and convergence speed (see Fig. \ref{fig:rps_vi} for a comparison of VI methods).
Our findings are consistent with those observed in GAN training \citep{chavdarova2021lamm}, where EG also provides only slight improvements over the baseline, while more contractive methods consistently yield better performance.

These results further confirm that \emph{MARL vector fields in these environments exhibit strong rotational dynamics}. For scenarios with highly competitive reward structures, we recommend using VI methods with greater contractiveness, such as employing multiple levels of LA.

\noindent\textbf{Limitations of Rewards as a Metric in MARL.}
While saturating rewards are commonly used in MARL, few works challenge their reliability~\citep[e.g.,][]{NIPS2004_88fee042}. Our results suggest that reward ``convergence'' \emph{does not necessarily indicate optimal policies}.
In multi-agent settings, \emph{rewards can stabilize at a target value even with suboptimal strategies}, leading to misleading evaluations. For example, in Figure~\ref{fig:snapshots_rps}, baseline agents repeatedly select a subset of actions, resulting in ties that yield a saturating reward but fail to reach equilibrium, making them vulnerable to a more skilled opponent.
Conversely, LA-MADDPG agents did not exhibit reward saturation, yet they learned near-optimal policies by randomizing their actions, which aligns with the expected equilibrium.
This highlights the need for stronger evaluation metrics in MARL, especially when the true equilibrium is unknown. For further discussion, see Appendix~\ref{app:convergence}.

\FloatBarrier
\section{Discussion}\label{sec:conclusion}

Poor reproducibility remains a major challenge in MARL, particularly when scaling to large neural networks used in real-world applications. Instead of introducing a new MARL learning objective, this work revisits the fundamental optimization process--a crucial but often overlooked aspect of MARL implementations.
MARL involves inherent competition, leading to complex learning dynamics where standard gradient-based methods, designed for minimization problems, fall short. 

To address this, we adopt a variational inequality (VI) perspective, which provides a unifying framework for MARL learning approaches. Leveraging insights from VI optimization, we introduce the general Algorithm \ref{alg:nested_lookahead_marl}, a computationally efficient method tailored for practical MARL applications.
Our findings are consistent and compelling: simply replacing the optimization method---while keeping everything else unchanged---yields significant performance gains. Moreover, since Algorithm \ref{alg:nested_lookahead_marl} relies only on averaging, without additional gradient computations, the computational overhead is negligible.

These results offer strong empirical evidence that optimization methods play a crucial role in MARL performance, underscoring the need for further research in this direction.

\FloatBarrier

\bibliography{main}
\bibliographystyle{icml2025}

\clearpage

\appendix
\onecolumn
\title{Appendix}

\section{Extended Related Works Discussion}\label{app:related_works} 

Our work is primarily grounded in two key areas: Multi-Agent Reinforcement Learning (MARL) and Variational Inequalities (VIs), which we discuss next. Additionally, we extend our discussion on related works on Linear-Quadratic (LQ) games and discuss relevant literature on independent MARL.

\paragraph{Multi-Agent Reinforcement Learning (MARL).}
Various MARL algorithms have been developed~\citep{lowe2017multi,Iqbal2018,Ackermann2019ReducingOB,Yu2021TheSE}, with some extending existing single-agent reinforcement learning (RL) methods \citep{rashid2018qmix,son2019qtran,yu2022surprising,kuba2022trust}.
\citet{lowe2017multi} extend an actor-critic algorithm to the MARL setting using the \emph{centralized training decentralized execution} framework. 
In the proposed algorithm, named \textit{multi-agent deep deterministic policy gradient (MADDPG)}, each agent in the game consists of two components: an \emph{actor} and a \textit{critic}. 
The actor is a policy network that has access only to the local observations of the corresponding agent and is trained to output appropriate actions. 
The critic is a value network that receives additional information about the policies of other agents and learns to output the Q-value; see Section~\ref{sec:prelim}. After a phase of experience collection, a batch is sampled from a replay buffer and used for training the agents.
To our knowledge, all deep MARL implementations rely on either stochastic gradient descent or \emph{Adam} optimizer \citep{kingma2014adam} to train all networks.  Game theory and MARL share many foundational concepts, and several studies explore the relationships between the two fields \citep{yang2021overview,article}, with some using game-theoretic approaches to model MARL problems \citep{zheng2021stackelberg}. This work proposes incorporating game-theoretic techniques into the optimization process of existing MARL methods to determine if these techniques can enhance MARL optimization.

\citet{li2019robust} introduced an algorithm called \emph{M3DDPG}, aimed at enhancing the robustness of learned policies. Specifically, it focuses on making policies resilient to worst-case adversarial perturbations, as well as uncertainties in the environment or the behaviors of other agents.

\paragraph{Variational Inequalities (VIs).}
VIs were first formulated to understand the equilibrium of a dynamical system~\citep{stampacchia1964formes}. Since then, they have been studied extensively in mathematics, including operational research and network games~\citep[see][and references therein]{facchinei2003finite}. More recently, after the shown training difficulties of GANs~\citep{goodfellow2014generative}---which are an instance of VIs---an extensive line of works in machine learning studies the convergence of iterative gradient-based methods to solve VIs numerically.
Since the last and average iterates can be far apart when solving VIs~\citep[see e.g.,][]{chavdarova2019}, these works primarily aimed at obtaining last-iterate convergence for special cases of VIs that are important in applications, including bilinear or strongly monotone games~\citep[e.g., ][]{tseng1995,malitsky2015,facchinei2003finite,daskalakis2018training,liang2018interaction,gidel19-momentum,azizian20tight,thekumparampil2022lifted},  VIs with cocoercive operators~\citep{diakonikolas2020halpern}, or monotone operators~\citep{chavdarova2021hrdes,gorbunov2022extra}.
Several works 
\textit{(i)} exploit continuous-time analyses 
\citep{ryu2019ode,bot2020,rosca2021,chavdarova2021hrdes,bot2022fast}, 
\textit{(ii)} establish lower bounds for some VI classes~\citep[e.g.,][]{golowich2020last,golowich2020noregret}, and \textit{(iii)} study the constrained setting~\citep{daskalakis2019constrained,cai2022constrVI,acvi,chavdarova2024acvi}, among other.
Due to the computational complexities involved in training neural networks, iterative methods that rely solely on first-order derivative computation are the most commonly used approaches for solving variational inequalities (VIs). However, standard gradient descent and its momentum-based variants often fail to converge even on simple instances of VIs. As a result, several alternative methods have been developed to address this issue. Some of the most popular first-order methods for solving VIs include the \emph{extragradient} method~\citep{korpelevich1976extragradient}, \emph{optimistic gradient} method~\citep{popov1980}, \emph{Halpern} method~\citep{diakonikolas2020halpern}, and (nested) \emph{Lookahead-VI} method~\citep{chavdarova2021lamm}; these are discussed in detail in Section \ref{sec:prelim} and Appendix \ref{app:vi_methods}.
In this work, we primarily focus on the nested Lookahead-VI (LA) method, which has achieved state-of-the-art results on the CIFAR-10~\citep{cifar10} benchmark for generative adversarial networks~\citep{goodfellow2014generative}.

\paragraph{General-sum linear quadratic (LQ) games.}
In LQ games, each agent's action linearly impacts the state process, and their goal is to minimize a quadratic cost function dependent on the state and control actions of both themselves and their opponents. LQ games are widely studied as they admit global Nash equilibria (NE), which can be analytically computed using coupled algebraic Riccati equations~\citep{lancaster1995algebraic}.

Several works establish global convergence for policy gradient methods in zero-sum settings.
\citet{zhang2019policyLQgames} propose an alternating policy update with projection for deterministic infinite-horizon settings, proving sublinear convergence.
\citet{bu2019globalconvpolicygradient} study leader-follower policy gradient in a deterministic setup, and showing sublinear convergence. 
\citet{zhang2021derivativefree} study the sample complexity of policy gradient with alternating policy updates.

For the deterministic $n$-agent setting, \citet{mazumdar2019policygrad} showed that policy gradient methods fail to guarantee even local convergence.
\citet{roudneshin2020} prove global convergence for policy gradient in a \emph{mean-field} LQ game with infinite horizon and stochastic dynamics.
\citet{hambly2023linQuadGames} show that the \textit{natural policy gradient} method achieves global convergence in finite-horizon general-sum LQ games, provided that a certain condition on an added noise to the system is satisfied.
Recently, \citet{guan2024linQuadGames} proposed a policy iteration method for the infinite horizon setting.

\paragraph{Independent MARL.} 
In independent MARL, each agent learns its policy independently, without direct access to the observations, actions, or rewards of other agents~\citep{matignon2012independent,foerster2017}. Each agent treats the environment as stationary and ignores the presence of other agents, effectively treating them as part of the environment.

\citep{daskalakis2020indepMARL} study two-agent zero-sum MARL setting of independent learning algorithms. The authors show that if both players run policy gradient methods jointly, their policies will converge to a min-max equilibrium of the game, as long as their learning rates follow a two-timescale rule.
\citep{arslan2015} propose a decentralized $Q$-learning algorithm for MARL setting where agents have limited information and access solely of their local observations and rewards. 
\citet{jiang22i2q} proposes a decentralized algorithm.
\citet{sayin2021decentralized} explore a decentralized $Q$-learning algorithm for zero-sum Markov games, where two competing agents learn optimal policies without direct coordination or knowledge of each other's strategies. Each agent relies solely on local observations and rewards, updating their $Q$-values independently while interacting in a stochastic environment. 
\citep{lu2021} study decentralized cooperative multi-agent setting with coupled safety constraints.

\citet{wei2017onlineRL} rely on the framework of average-reward stochastic games to model single player with a perfect adversary, yielding a two-player zero-sum game, in a Markov environment, and study the regret bound.

\section{Additional Background} 
In this section, we further discuss VIs, and provide additional background and relevant algorithms.

\subsection{VI Discussion}\label{app:vi_intro}

\paragraph{Variational Inequality.}
We first recall the definition of VIs.
A $\textbf{VI}(F, \Z)$ is defined as:
\begin{equation} \label{eq:vi-restated} \tag{VI}
	\text{find } \vz^\star\in\Z \quad \text{ s.t. }\qquad \langle \vz-\vz^\star, F(\vz^\star) \rangle \geq 0, \quad \forall \vz \in \mathcal{Z} \,,
\end{equation}
where $F\colon \mathcal{Z}\to \R^d$, referred to as the \emph{operator}, is continuous, and $\mathcal{Z}$ is a subset of the Euclidean $d$-dimensional space $\R^d$.

When $F\equiv \nabla f$ and $f$ is a real-valued function $f\colon \Z \to \R$, the problem \ref{eq:vi-restated} is equivalent to standard minimization.
However, by allowing $F$ to be a more general vector field, VIs also model problems such as finding equilibria in zero-sum and general-sum games~\citep{cottle_dantzig1968complementary,rockafellar1970monotone}.
We refer the reader to ~\citep{facchinei2003finite} for an introduction and examples.

To illustrate the relevance of VIs to multi-agent problems, consider the following example.
Suppose we have $n$ agents, each with a strategy $\vz_i\in \R^{d_i}$, and let us denote the joint strategy with 
$$
\vz \equiv 
\begin{bmatrix}
   \vz_1 \\
    \vdots \\
    \vz_n 
\end{bmatrix}
\in \R^d, \qquad \text{with} \quad d = \sum_{i=1}^n d_i  \,.
$$
Each agent $i \in [n]$ aims to optimize its objective $f_i \colon \R^d \to \R$, which, in the general case, depends on all players' strategies.
Then, finding an equilibrium in this game is equivalent to solving a~\ref{eq:vi} where the operator $F$ corresponds to:
\begin{equation} \tag{$F_{n\text{-agents}}$} \label{eq:f_n_agents}
F_{n\text{-agents}}(\vz)\equiv
\begin{bmatrix}
\nabla_{\vz_1} f_1(\vz) \\
\vdots\\
\nabla_{\vz_n} f_n(\vz) \\
\end{bmatrix}   \,.
\end{equation}

An instructive way to understand the difference between non-rotational and rotational learning dynamics is to consider the second-derivative matrix $J\colon R^d \to \R^{d\times d}$ 
referred herein as the \emph{Jacobian}. For the above \eqref{eq:f_n_agents} problem the Jacobian is as follows:
\begin{equation}\label{eq:jacobian_def}\tag{$J_{n\text{-agents}}$}
J_{n\text{-agents}}(\vz)\equiv
\begin{bmatrix}
\nabla^2_{\vz_1^2} f_1(\vz)   &  \nabla^2_{\vz_1\vz_2} f_1(\vz) &\dots & \nabla^2_{\vz_1\vz_n} f_1(\vz)  \\
\vdots & \vdots & \dots & \vdots \\
\nabla^2_{\vz_n \vz_1} f_n(\vz)   &  \nabla^2_{\vz_n \vz_2} f_n(\vz) &\dots & \nabla^2_{\vz_n^2} f_n(\vz)  \\
\end{bmatrix}   \,.
\end{equation}
Notably, unlike in minimization, where the second-derivative matrix---the so-called \emph{Hessian}---is always symmetric, the Jacobian is not necessarily symmetric. Hence its eigenvalues may belong to the complex plane. 
In some cases, the Jacobian of the associated vector field can be decomposed into a symmetric and antisymmetric component~\citep{balduzzi2018mechanics}, where each behaves as a \textit{potential}~\citep{monderer1996potential} and a \textit{Hamiltonian} (purely rotational) game, resp.

In Section~\ref{app:vi_presp} we will also rely on a more general problem, referred to as the \textit{Quasi Variational Inequality}.

\paragraph{Quasi Variational Inequality.}
Given a map $F\colon\X\to\R^n$---herein referred as an \emph{operator}---the goal is to:
\begin{equation} \label{eq:qvi} \tag{QVI}
  \text{\emph{find }}\ \vx^\star \qquad\text{s.t.}\qquad \langle \vx -\vx^\star, F(\vx^\star)\rangle \geq 0, \quad \forall \vx \in \K(\vx^\star) \,,
\end{equation}
where $\K(\vx) \subseteq \R^d$ is a point-to-set mapping from $\R^d$ into subsets of $\R^d$ such that for every $\vx\in \X$, $\K(\vx)\subseteq \R^d$ which can be possibly empty.

In other words, the constraint set for \ref{eq:qvi}s depends on the variable $\vx$. 
This contrasts with a standard variational inequality \eqref{eq:vi}, where the constraint set $\K$ is fixed and does not depend on $\vx$. 
\ref{eq:qvi}s were introduced in a series of works by \citet{Bensoussan1973qvi1,Bensoussan1973qvi2,Bensoussan1974}.

\subsubsection{VI classes and additional methods}\label{app:vi_methods}

The following VI class is often referred to as the generalized class for VIs to that of convexity in minimization.

\begin{definition}[monotonicity]\label{def:monotone}
An operator $F: \R^d \to \R^d $ is \emph{monotone} if
$
    \langle \vz-\vz', F(\vz)-F(\vz') \rangle \geq 0, \enspace \forall \vz, \vz' \in \R^d \,.
$
$F$ is $\mu$-strongly monotone if:
$\langle \vz-\vz', F(\vz)-F(\vz') \rangle \geq \mu\|\vz-\vz'\|^2$ for all $\vz, \vz' \in \R^d$.
\end{definition}

The following provides an alternative but equivalent formulation of \ref{eq:lookahead_mm}. \ref{eq:lookahead_mm} was originally proposed for minimization problems~\citep{Zhang2019}.

\noindent\textbf{LA equivalent formulation.}
We can equivalently write \eqref{eq:lookahead_mm} as follows.
At a step $t$:
\begin{enumerate*}[series = tobecont, label=(\roman*),font=\itshape]
\item a copy of the current iterate $\tilde \vz_t$ is made: $ \tilde \vz_t \leftarrow  \vz_t$, 
\item $ \tilde \vz_t$ is updated $k \geq 1$ times using $B$, yielding $\tilde \vz_{t+k}$, and finally
\item the actual update $\vz_{t+1}$ is obtained as a \textit{point that lies on a line between} 
the current $\vz_{t}$ iterate and the predicted one $\tilde \vz_{t+k}$: 
\end{enumerate*}
\begin{equation*}
\tag{LA}
\vz_{t+1} \leftarrow \vz_t + \alpha (\tilde  \vz_{t+k} -  \vz_t), \quad \alpha \in [0,1]  
\,. 
\end{equation*}

In addition to those presented in the main part, we describe the following popular VI method.

\noindent\textbf{
Optimistic Gradient Descent (OGD).}
The update rule of Optimistic Gradient Descent 
OGD~\citep[(OGD)][]{popov1980} is:

\begin{equation} \label{eq:ogda} \tag{OGD}
    \vz_{t+1} = 
    \vz_{t} - 2\eta F(\vz_t) + \eta F(\vz_{t-1}) 
    \,,
\end{equation}
where $\eta \in(0,1)$ is the learning rate.

\subsubsection{Pseudocode for Extragradient}\label{app:eg-opt}

In Algorithm~\ref{alg:extragradient} outlines the \textit{Extragradient} optimizer~\citep{korpelevich1976extragradient}, which we employ in EG-MARL. This method uses a gradient-based optimizer to compute the extrapolation iterate, then applies the gradient at the extrapolated point to perform an actual update step. 
The extragradient optimizer is used to update all agents' actor and critic networks. In our experiments, we use Adam for both the extrapolation and update steps, maintaining the same learning intervals and parameters as in the baseline algorithm.

\begin{algorithm}
\caption{Extragradient optimizer; Can be used as $B$ in algorithm~\ref{alg:nested_lookahead_marl}.}
\begin{algorithmic}[1]

\STATE {\bfseries Input:} 
              learning rate $\eta_\psi$, initial weights $\psi$, 
              loss $\ell^{\psi}$, extrapolation steps $t$
\STATE $\psi^{\textit{copy}} \leftarrow \psi$ \hfill \emph{\color{gray}(Save current parameters)}
 \FOR{$i \in 1, \dots, t$} 
    \STATE $\psi = \psi - \eta_\psi \nabla_{\psi}  \ell^{\psi}(\psi) $
        \hfill\emph{\color{gray}(Compute the extrapolated $\psi$)}
\ENDFOR  
 \STATE $\psi = \psi^{\textit{copy}} - \eta_\psi \nabla_{\psi}  \ell^{\psi}(\psi ) $     \hfill\emph{\color{gray}(update $\psi$)}
 \STATE {\bfseries Output:} $\psi$

\end{algorithmic}
\label{alg:extragradient}
\end{algorithm}

\subsubsection{Pseudocode for Nested Lookahead for a Two-Player Game}\label{app:la-minmax}

For completeness, in Algorithm~\ref{alg:nested_lookahead_minimax} we give the details of adapted version of the nested Lookahead-Minmax algorithm proposed in~\citep[Algorithm 6,][]{chavdarova2021lamm} with two-levels.

In the given algorithm,  the actor and critic parameters are first updated using a gradient-based optimizer. At interval $k^{(1)}$, backtracking is done between the current parameters and first-level copies (slow weights) and they get updated. At interval $k^{(2)}= c_j k^{(1)}$ backtracking is performed again with second-level copies (slower weights), updating both first- and second-level copies with the averaged version.

\begin{algorithm}[tbh]
\begin{algorithmic}[1]
   \STATE {\bfseries Input:} 
               number of episodes $t$,
               learning rates $\eta_\vtheta, \eta_\vw$, 
               initial weights $\{\vtheta, \vtheta^{(1)}, \vtheta^{(2)}\}$ and $\{(\vw, \vw^{(1)}, \vw^{(2)})\}$,
                LA hyperparameters: levels $l=2$, $(k^{(1)}$, $k^{(2)})$
                and $(\alpha_\vtheta, \alpha_\vw)$,
              losses $\ell^{\vtheta}$, $\ell^{\vw}$, 
              real--data distribution $p_d$, noise--data distribution $p_z$.
   \FOR{$r \in 1, \dots, t$}
        
            \STATE $\vx \sim p_d$, $\vz \sim p_z$ 
            \STATE $\vw \leftarrow \vw - \eta_\vw \nabla_{\vw}  \ell^{\vw}(\vw, \vx, \vz ) $     \hfill\emph{\color{gray}(update $\vw$)} 
       
        \STATE $\vtheta \leftarrow \vtheta - \eta_\vtheta \nabla_{\vtheta}  \ell^{\vtheta}(\vtheta, \vx, \vz ) $ \hfill\emph{\color{gray}(update $\vtheta$ )} 
        
        \IF{ $r \% k^{(1)} == 0$}
                \STATE $\vw \leftarrow \vw^{(1)}+ \alpha_{\vw} (\vw - \vw^{(1)}) $  \hfill\emph{\color{gray}(backtracking on interpolated line $\vw^{(1)}$, $\vw$)} 
                \STATE $\vtheta \leftarrow \vtheta^{(1)} + \alpha_{\vtheta} (\vtheta - \vtheta^{(1)}) $  \hfill\emph{\color{gray}(backtracking on interpolated line $\vtheta^{(1)}$, $\vtheta$)} 
                \STATE $(\vtheta^{(1)} , \vw^{(1)} ) 
            \leftarrow (\vtheta, \vw )$ \hfill\emph{\color{gray}(update slow checkpoints)} 
        \ENDIF
        \IF{ $r \% k^{(2)} == 0$}
                \STATE $\vw \leftarrow \vw^{(2)}+ \alpha_{\vw} (\vw - \vw^{(2)}) $  \hfill\emph{\color{gray}(backtracking on interpolated line $\vw^{(2)}$, $\vw$)} 
                \STATE $\vtheta \leftarrow \vtheta^{(2)} + \alpha_{\vtheta} (\vtheta - \vtheta^{(2)}) $  \hfill\emph{\color{gray}(backtracking on interpolated line $\vtheta^{(2)}$, $\vtheta$)} 
                \STATE $(\vtheta^{(2)} , \vw^{(2)} ) 
            \leftarrow (\vtheta, \vw )$ \hfill\emph{\color{gray}(update super-slow checkpoints)} 
                \STATE $(\vtheta_{(1)} , \vw_{(1)} ) 
            \leftarrow (\vtheta, \vw )$ \hfill\emph{\color{gray}(update slow checkpoints)}
        \ENDIF
   \ENDFOR   
   \STATE {\bfseries Output:} $\vtheta^{(2)}$, $\vw^{(2)}$   
\end{algorithmic}
   \caption{Pseudocode of Two-Level Nested Lookahead--Minmax.~\citep{chavdarova2021lamm}}
   \label{alg:nested_lookahead_minimax}
\end{algorithm}

\subsection{MARL algorithms}\label{app:marl_details}

\subsubsection{Details on the MADDPG Algorithm}\label{app:maddpg_details}

The MADDPG algorithm \citep{lowe2017multi} is outlined in Algorithm~\ref{alg:maddpg}.
An empty replay buffer $\mathcal{D}$ is initialized to store experiences.
In each episode, the environment is reset and agents choose actions to perform accordingly. After, experiences in the form of \emph{(state, action, reward, next state)} are saved to $\mathcal{D}$.

After a predetermined number of random iterations, learning begins by sampling batches from $\mathcal{D}$.
The critic of agent $i$ receives the sampled joint actions $\va$ of all agents and the state information of agent $i$ to output the predicted $Q$-value of agent $i$. Deep Q-learning~\citep{mnih2015humanlevel} is then used to update the critic network; lines \ref{line21}--\ref{line22}.  
Then, the agents' policy network is optimized using policy gradient; refer to \ref{line24}. 
Finally, following each learning iteration, the target networks are updated towards current actor and critic networks using a fraction $\tau$. Then the process repeats until the end of training.

All networks are optimized using the Adam optimizer~\citep{kingma2014adam}. Once training is complete, each agent's actor operates independently during execution. This approach is applicable across cooperative, competitive, and mixed environments.

\begin{algorithm}
    \begin{algorithmic}[1]
    \STATE {\bfseries Input:}
              Environment $\mathcal{E}$,
              number of agents $n$,
              number of episodes $t$,
              action spaces $\{\A_i\}_{i=1}^n$,
              number of random steps $t_{\text{rand}}$ before learning,
              learning interval $t_{\text{learn}}$,
              actor networks $\{ \vmu_i \}_{i=1}^n$, with initial weights $\vtheta \equiv \{ \vtheta_i \}_{i=1}^n$,
              critic networks $\{ \mathbf{Q}_i \}_{i=1}^n$ with initial weights $\vw \equiv \{ \vw_i \}_{i=1}^n$,
              learning rates $\eta_\vtheta, \eta_\vw$,
              optimizer $B$ (e.g., Adam),
              discount factor $\gamma$,
              soft update parameter $\tau$.
    \STATE {\bfseries Initialize:}
    \begin{ALC@g}
        \STATE Replay buffer $\mathcal{D} \leftarrow \varnothing$ 
    \end{ALC@g}
  
   \FORALL{episode $e \in 1, \dots, t$}
        \STATE $\vx \leftarrow \textit{Sample}(\mathcal{E})$ \hfill\emph{\color{gray}(sample from environment $\mathcal{E}$}) 
        \STATE $step \leftarrow 1$
        \REPEAT \label{line:maddpg}
            \IF{$e \leq t_{\text{rand}}$ } 
                \STATE for each agent $i$, $a_i \sim \A_i$ \hfill\emph{\color{gray}(sample actions randomly)}
            \ELSE
                \STATE for each agent $i$, select action $a_i = \vmu_i(\vo_i)+\mathcal{N}_t$ using current policy and exploration noise
            \ENDIF
            \STATE Execute actions $\va = (a_1,\dots,a_n)$, observe rewards $\mathbf{r}$ and new state $\vx'$ \hfill\emph{\color{gray}(apply actions and record results)}
            \STATE replay buffer $ \mathcal{D} \leftarrow (\vx, \va, \mathbf{r}, \vx')$\alglinelabel{line-buffer}
            \STATE $\vx \leftarrow \vx'$
            
            \STATE \hfill\emph{\color{gray}(apply learning step if applicable)}
            \IF{$step \% t_{\text{learn}} = 0$}
                \FORALL{agent $i \in 1,\dots,n$}
                    \STATE sample batch $\mathcal{B}:  \{ (\vx^j, \va^j, \mathbf{r}^j, \vx'^j )\}_{j=1}^{|\B|}$ 
                    from $\mathcal{D}$ 
                       \STATE $y^j \leftarrow r^j_i + \gamma\mathbf{Q}^{\bar\vmu}(\vx'^j,a'_1,\dots,a'_n)$, where $a'_k = \bar\vmu_k(\vo'^{j}_k)$ \alglinelabel{line21}
                    \STATE Update critic by minimizing the loss (using optimizer $B$ ): \\ \hspace{2em} $\ell(\vw_i) = \frac{1}{|\B|}\sum_j{\left(y^j - \mathbf{Q}^\vmu_i(\vx^j,a^j_1,\dots,a^j_n)\right)^2}$  \alglinelabel{line22}\\
                    \STATE Update actor policy using policy gradient formula and optimizer $B$
                    \STATE $\nabla_{\vtheta_i}J \approx \frac{1}{|\B|}\sum_j\nabla_{\vtheta_i}\vmu_i(\vo^j_i)\nabla_{a_i}\mathbf{Q}^\vmu_i(\vx^j,a_1^j,\dots,a_i,\dots,a_n^j)$, where $a_i = \vmu_i(\vo^j_i)$ \alglinelabel{line24}
                \ENDFOR
                \FORALL{agent $i \in [n]$}
                    \STATE $\bar\vtheta_i \leftarrow \tau\vtheta_i + (1 - \tau)\bar\vtheta_i$
                    \hfill\emph{\color{gray}(update target networks)}
                    \STATE $\bar\vw_i \leftarrow \tau\vw_i + (1 - \tau)\bar\vw_i$
                \ENDFOR
            \ENDIF
            \STATE $step \leftarrow step + 1$
        \UNTIL{environment terminates}
    \ENDFOR
    \STATE {\bfseries Output:} $\vtheta$, $\vw$  
    \end{algorithmic}
    \caption{Pseudocode for MADDPG~\citep{lowe2017multi}.}
   \label{alg:maddpg}
\end{algorithm}

\subsubsection{MATD3 Algorithm}
We provide a psuedo code for MATD3 algorithm from \citep{Ackermann2019ReducingOB} in algorithm \ref{alg:matd3}. 
As discussed in the main section, MATD3 was introduced as an improvement to MADDPG and follows a similar structure, except for the learning steps. After sampling a batch from the replay buffer $\mathcal{D}$, both critics of each agent are updated using Deep Q-learning, with the target computed using the minimum of the two critic values (notice the difference in lines \ref{line21}  and \ref{line21matd3}of the two algorithms). The actor networks are then updated via policy gradient, using only the Q-value from the first critic; see line \ref{line25matd3}.

\begin{algorithm}
    \begin{algorithmic}[1]
    \STATE {\bfseries Input:}
              Environment $\mathcal{E}$,
              number of agents $n$,
              number of episodes $t$,
              action spaces $\{\A_i\}_{i=1}^n$,
              number of random steps $t_{\text{rand}}$ before learning,
              learning interval $t_{\text{learn}}$,
              actor networks $\{ \vmu_i \}_{i=1}^n$, with initial weights $\vtheta \equiv \{ \vtheta_i \}_{i=1}^n$,
             both critic networks, $\{ \mathbf{Q}_{i,1}, \mathbf{Q}_{i,2} \}_{i=1}^n$ with initial weights $\vw \equiv \{ \vw_{i,1}, \vw_{i,2} \}_{i=1}^n$,
              learning rates $\eta_\vtheta, \eta_\vw$,
              optimizer $B$ (e.g., Adam),
              discount factor $\gamma$,
              soft update parameter $\tau$,
              policy update frequency $p$.
    \STATE {\bfseries Initialize:}
    \begin{ALC@g}
        \STATE Replay buffer $\mathcal{D} \leftarrow \varnothing$
    \end{ALC@g}
  
   \FORALL{episode $e \in 1, \dots, t$}
        \STATE $\vx \leftarrow \textit{Sample}(\mathcal{E})$ \hfill\emph{\color{gray}(sample from environment $\mathcal{E}$)}
        \STATE $step \leftarrow 1$
        \REPEAT \label{line:matd3}
            \IF{$e \leq t_{\text{rand}}$ } 
                \STATE for each agent $i$, $a_i \sim \A_i$ \hfill\emph{\color{gray}(sample actions randomly)}
            \ELSE
                \STATE for each agent $i$, select action $a_i = \vmu_i(\vo_i)+\epsilon$ using current policy with some exploration noise
            \ENDIF
            
            \STATE Execute actions $\va = (a_1,\dots,a_n)$, observe rewards $\mathbf{r}$ and new state $\vx'$ \hfill\emph{\color{gray}(apply actions and record results)}
            \STATE replay buffer $ \mathcal{D} \leftarrow (\vx, \va, \mathbf{r}, \vx')$
            \STATE $\vx \leftarrow \vx'$
            
            \STATE \hfill\emph{\color{gray}(apply learning step if applicable)}
            \IF{$step \% t_{\text{learn}} = 0$} 
            
                \FORALL{agent $i \in [n]$}
                    \STATE sample batch $\{ (\vx^j, \va^j, \mathbf{r}^j, \vx'^j )\}_{j=1}^{|\B|}$ 
                    from $\mathcal{D}$ 
                       \STATE $y^j \leftarrow r^j_i + \gamma\min_{m=1,2}\mathbf{Q}^{\bar\vmu}_{i,m}(\vx'^j,a'_1,\dots,a'_n)$, where $a'_k = \bar\vmu_k(\vo'^j_k)+\epsilon$ \alglinelabel{line21matd3}
                    \STATE Update both critics, $m=1,2$ by minimizing the loss (using optimizer $B$ ): \\ \hspace{2em} $\ell(\vw_{i,m}) = \frac{1}{|\B|}\sum_j{\left(y^j - \mathbf{Q}^\vmu_{i,m}(\vx^j,a^j_1,\dots,a^j_n)\right)^2}$  
                    \IF{$step \% p = 0$}
                    \STATE Update actor policy using policy gradient formula and optimizer $B$
                    \STATE $\nabla_{\vtheta_i}J \approx \frac{1}{|\B|}\sum_j\nabla_{\vtheta_i}\vmu_i(\vo^j_i)\nabla_{a_i}\mathbf{Q}^\vmu_{i,1}(\vx^j,a_1^j,\dots,a_i,\dots,a_n^j)$, where $a_i = \vmu_i(\vo^j_i)$\alglinelabel{line25matd3}
                    \STATE $\bar\vtheta_i \leftarrow \tau\vtheta_i + (1 - \tau)\bar\vtheta_i$
                    \hfill\emph{\color{gray}(update target networks)}
                    \STATE $\bar\vw_{i,m} \leftarrow \tau\vw_{i,m} + (1 - \tau)\bar\vw_{i,m}$
                    \ENDIF
                \ENDFOR
            \ENDIF
            \STATE $step \leftarrow step + 1$
        \UNTIL{environment terminates}
    \ENDFOR
    \STATE {\bfseries Output:} $\vtheta$, $\vw$  
    \end{algorithmic}
    \caption{Pseudocode for MATD3~\citep{Ackermann2019ReducingOB}.}
   \label{alg:matd3}
\end{algorithm}

\subsubsection{Counterfactual multi-agent policy gradients (\textbf{COMA}, \citep{foerster2018counterfactual})}
COMA is an actor-critic multi-agent algorithm based on the CTDE paradigm, with one centralized critic and $n$ decentralized actors. Additionally, COMA directly addresses the credit assignment problem in multi-agent settings by: \begin{enumerate*}[series = tobecont, label=(\roman*),font=\itshape]
    \item computing a counterfactual baseline for each agent $b_i(s, \mathbf{a}_{-i})$, \item using this baseline to estimate the advantage $A_i$ of the chosen action over all others in $\A_{i}$, and \item  leveraging this advantage to update individual policies. 
\end{enumerate*} This ensures that policy updates reflect each agent's true contribution to the overall reward.

\subsubsection{Multi-agent Trust Region Policy Optimization (\textbf{MATRPO}, \citep{li2023multiagenttrustregionpolicy})}

Trust Region Policy Optimization \citep[TRPO,][]{Schulman2015TrustRP} is a policy optimization method that ensures stable updates by constraining policy changes within a trust region. This constraint is enforced using the KL-divergence, and the update step is computed using natural gradient descent.

Extending TRPO to the cooperative multi-agent setting introduces challenges due to non-stationarity. To address this, MATRPO employs a centralized critic, represented by a central value function $V(\vs)$, which leverages shared information among agents to estimate the Generalized Advantage Estimator (GAE) $A_i$. The advantage function is then used in the policy gradient update, while ensuring that the KL-divergence constraint is respected, maintaining stable and coordinated learning across agents.

\subsubsection{Multi-agent Proximal policy Optimization \citep[MAPPO,][]{Yu2021TheSE}}

One of the widely used algorithms in practice is MAPPO, an extension of Proximal Policy Optimization \citep[PPO,][]{Schulman2017ProximalPO} to the multi-agent setting. Similar to TRPO, PPO ensures that policy updates remain within a small, stable region, but instead of enforcing a KL-divergence constraint, it uses clipping. This clipping mechanism simplifies the optimization process, allowing updates to be performed efficiently using standard gradient ascent methods.

MAPPO is an on-policy algorithm that employs a centralized critic while maintaining decentralized actor networks for each agent. Its critic update follows the same rule as MATRPO, but for the policy update, it optimizes a clipped surrogate objective, which restricts the policy update step size, ensuring stable and efficient learning.

\section{VI MARL Perspectives \& Details on the Proposed Algorithms} \label{app:vi_presp}
In the main text, we introduced the general VI operator for multi-agent actor-critic algorithms \eqref{eq:F_marl} and provided the specific equations for MADDPG in (\ref{eq:cr_maddpg} \& \ref{eq:ac_maddpg}), with the operator corresponding to: 
\begin{equation}
\tag{$F_{\text{MADDPG}}$}\label{eq:F_marl}
F_{\text{MADDPG}} \Big(\begin{bmatrix} \vdots \\
\vw_i\\
\vtheta_i\\
\vdots\\
\end{bmatrix}   \Big) {\equiv}
\begin{bmatrix} \vdots \\
\nabla_{\vw_i} \Big( { \frac{1}{|\B|}\sum\limits_{j=1}^{|\B|}{ \left( 
r_i^j + \gamma\mathbf{Q}_i^{\bar\vmu}\left.(\vx'^j,\va'; \vw_i')  \right|_{\va' = \bar\vmu(\vo'^j)}
- \mathbf{Q}^\vmu_i(\vx^j,\va^j ; \vw_i )
\right)^2
}
}   
\Big)\\ 
\nabla_{\vtheta_i} \Big( { \frac{1}{|\B|}\sum\limits_{j=1}^{|\B|}{ 
\vmu_i(\vo^j_i; \vtheta_i)\nabla_{a_i}\left. \mathbf{Q}^\vmu_i(\vx^j,a_1^j,\dots,a_i,\dots,a_n^j; \vw_i)\right|_{a_i=\vmu_i(\vo^j_i)}
}} 
\Big)\\
 \vdots \\
\end{bmatrix} \,,
\end{equation}
\vspace{.3em} 
where the parameter space is $\mathcal{Z}\equiv \R^d$, with $d=\sum_{i=1}^n (d_i^{Q} + d_i^{\mu})$.

We now present the VI operators of additional MARL algorithms within the centralized critic CTDE paradigm.
After, we provide detailed versions of Algorithm \ref{alg:nested_lookahead_marl} for MADDPG and MATD3, outlining the full training process when incorporating LA or LA-EG.

\subsection{VI MARL Perspectives}\label{app:vip}
We show how update equations for several well-known MARL algorithms---that follow the CTDE paradigm with a centralized critic---can be written as a VI. Our VI-based methods can also be applied to these algorithms using the operators below.

For a more general notation, for each agent $i \in [n]$ we assume: 
\begin{enumerate}[label=(\roman*),font=\itshape] 
\item central critic network (one or multiple) that estimates either action value $Q$--Network($\vs,\va$): $\mathbf{Q}_i(\vx_t,\va_t ; \vw_i )$, or state value $V$--network(s): $\mathbf{V}_i(\vx_t; \vw_i )$, and 
\item a decentralized policy network that can be deterministic $\vmu_i(\vo_i;\vtheta_i)$ or stochastic $\vpi_i(\vo_i;\vtheta_i)$, depending on the algorithm.
\end{enumerate}
Given a batch of experiences $\B$: $(\vx^j, \va^j, \mathbf{r}^j, \vx'^j )$, sampled from a replay buffer ($\mathcal{D}$), we provide the necessary equations and the final operator ($F$) for each of the following popular MARL algorithms.

\subsubsection{MATD3}\label{app:vip-matd3}

The VI formulation for MATD3 is very similar to MADDPG, except here, for each agent we have two critic networks, we write: $\vw_i \equiv \{\vw_{i,1},\vw_{i,2}\}$. Accordingly, target computation for the critic ($Q_{i,m}$) is calculated by taking the minimum of both critic networks, but only the value of critic $1$ is used for the actor (policy network) update. We have:
\begin{equation}
\tag{$F_{\text{MATD3}}$}\label{eq:F_matd3}
F_{\text{MATD3}} \Big(\begin{bmatrix} \vdots \\
\vw_{i,1}\\
\vw_{i,2}\\
\vtheta_i\\
\vdots\\
\end{bmatrix}   \Big) {\equiv}
\begin{bmatrix} \vdots \\ 
\nabla_{\vw_{i,1}} \Big( { \frac{1}{|\B|}\sum\limits_{j=1}^{|\B|}{ \left(
\underbrace{r^j_i + \gamma\min_{m \in\{1,2\}}\mathbf{Q}^{\bar\vmu}_{i,m}\left.(\vx'^j,a'_1,\dots,a'_n) \right|_{\va' = \bar\vmu(\vo'^j)}}_{\text{target } y_i}
- \mathbf{Q}^\vmu_{i,1}(\vx^j,\va^j ; \vw_{i,1} )\right)^2}} \Big)\\  
\nabla_{\vw_{i,2}} \Big( { \frac{1}{|\B|}\sum\limits_{j=1}^{|\B|}{ \left( 
\underbrace{r^j_i + \gamma\min_{m \in\{1,2\}}\mathbf{Q}^{\bar\vmu}_{i,m}\left.(\vx'^j,a'_1,\dots,a'_n) \right|_{\va' = \bar\vmu(\vo'^j)}}_{\text{target } y_i}
- \mathbf{Q}^\vmu_{i,2}(\vx^j,\va^j ; \vw_{i,2} )\right)^2}} \Big)\\  
\Big( { \frac{1}{|\B|}\sum\limits_{j=1}^{|\B|}{\nabla_{\vtheta_i} \vmu_i(\vo^j_i; \vtheta_i)\nabla_{a_i}\left.\mathbf{Q}^\vmu_{i,1}(\vx^j,a_1^j,\dots,a_i,\dots,a_n^j)\right|_{a_i=\vmu_i(\vo^j_i)}}} \Big)\\     
 \vdots \\
\end{bmatrix} \,.
\end{equation}

\subsubsection{COMA}
In COMA, critic is trained using a $TD(\lambda)$ target ($y^\lambda$) computed from a target network parameterized by $\bar\vw$ that get updated to main network weights every couple iterations. Given the following Advantage $A_i$ calculations:
\[
A_i(\vx, \va) = Q(\vx, \va) - b_i(\vx, \va_{-i})
\]
\[
b_i(\vx, \va_{-i}) = \sum_{a_i} \vpi_i(a_i | \vo_i) Q(\vx, (a_i, \va_{-i}))\,,
\]
the operator for \emph{COMA} corresponds to:

\begin{equation}
\tag{$F_{\text{COMA}}$}\label{eq:F_coma}
F_{\text{COMA}} \Big(\begin{bmatrix} \vdots \\
\vw_i\\
\vtheta_i\\
\vdots\\
\end{bmatrix}   \Big) {\equiv}
\begin{bmatrix} \vdots \\
\nabla_{\vw_i} \mathbb{E}   \left[ \left(  y^\lambda_i - Q_{i}(\vx^j, \va; \vw_i) \right)^2\right]\\ 
{ \mathbb{E} \left[ \nabla_{\vtheta_i} \sum_{i} A_i(\vx, \va) \log \vpi_{\vtheta_i}(a_i | \vo_i)\right]}\\
 \vdots \\
\end{bmatrix} \,.
\end{equation}

\subsection{Detailed Algorithms}

Herein we provide two pseudocodes; considered as extended versions of the main algorithm in algorithm \ref{alg:nested_lookahead_marl}; in which we detail how the lookahead approach can be integrated in the training process of MADDPG and MATD3.

\subsubsection{Extended version of LA-MADDPG pseudocode}
We include an extended version for the LA-MADDPG algorithm without VI notations in algorithm \ref{alg:nested_lookahead_maddpg_extended}.

\begin{algorithm*}
\begin{algorithmic}[1]
   \STATE {\bfseries Input:}
              Environment $\mathcal{E}$,
              number of agents $n$,
              number of episodes $t$,
              action spaces $\{\A_i\}_{i=1}^n$,
              number of random steps $t_{\text{rand}}$ before learning,
              learning interval $t_{\text{learn}}$,
              actor networks $\{ \vmu_i \}_{i=1}^n$, with initial weights $\vtheta \equiv \{ \vtheta_i \}_{i=1}^n$,
              critic networks $\{ \mathbf{Q}_i \}_{i=1}^n$ with initial weights $\vw \equiv \{ \vw_i \}_{i=1}^n$,
              learning rates $\eta_\vtheta, \eta_\vw$,
              base optimizer $B$ (e.g., Adam),
              discount factor $\gamma$,
              lookahead hyperparameters $\mathcal{L}\equiv (l, \{k^{(j)}\}_{j=1}^l, \alpha_\vtheta, \alpha_\vw )$,
              soft update parameter $\tau$.
   \STATE {\bfseries Initialize:}
   \begin{ALC@g}
        \STATE Replay buffer $\mathcal{D} \leftarrow \varnothing$ 
        \STATE LA parameters: $\vphi \leftarrow \{ \vtheta \}_{\times l}, \{\vw \}_{\times l} $  \hfill\emph{\color{gray}(store snapshots for LA)}
    \end{ALC@g}
  
   \FORALL{episode $e \in 1, \dots, t$}
        \STATE $\vx \leftarrow \textit{Sample}(\mathcal{E})$ \hfill\emph{\color{gray}(sample from environment $\mathcal{E}$)}
        \STATE $step \leftarrow 1$
        \REPEAT
            \IF{$e \leq t_{\text{rand}}$ } 
                \STATE for each agent $i$, $a_i \sim \A_i$ \hfill\emph{\color{gray}(sample actions randomly)}
            \ELSE
                \STATE for each agent $i$, select action $a_i$ using current policy and exploration
            \ENDIF
            \STATE \hfill\emph{\color{gray}(apply actions and record results)}
            \STATE Execute actions $\va = (a_1,\dots,a_n)$, observe rewards $\mathbf{r}$ and new state $\vx'$
            \STATE replay buffer $ \mathcal{D} \leftarrow (\vx, \va, \mathbf{r}, \vx')$
            \STATE $\vx \leftarrow \vx'$
            
            \STATE \hfill\emph{\color{gray}(apply learning step if applicable)}
            \IF{$step \% t_{\text{learn}} = 0$}
                \FORALL{agents $i \in 1,\dots,n$}
                    \STATE sample batch $\{ (\vx^j, \va^j, \mathbf{r}^j, \vx'^j )\}_{j=1}^{|\B|}$ 
                    from $\mathcal{D}$
                        \STATE $y^j \leftarrow r^j_i + \gamma\mathbf{Q}^{\bar\vmu}(\vx'^j,a'_1,\dots,a'_n)$, where $a'_k = \bar\vmu_k(\vo'^j_k)$
                    \STATE Update critic by minimizing the loss $\ell(\vw_i) = \frac{1}{|\B|}\sum_j{\left(y^j - \mathbf{Q}^\vmu_i(\vx^j,a^j_1,\dots,a^j_n)\right)^2}$ using $B$
                    \STATE Update actor policy using policy gradient formula $B$
                   \STATE $\nabla_{\vtheta_i}J \approx \frac{1}{|\B|}\sum_j\nabla_{\vtheta_i}\vmu_i(\vo^j_i)\nabla_{a_i}\mathbf{Q}^\vmu_i(\vx^j,a^j_1,\dots,a_i,\dots,a^j_n)$, where $a_i = \vmu_i(\vo^j_i)$\
                \ENDFOR
                \FORALL{agents $i \in [n]$}
                    \STATE $\bar\vtheta_i \leftarrow \tau\vtheta_i + (1 - \tau)\bar\vtheta_i$
                    \hfill\emph{\color{gray}(update target networks)}
                    \STATE $\bar\vw_i \leftarrow \tau\vw_i + (1 - \tau)\bar\vw_i$
                \ENDFOR
            \ENDIF
            \STATE $step \leftarrow step + 1$
        \UNTIL{environment terminates}
         \STATE {\scshape NestedLookAhead}($n, e,
        \vphi,
        \mathcal{L}
        $)
   \ENDFOR   
   \STATE {\bfseries Output:} $\vtheta$, $\vw$   
\end{algorithmic}
   \caption{Pseudocode for LA--MADDPG: MADDPG with (Nested) Lookahead.}
   \label{alg:nested_lookahead_maddpg_extended}
\end{algorithm*}

\subsubsection{Extended version of LA-MATD3 pseudocode}
We include an extended version for the LA-MATD3 algorithm without VI notations in algorithm \ref{alg:nested_lookahead_matd3_extended}.

\begin{algorithm*}
\begin{algorithmic}[1]
   \STATE {\bfseries Input:}
              Environment $\mathcal{E}$,
              number of agents $n$,
              number of episodes $t$,
              action spaces $\{\A_i\}_{i=1}^n$,
              number of random steps $t_{\text{rand}}$ before learning,
              learning interval $t_{\text{learn}}$,
              actor networks $\{ \vmu_i \}_{i=1}^n$, with initial weights $\vtheta \equiv \{ \vtheta_i \}_{i=1}^n$,
              both critic networks, $\{ \mathbf{Q}_{i,1}, \mathbf{Q}_{i,2} \}_{i=1}^n$ with initial weights $\vw \equiv \{ \vw_{i,1}, \vw_{i,2} \}_{i=1}^n$,
              learning rates $\eta_\vtheta, \eta_\vw$,
              base optimizer $B$ (e.g., Adam),
              discount factor $\gamma$,
              lookahead hyperparameters $\mathcal{L}\equiv (l, \{k^{(j)}\}_{j=1}^l, \alpha_\vtheta, \alpha_\vw )$,
              soft update parameter $\tau$,
              policy update frequency $p$.
   \STATE {\bfseries Initialize:}
   \begin{ALC@g}
        \STATE Replay buffer $\mathcal{D} \leftarrow \varnothing$ 
        \STATE LA parameters: $\vphi \leftarrow \{ \vtheta \}_{\times l}, \{\vw \}_{\times l} $  \hfill\emph{\color{gray}(store snapshots for LA)}
    \end{ALC@g}

   \FORALL{episode $e \in 1, \dots, t$}
        \STATE $\vx \leftarrow \textit{Sample}(\mathcal{E})$ \hfill\emph{\color{gray}(sample from environment $\mathcal{E}$)} 
        \STATE $step \leftarrow 1$
        \REPEAT
            \IF{$e \leq t_{\text{rand}}$ } 
                \STATE for each agent $i$, $a_i \sim \A_i$ \hfill\emph{\color{gray}(sample actions randomly)}
            \ELSE
                \STATE for each agent $i$, select action $a_i$ using current policy and exploration
            \ENDIF
            \STATE \hfill\emph{\color{gray}(apply actions and record results)}
            \STATE Execute actions $\va = (a_1,\dots,a_n)$, observe rewards $\mathbf{r}$ and new state $\vx'$
            \STATE replay buffer $ \mathcal{D} \leftarrow (\vx, \va, \mathbf{r}, \vx')$
            \STATE $\vx \leftarrow \vx'$
            
            \STATE \hfill\emph{\color{gray}(apply learning step if applicable)}
            \IF{$step \% t_{\text{learn}} = 0$}
                \FORALL{agent $i \in [n]$}
                    \STATE sample batch $\{ (\vx^j, \va^j, \mathbf{r}^j, \vx'^j )\}_{j=1}^{|\B|}$ 
                    from $\mathcal{D}$ 
                       \STATE $y^j \leftarrow r^j_i + \gamma\min_{m=1,2}\mathbf{Q}^{\bar\vmu}_{i,l}(\vx'^j,a'_1,\dots,a'_n)$, where $a'_k = \bar\vmu_k(\vo'^j_k)+\epsilon$ 
                    \STATE Update both critics, $m=1,2$ by minimizing the loss (using optimizer $B$ ): \\ \hspace{2em} $\ell(\vw_{i,m}) = \frac{1}{|\B|}\sum_j{\left(y^j - \mathbf{Q}^\vmu_{i,m}(\vx^j,a^j_1,\dots,a^j_n)\right)^2}$  
                    \IF{$step \% p = 0$}
                    \STATE Update actor policy using policy gradient formula and optimizer $B$
                    \STATE $\nabla_{\vtheta_i}J \approx \frac{1}{|\B|}\sum_j\nabla_{\vtheta_i}\vmu_i(\vo^j_i)\nabla_{a_i}\mathbf{Q}^\vmu_{i,1}(\vx^j,a_1^j,\dots,a_i,\dots,a_n^j)$, where $a_i = \vmu_i(\vo^j_i)$
                    \STATE $\bar\vtheta_i \leftarrow \tau\vtheta_i + (1 - \tau)\bar\vtheta_i$
                    \hfill\emph{\color{gray}(update target networks)}
                    \STATE $\bar\vw_{i,m} \leftarrow \tau\vw_{i,m} + (1 - \tau)\bar\vw_{i,m}$
                    \ENDIF
                \ENDFOR
            \ENDIF
            \STATE $step \leftarrow step + 1$
        \UNTIL{environment terminates}
         \STATE {\scshape NestedLookAhead}($n, e,
        \vphi,
        \mathcal{L}%
        $)
   \ENDFOR   
   \STATE {\bfseries Output:} $\vtheta$, $\vw$   
\end{algorithmic}
   \caption{Pseudocode for LA--MATD3: MATD3 with (Nested) Lookahead.}
   \label{alg:nested_lookahead_matd3_extended}
\end{algorithm*}

\section{Details On The Implementation} \label{app:implementation}

We used the configurations and hyperparameters from the original MADDPG paper for our implementation. For completeness, these are listed in Table \ref{table:hps}. We ran $t=60000$ training episodes for all environments, with a maximum of $25$ environment steps ($step$) per episode.

In all experiments, we used a $2$-layer MLP with $64$ units per layer. ReLU activation was applied between layers for both the policy and value networks of all agents.

\subsection{Hyperparameter Selection for Lookahead}\label{app:hyperparam-selection}

In this section, we discuss and share guidelines for hyperparameter selection based on our experiments.

\paragraph{Summary.}
\begin{itemize}
    \item We observed two- or three-level of Lookahead outperform single-level Lookahead. 
    \item Each level  $j\in[l]$ has different $k$, denoted here with $k^{(j)}$. These should be selected as multiple of the selected $k$ for the level before, that is, $k^{(j)}= c_j \cdot k^{(j-1)}$, where $ c_j$ is positive integer.
    \item We observed that for the innermost lookahead, small values for $k^{(1)}$, such as smaller than $50$,  perform better than using large values. For the outer $k^{(j)}, j>1$ large values work well, such as in the range between $5-10$ for the $ c_{j},$.
    \item We typically used $\alpha=0.5$, and we observed lower values, such as $\alpha=0.3$, give better performances then $\alpha>0.5$.
\end{itemize}

\paragraph{Discussion.}
\begin{itemize}
    \item To give an intuition regarding the above-listed conclusions, small values for $k^{(1)}$ help because the MARL setting is very noisy and the vector field is rotational. If large values are used for $k_{s}$, then the algorithm will diverge away. It is known that the combination of noise and rotational vector field can cause methods to diverge away~\citep{chavdarova2019}.
    \item Relative to the analogous conclusions for GANs~\citep{chavdarova2021lamm}, the differences is that:
    \begin{itemize}
        \item The better-performing values for $k^{(1)}$ are of a similar range as for Lookahead with GD for GANs; however they are smaller than those used for Lookahead with EG for GANs. 
    \end{itemize}
\end{itemize}

\begin{table}
  \caption{Hyperparameters used for LA-MADDPG experiments.}
  \vspace{.2cm}
  \label{table:hps}
  \centering
  \begin{tabular}{lll}
    \toprule
    \cmidrule(r){1-2}
    Name     & Description    \\
    \midrule
Adam $lr$    & $0.01$      \\
Adam $\beta_1$    & $0.9$      \\
Adam $\beta_2$    & $0.999$    \\
Batch-size        & $1024$      \\
Update ratio $\tau$  & $0.01$   \\
Discount factor $\gamma$ & $0.95$ \\
Replay Buffer & $1.5\times10^6$    \\
learning step $t_{\text{learn}}$   & $100$  \\
$t_{\text{rand}}$ & $1024$ \\
Policy update ratio (MATD3) $p$ & 2 \\
Noise std (MATD3) & 0.2 \\
Noise clip (MATD3) & 0.5 \\

Lookahead $\alpha$& $0.5$      \\
    \bottomrule
  \end{tabular}
\end{table}

\section{Additional Empirical Results}\label{app:addresults}

\subsection{Rock--paper--scissors: Buffer Structure}\label{app:buffer-conf}

For the Rock--paper--scissors (RPS) game, using a buffer size of $1$M wasn’t sufficient to store all experiences from the $60$K training episodes. We observed a change in algorithm behavior around $40$K episodes. To explore the impact of buffer configurations, we experimented with different sizes and structures, as experience storage plays a critical role in multi-agent reinforcement learning.

\noindent\textbf{Full buffer.} The buffer is configured to store all experiences from the beginning to the end of training without any loss.

\noindent\textbf{Buffer clearing.} In this setup, a smaller buffer is used, and once full, the buffer is cleared completely, and new experiences are stored from the start.

\noindent\textbf{Buffer shifting.} Similar to the small buffer setup, but once full, old experiences are replaced by new ones in a first-in-first-out (FIFO) manner.

\begin{figure*}[!htbp]
  \centering
  \begin{subfigure}[Full buffer]{\includegraphics[width=.45\textwidth,trim={.25cm .3cm .25cm .3cm},clip]{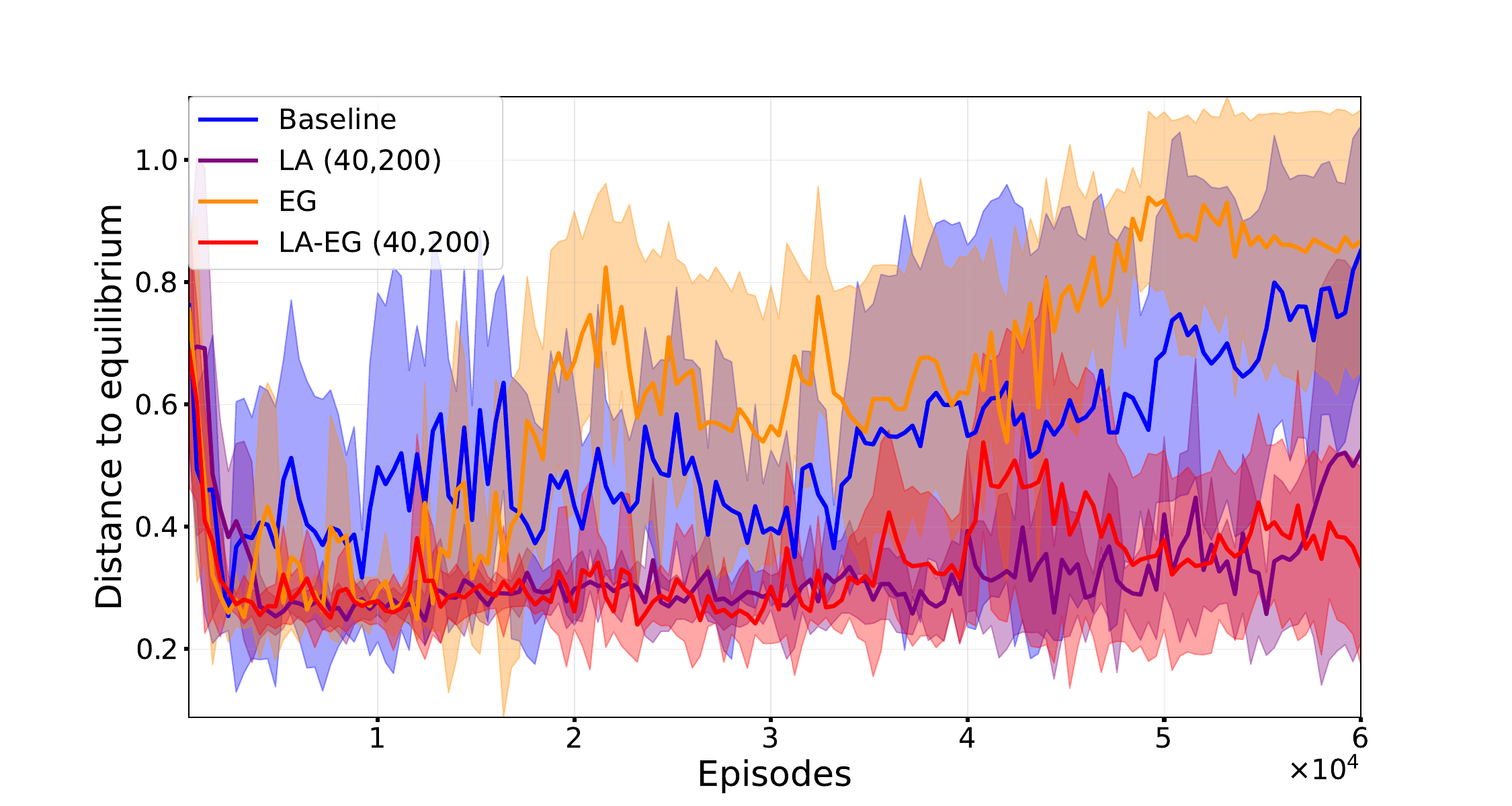}}
  \end{subfigure}
  \begin{subfigure}[Clearing buffer ($20$K)]{ \includegraphics[width=.45\textwidth,trim={.25cm .3cm .25cm .3cm},clip]{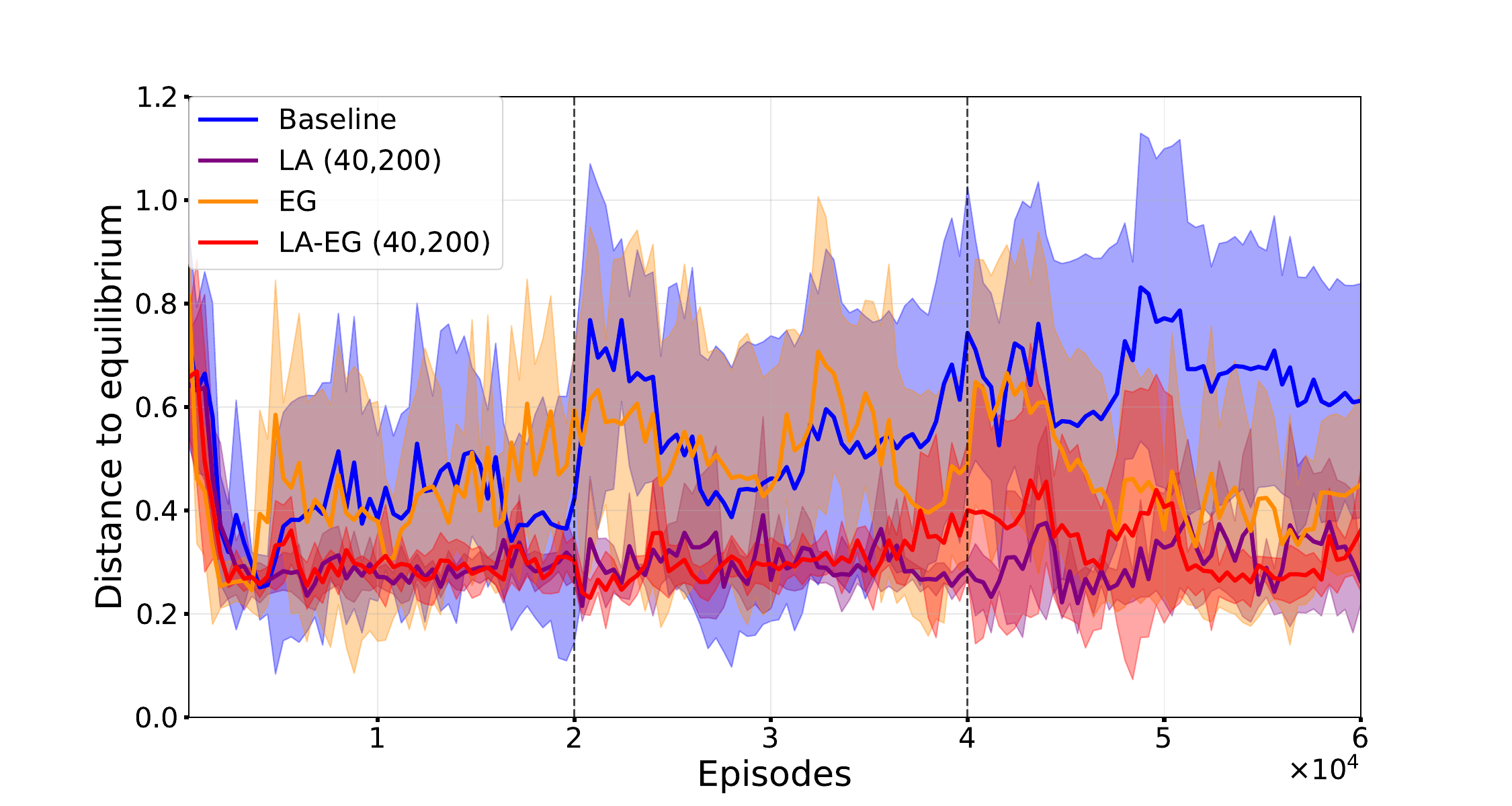}}
  \end{subfigure}

  \vspace{0cm}

  \begin{subfigure}[Shifting buffer ($20$K)]{ \includegraphics[width=.45\textwidth,trim={.25cm .3cm .25cm .3cm},clip]{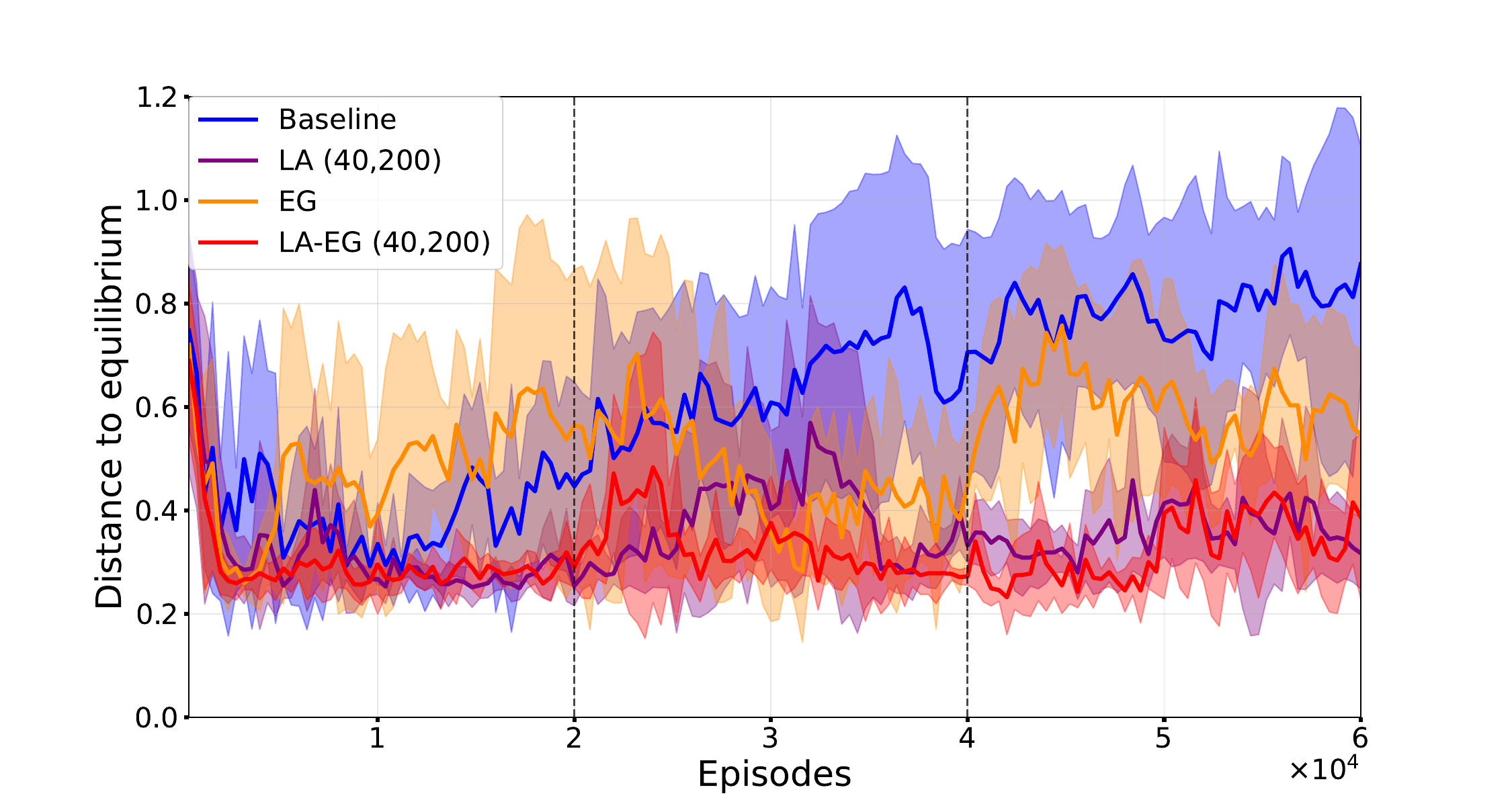}}
  \end{subfigure}
  \begin{subfigure}[Shifting buffer ($40$K)]{ \includegraphics[width=.45\textwidth,trim={.25cm .3cm .25cm .3cm},clip]{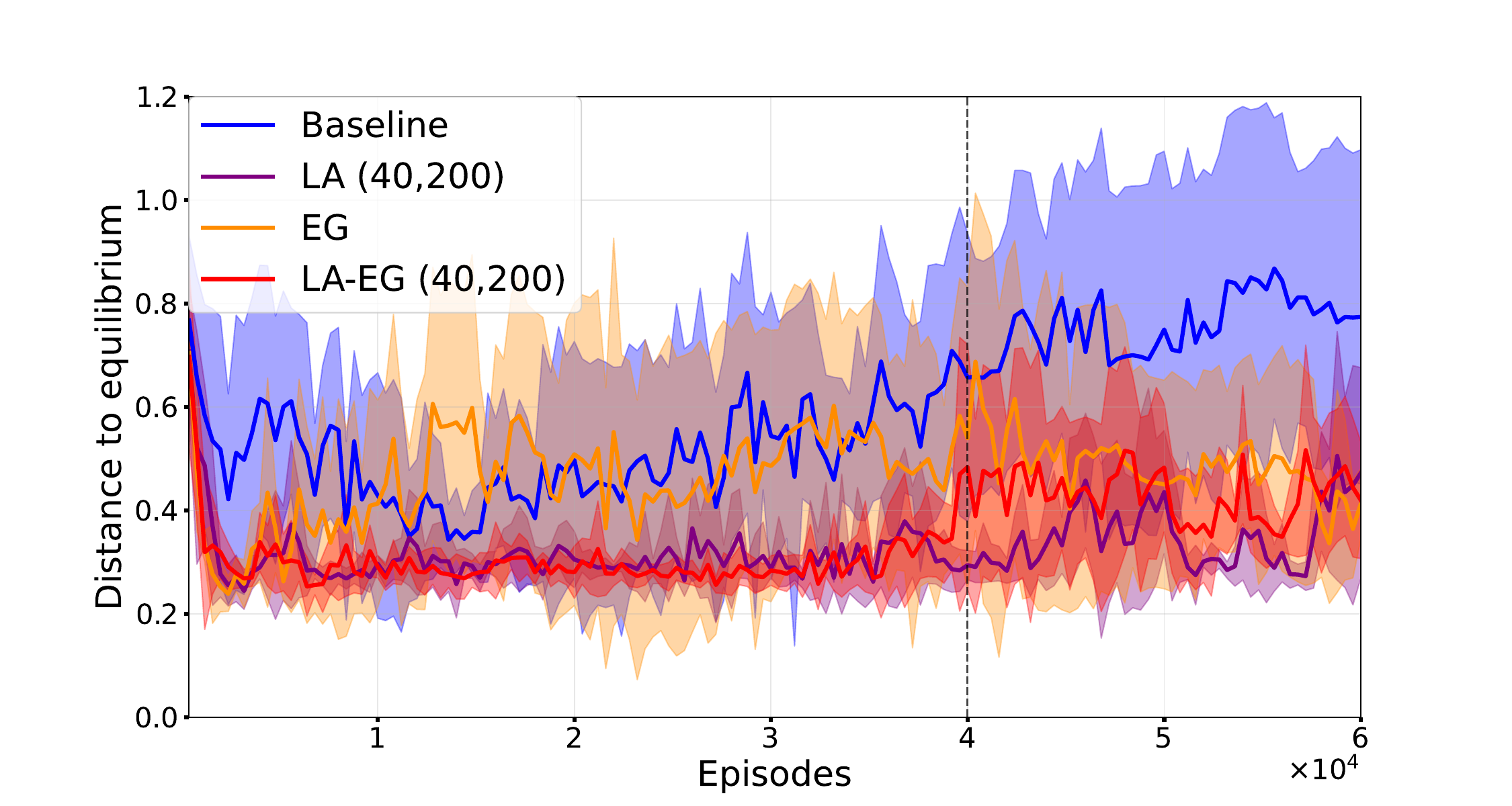}}
   \end{subfigure}

 \caption{\textbf{Comparison of different buffer configurations (see Appendix \ref{app:buffer-conf}) and methods on Rock--paper--scissors game. }
  $x$-axis: training episodes.
  $y$-axis: $5$-seed average norm between the two players' 
    policies and equilibrium policy $(\frac{1}{3}, \frac{1}{3}, \frac{1}{3})^2$. 
  The dotted line indicates the point at which the buffer begins to change, either through shifting or clearing.
 }\label{fig:diff_buffer_rps}
\end{figure*}

\noindent\textbf{Results.}
Figure~\ref{fig:diff_buffer_rps} depicts the results when using different buffer options for the RPS game. 

\subsection{Rock--paper--scissors: Scheduled learning rate}

We experimented with gradually decreasing the learning rate (LR) during training to see if it would aid convergence to the optimal policy in RPS. While this approach reduced noise in the results, it also led to increased variance across all methods except for LA-MADDPG.

\begin{figure}[tbh]
\centering
\vspace{-.5cm}
\includegraphics[width=\columnwidth,trim={.1cm .25cm .1cm .25cm},clip]{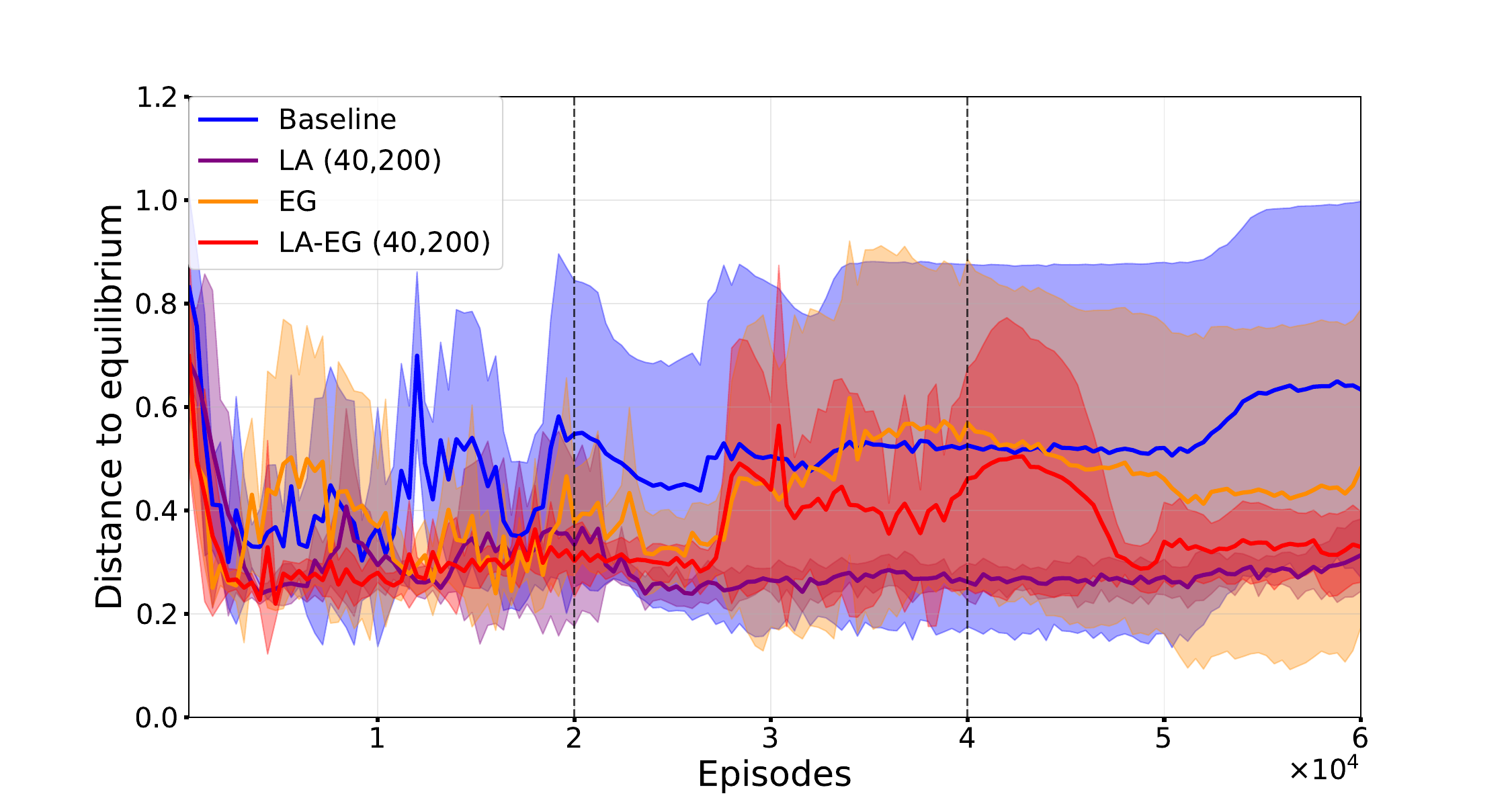}
\caption{\textbf{Compares MADDPG with different \emph{LA-MADDPG} configurations to the baseline MADDPG with (\emph{Adam}) in Rock--paper--scissors with a scheduled learning rate}. 
$x$-axis: training episodes. 
$y$-axis: $5$-seed average norm between the two players' 
policies and equilibrium policy $(\frac{1}{3}, \frac{1}{3}, \frac{1}{3})^2$. 
The dotted lines depict the times when the learning rate was decreased by a factor of $10$. 
}\label{fig:rps_learning_rates}
\end{figure}

Figure~\ref{fig:rps_learning_rates}
depicts the average distance to the equilibrium policy over $5$ different seeds for each methods, using periodically decreased step sizes.

\subsection{MPE: Predator-prey  Full results}\label{app:addresults_simpletag}

\begin{figure*}[tp]
\centering

\begin{tabular}{
  @{}
  *{5}{c@{\hspace{2pt}}}
  c
  @{}
}
\column{
\vspace{.1cm}
  \fbox{\includegraphics[width=2cm]{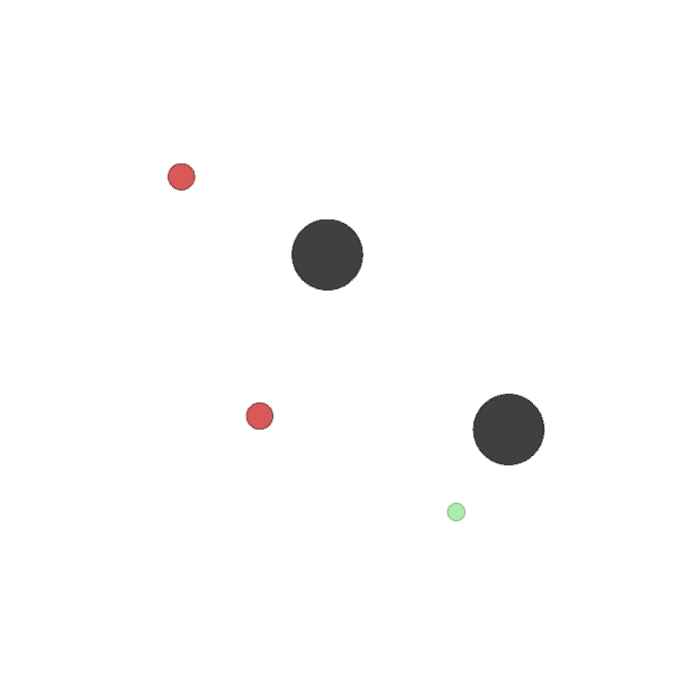}}\\
  \vspace{.1cm}
  \fbox{\includegraphics[width=2cm]{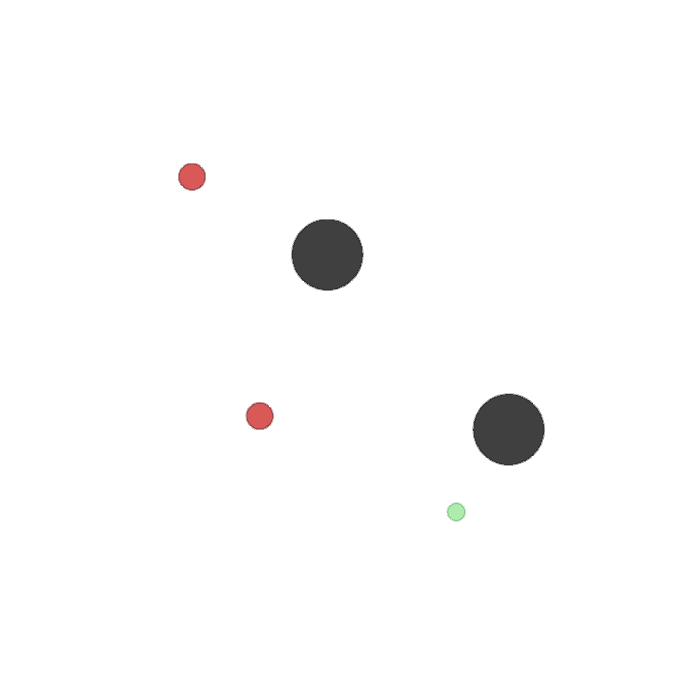}}\\
  \vspace{.1cm}
  \fbox{\includegraphics[width=2cm]{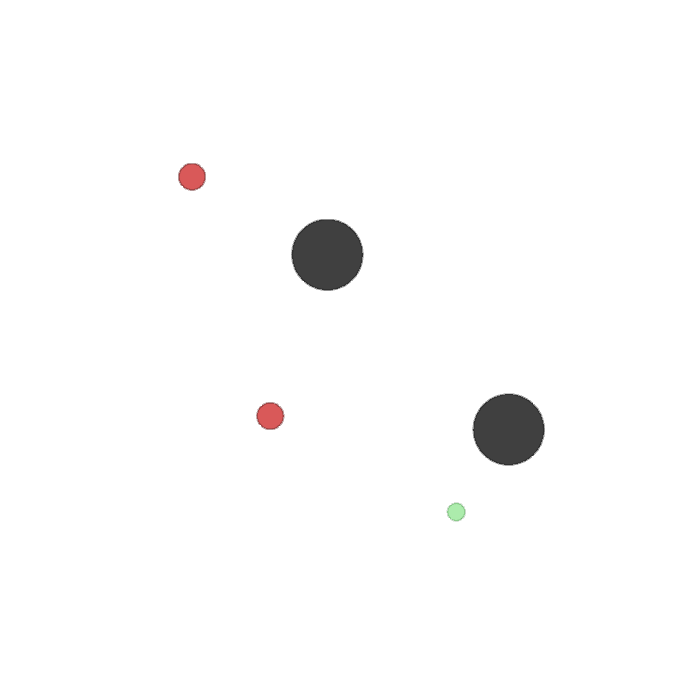}}\\
  \vspace{.1cm}
  \fbox{\includegraphics[width=2cm]{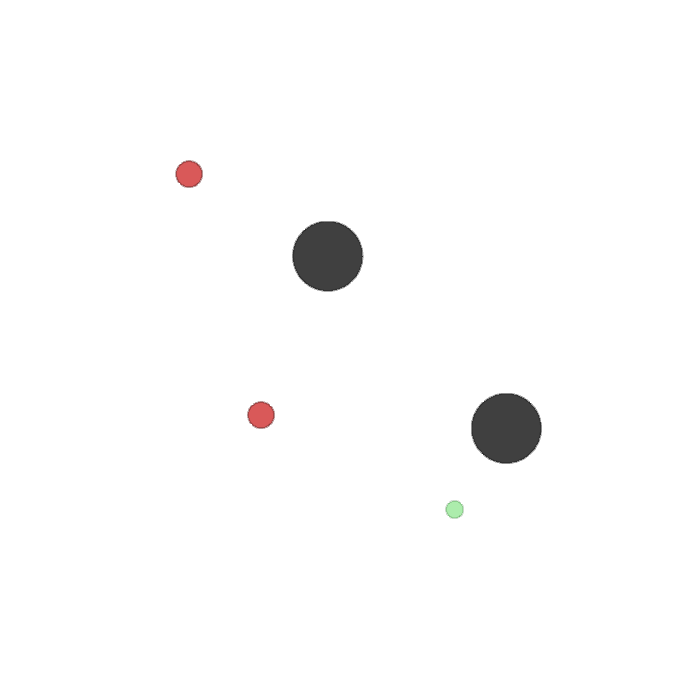}}
}{$t=1$}
&
\column{
\vspace{.1cm}
  \fbox{\includegraphics[width=2cm]{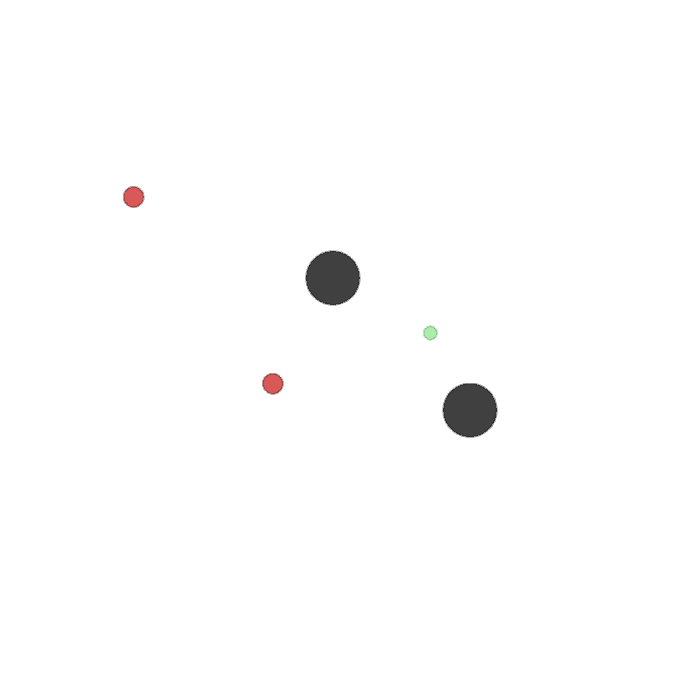}}\\
  \vspace{.1cm}
  \fbox{\includegraphics[width=2cm]{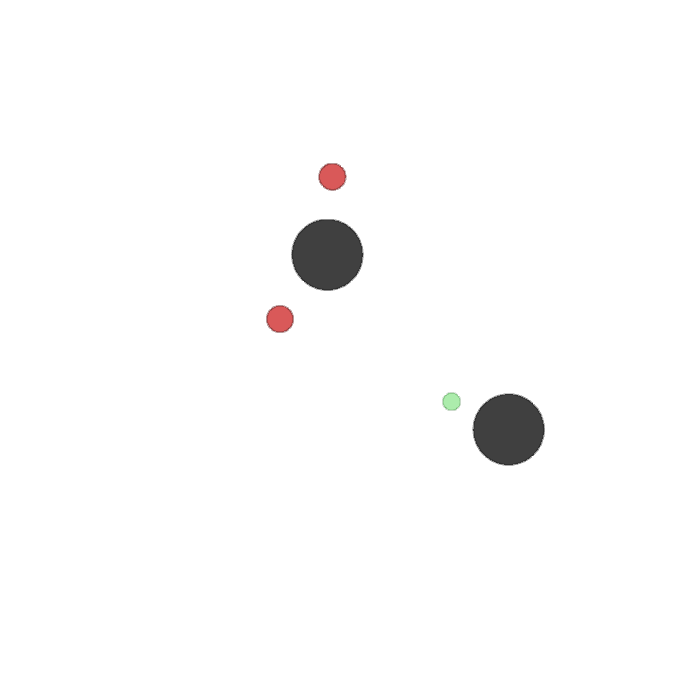}}\\
  \vspace{.1cm}
  \fbox{\includegraphics[width=2cm]{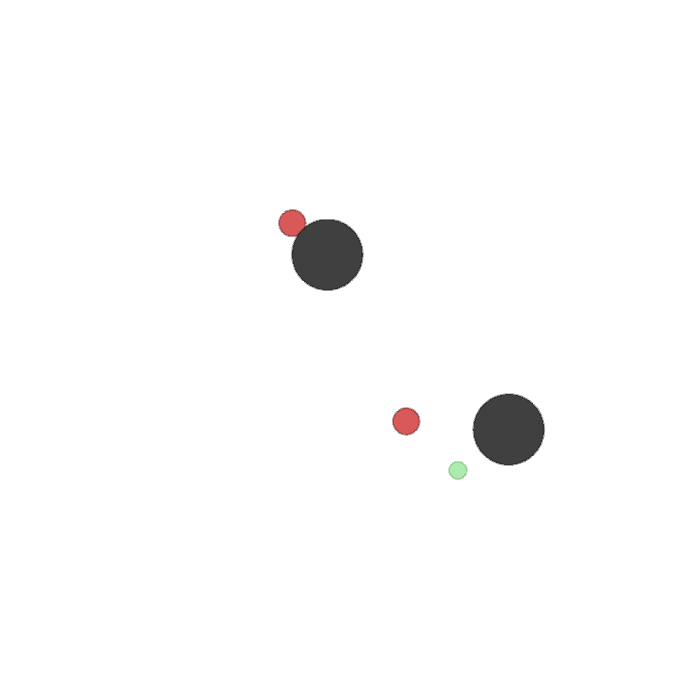}}\\
  \vspace{.1cm}
  \fbox{\includegraphics[width=2cm]{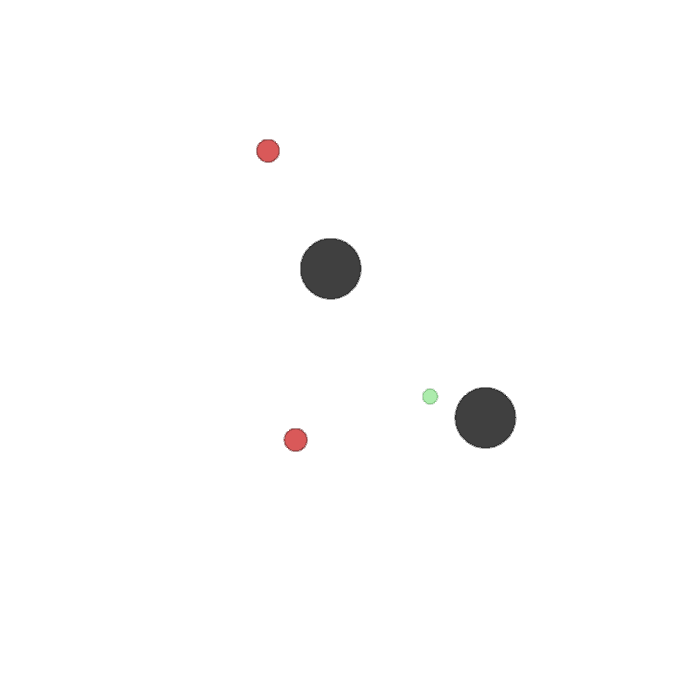}}
}{$t=10$}
&
\column{
\vspace{.1cm}
  \fbox{\includegraphics[width=2cm]{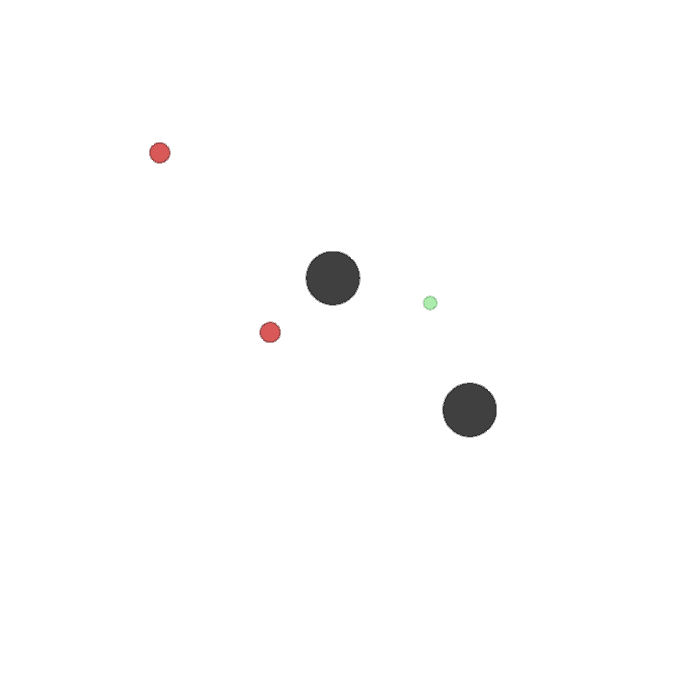}}\\
  \vspace{.1cm}
  \fbox{\includegraphics[width=2cm]{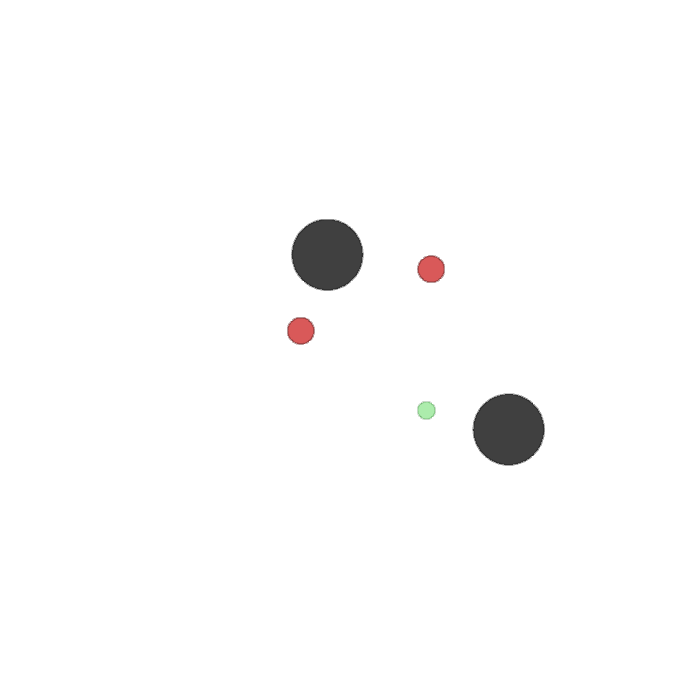}}\\
  \vspace{.1cm}
  \fbox{\includegraphics[width=2cm]{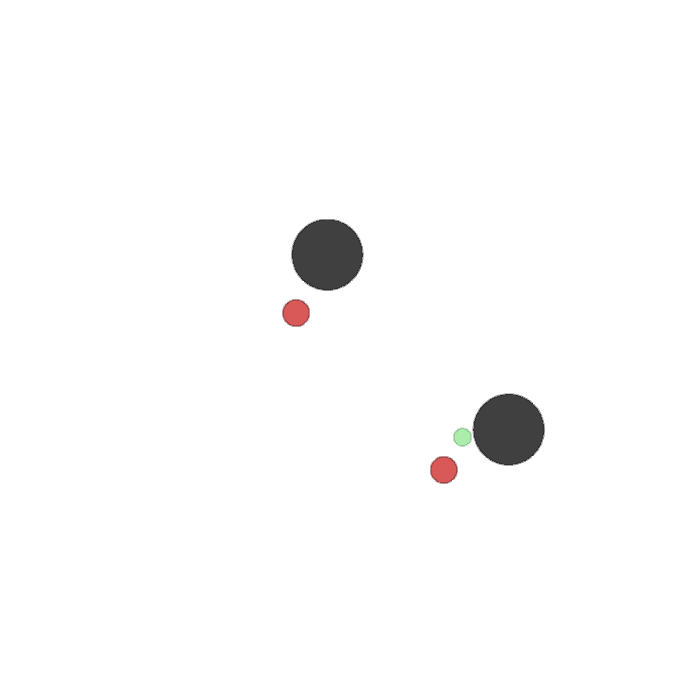}}\\
  \vspace{.1cm}
  \fbox{\includegraphics[width=2cm]{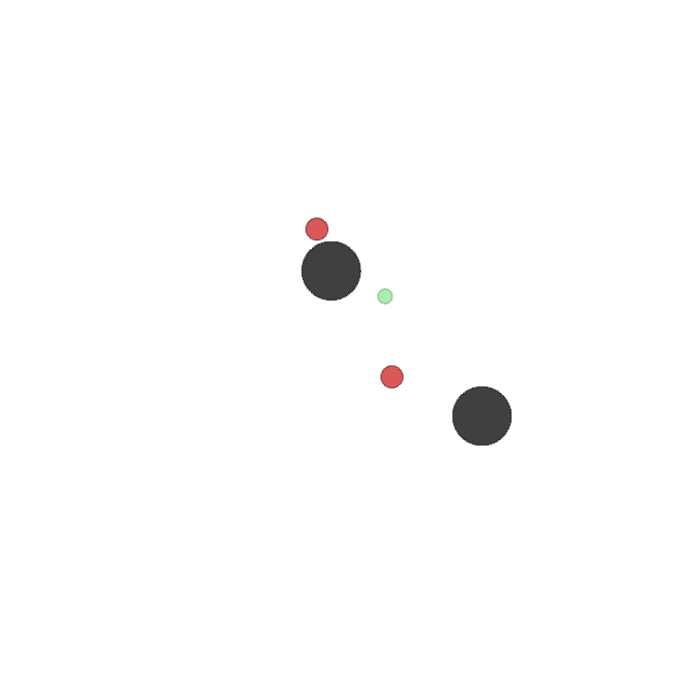}}
}{$t=20$}
&
\column{
\vspace{.1cm}
 \fbox{\includegraphics[width=2cm]{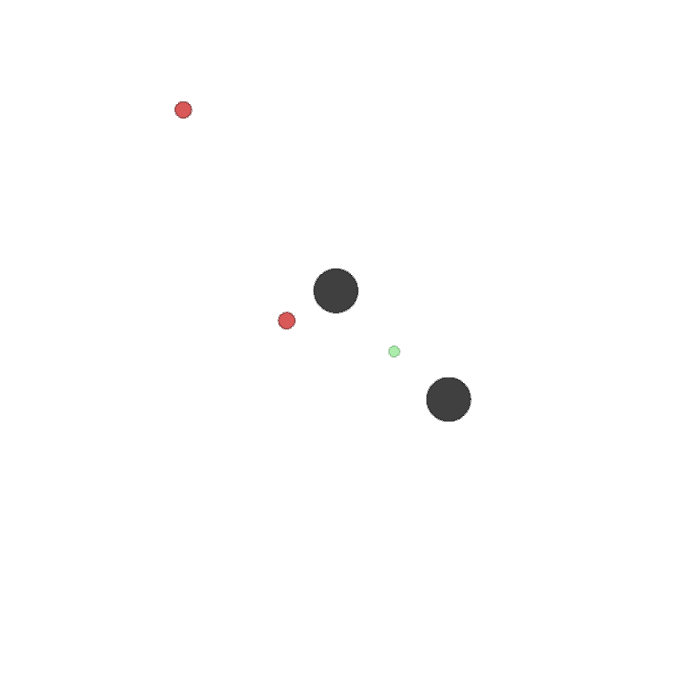}}\\
 \vspace{.1cm}
  \fbox{\includegraphics[width=2cm]{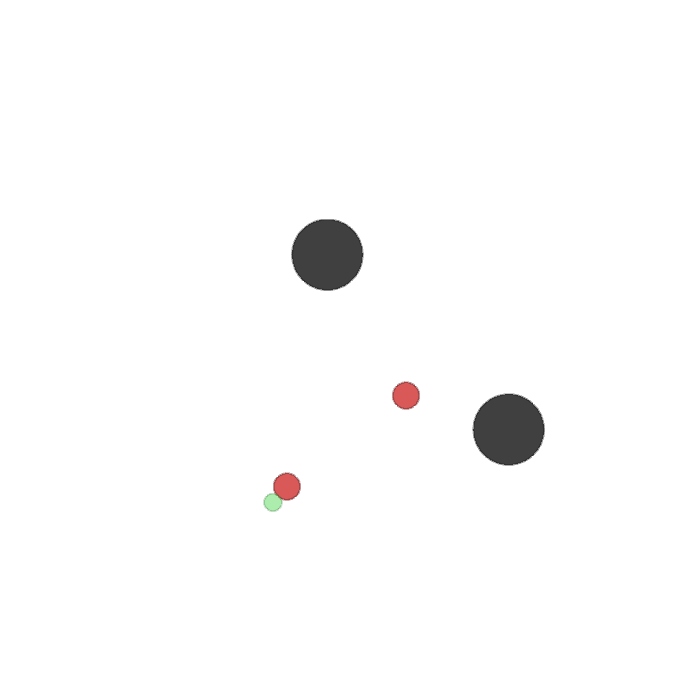}}\\
  \vspace{.1cm}
 \fbox{ \includegraphics[width=2cm]{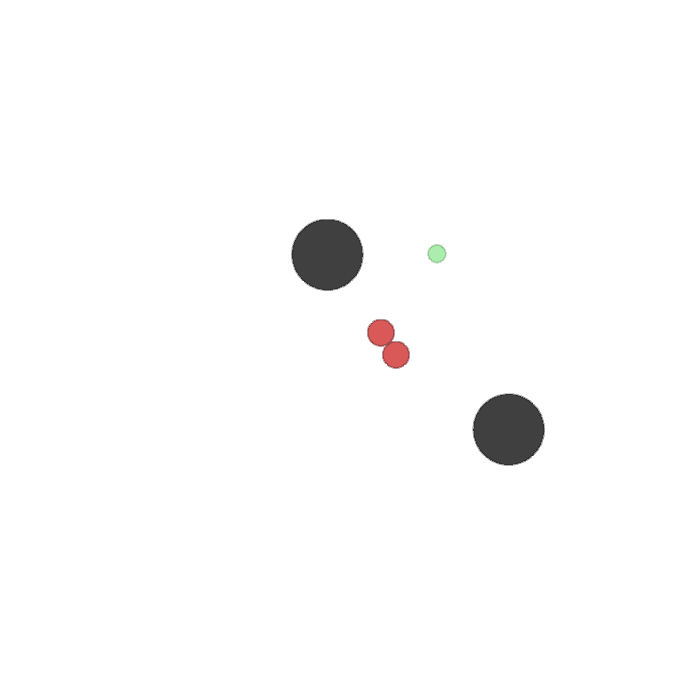 }}\\
  \vspace{.1cm}
  \fbox{\includegraphics[width=2cm]{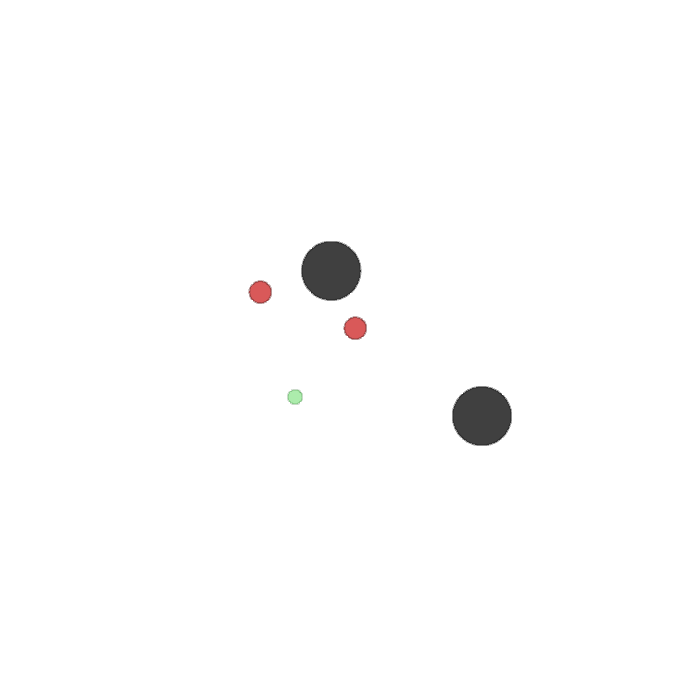}}
}{$t=30$}
&
\column{
\vspace{.1cm}
 \fbox{\includegraphics[width=2cm]{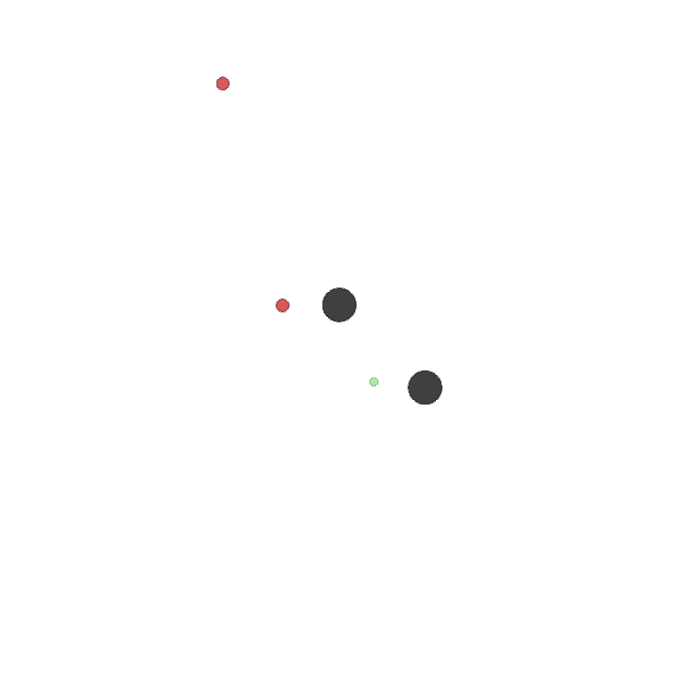}}\\
 \vspace{.1cm}
  \fbox{\includegraphics[width=2cm]{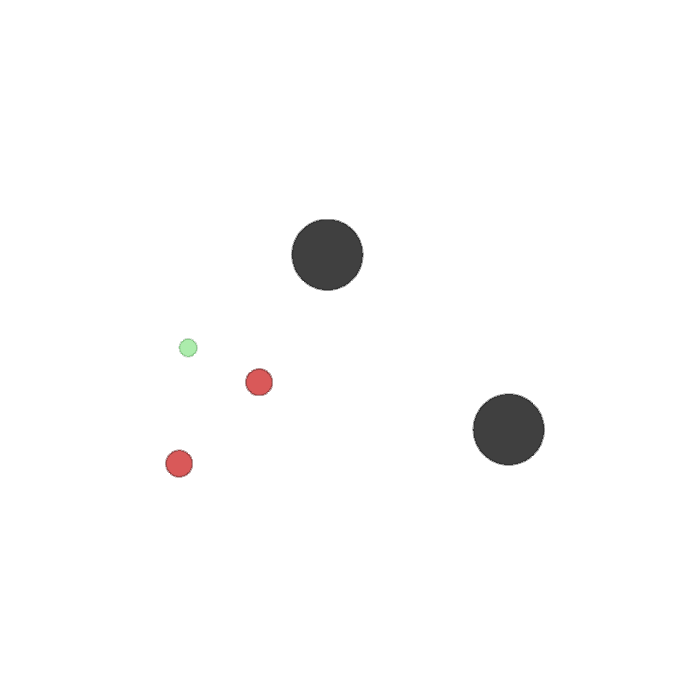}}\\
  \vspace{.1cm}
  \fbox{\includegraphics[width=2cm]{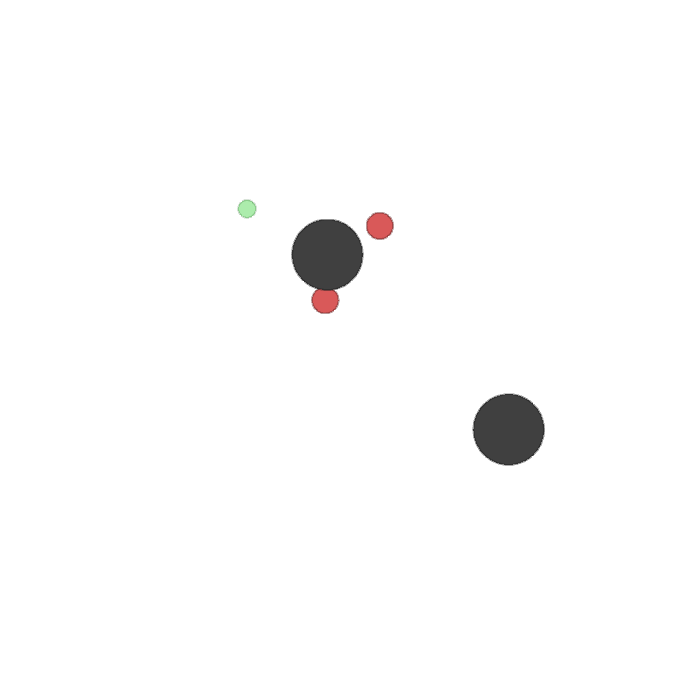}}\\
  \vspace{.1cm}
  \fbox{\includegraphics[width=2cm]{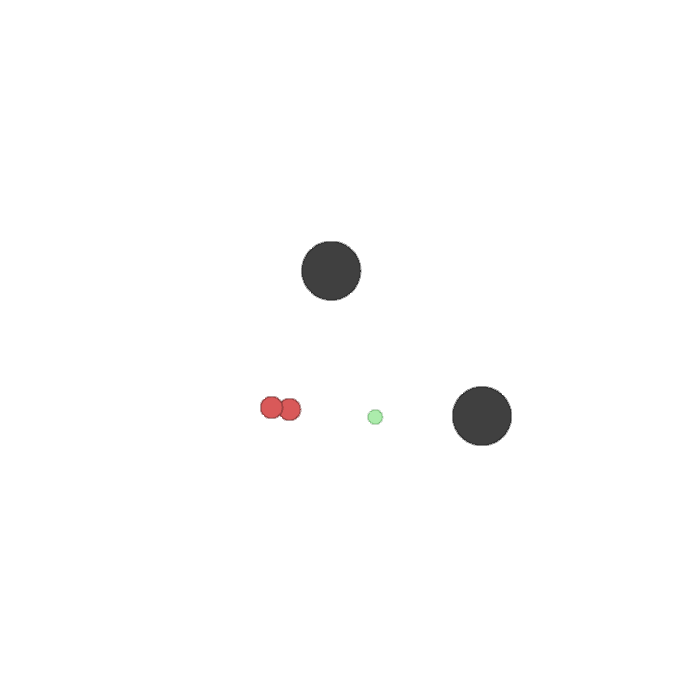}}
}{$t=40$}
&
\column{
\vspace{.1cm}
 \fbox{\includegraphics[width=2cm]{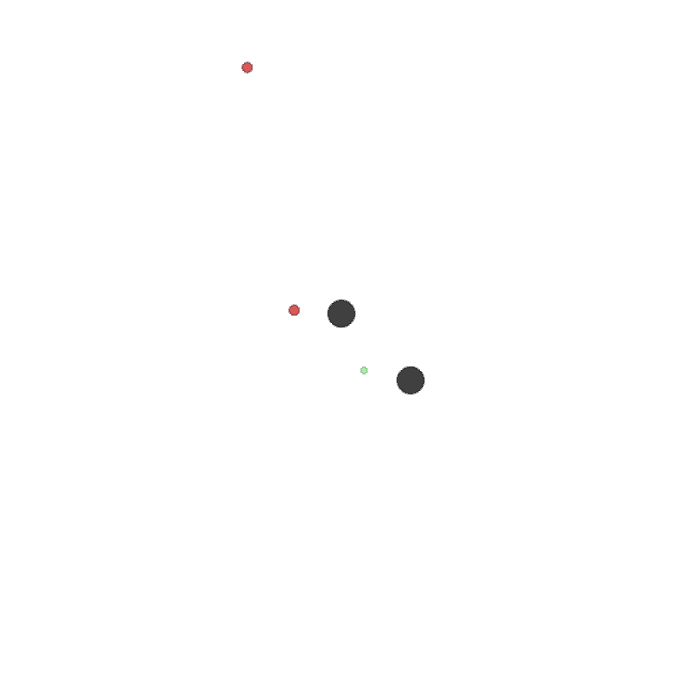}}\\
 \vspace{.1cm}
  \fbox{\includegraphics[width=2cm]{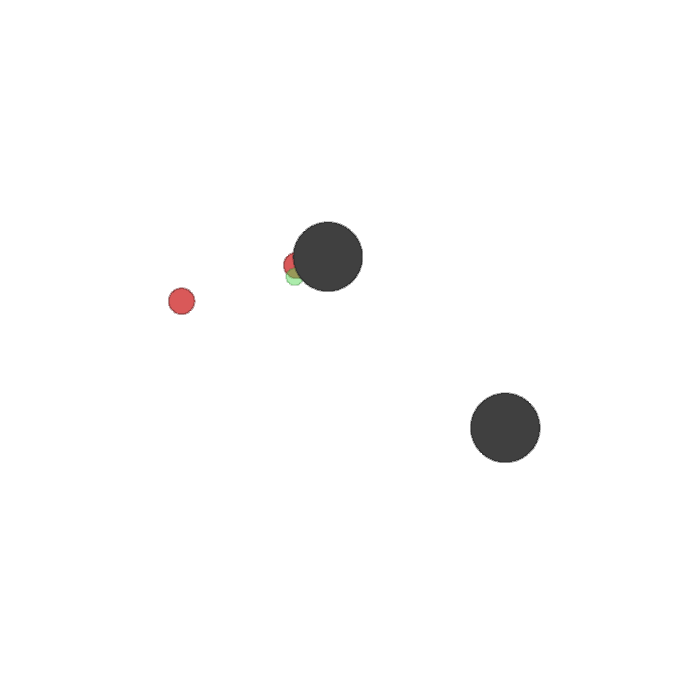}}\\
  \vspace{.1cm}
  \fbox{\includegraphics[width=2cm]{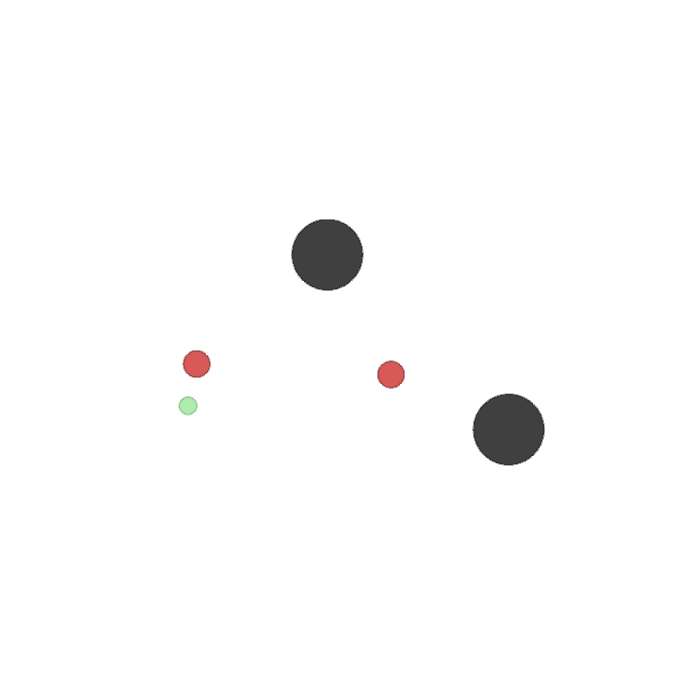}}\\
  \vspace{.1cm}
  \fbox{\includegraphics[width=2cm]{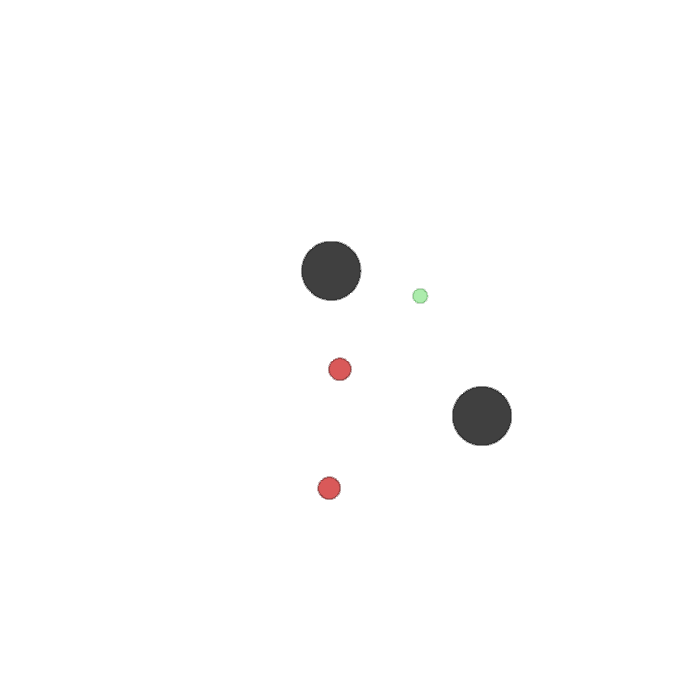}}
}{$t=50$}

\end{tabular}

\caption{\textbf{Agents' trajectories of fully trained models with all considered optimization methods on the same environment seed of MPE: Predator-prey.} Snapshots show the progress of agents as time progresses in a $50$ steps long environment. Each row contains snapshots of one method, from top to bottom: \emph{GD-MADDPG},  \emph{LA-MADDPG}, \emph{EG-MADDPG} and \emph{LA-EG-MADDPG}. Big dark circles represent landmarks, small red circles are adversary agents and green one is the good agent.}.
\label{fig:snapshots}
\end{figure*}

We also evaluated the trained models of all methods on an instance of the environment that runs for $50$ steps to compare learned policies. We present snapshots from it in Figure \ref{fig:snapshots}. Here, you can clearly anticipate the difference between the policies from baseline and our optimization methods. As in the baseline, only one agent will chase at the beginning of episode. Moreover,  for the baseline (topmost row), the agents move further away from the landmarks and the good agent, which is suboptimal. This can be noticed from the decreasing agents’ size in the figures. While in ours, both adversary agents engage in chasing the good agent until the end.

\begin{figure}[tbh]
\centering
\includegraphics[width=\columnwidth]{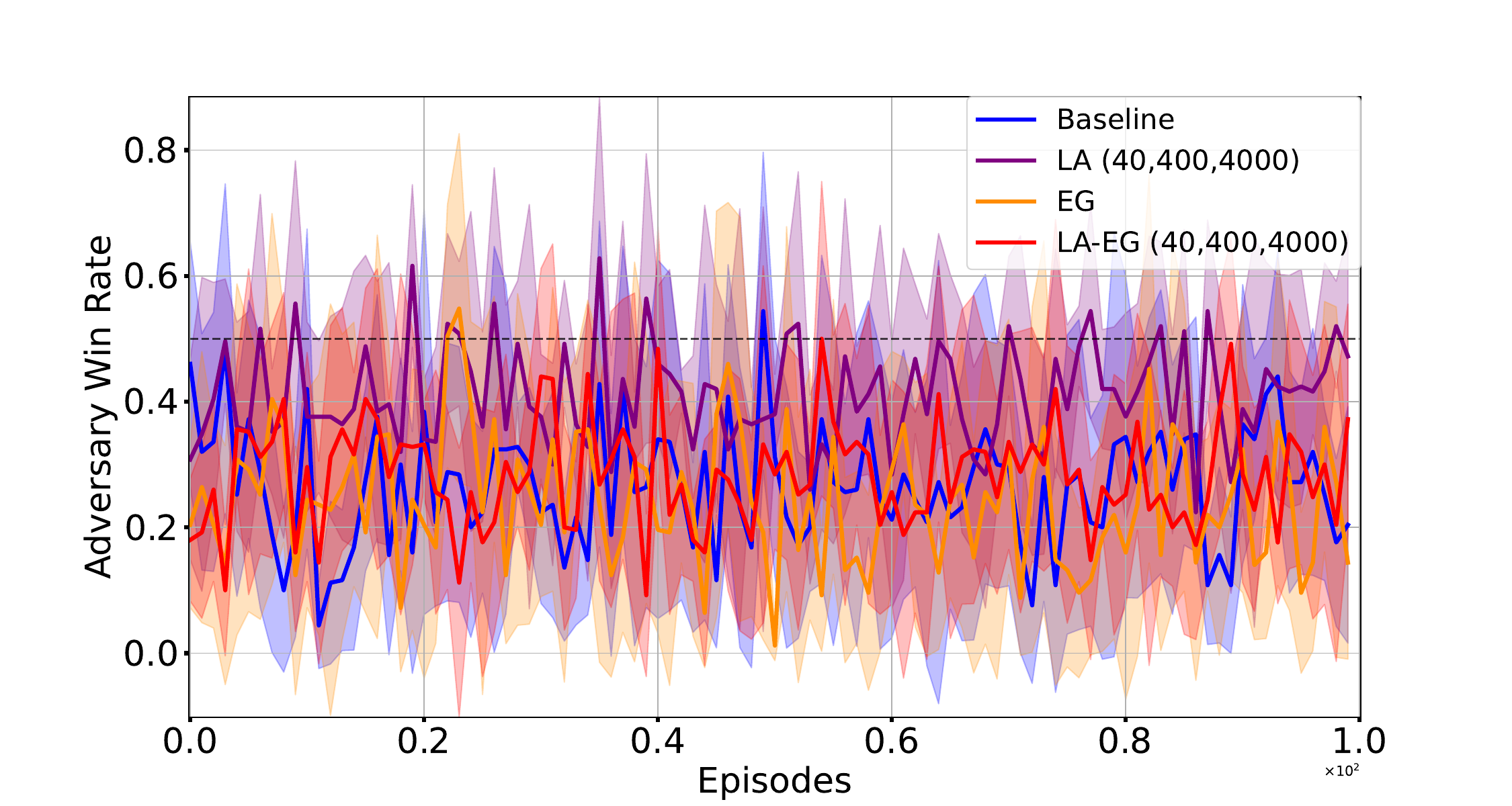}
\caption{\textbf{Comparison on the MPE--Predator-prey  game between the \emph{GD-MADDPG},  \emph{LA-MADDPG}, \emph{EG-MADDPG} and \emph{LA-EG-MADDPG} optimization methods, denoted as \textit{Baseline}, \textit{LA}, \textit{EG}, \textit{LA-EG}, resp.} 
$x$-axis: evaluation episodes. 
$y$-axis: mean adversaries win rate, averaged over $5$ runs with different seeds. 
}\label{fig:simpletag_all}
\end{figure}

\subsection{MPE: Predator-Prey and Physical deception training figures}

\begin{figure*}[!htbp]
    \centering
    \begin{subfigure}[GD-MADDPG]{ \includegraphics[width=.4\textwidth]{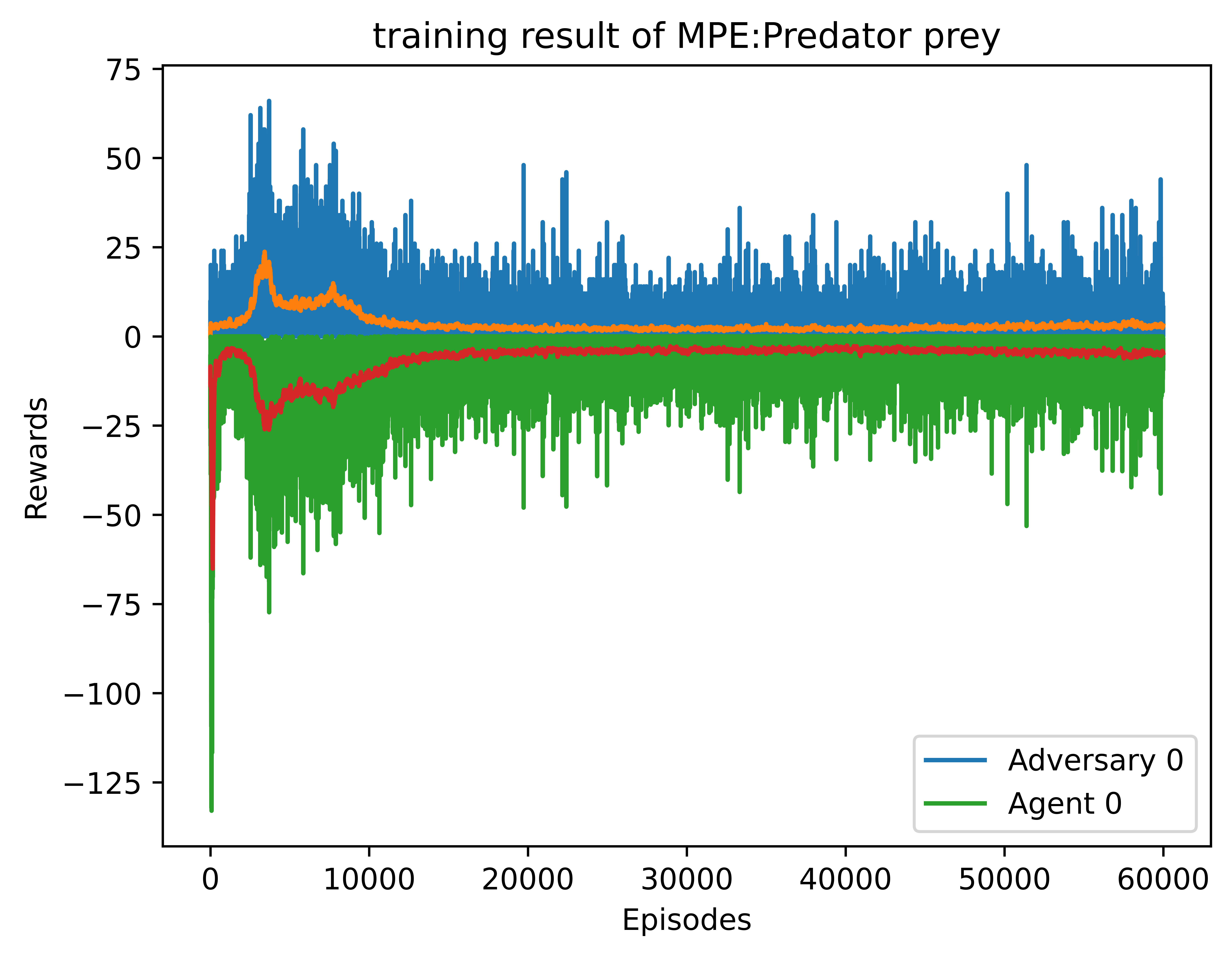}\label{fig:adam_pp}}
    \end{subfigure}
    \begin{subfigure}[LA-MADDPG]{\includegraphics[width=.4\textwidth]{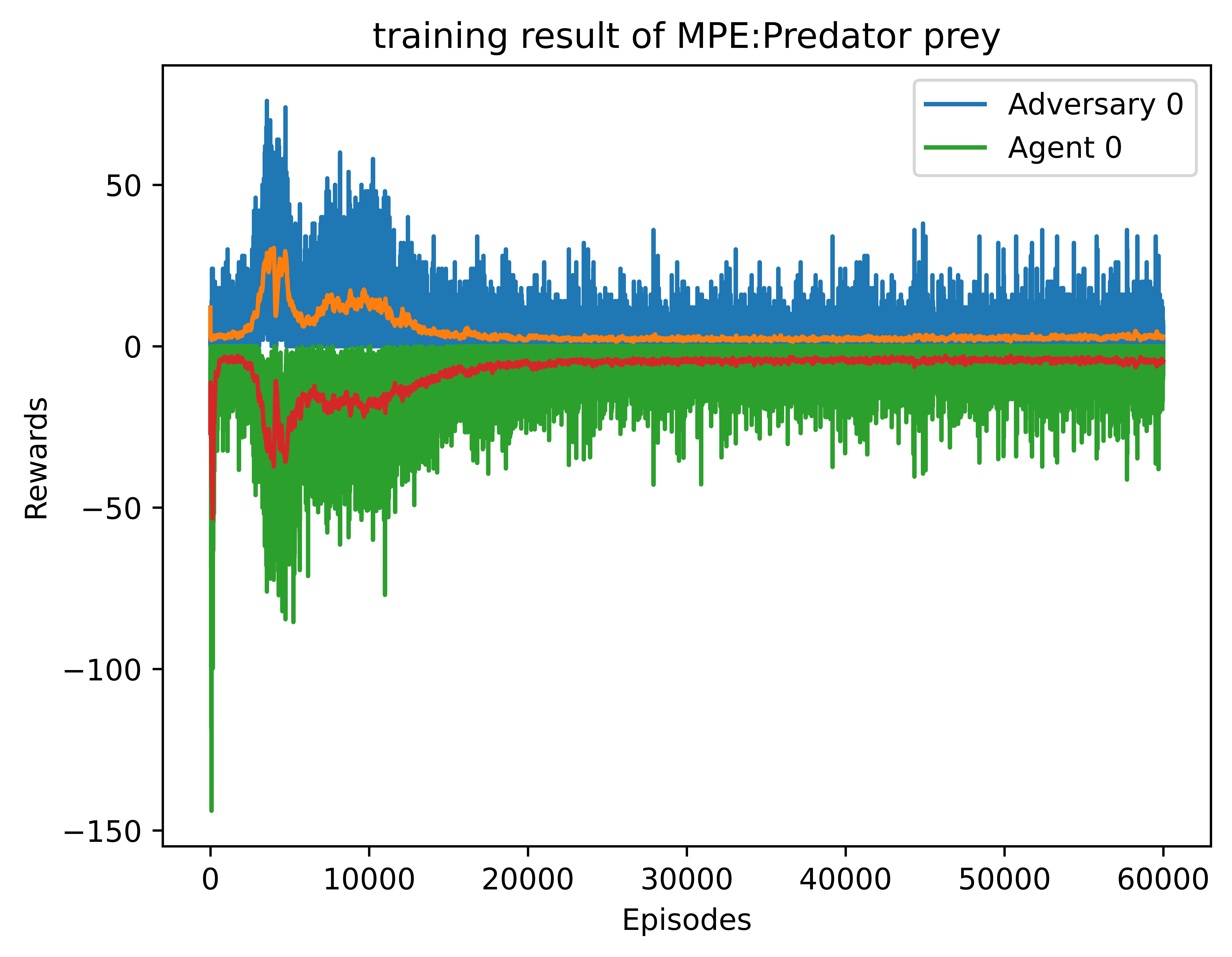}
        \label{fig:la_pp}}
    \end{subfigure}
\caption{\textbf{The figure shows the learning curves during training of GD-MADDPG and LA-MADDPG for MPE: Predator-Prey.} 
$x$-axis: training episodes. 
$y$-axis: agents rewards and their moving average with a window size of 100, calculated over $5$-seeds over $5$ seeds.
}
\end{figure*}

In figures \ref{fig:adam_pp} and \ref{fig:la_pp} we include the rewards achieved during the training of GD-MADDPG and LA-MADDPG resp. for MPE: Predator-prey. The figures show individual rewards for the agent (prey) and one adversary (predator). Blue and green show the individual rewards received at each episode while the orange and red lines are the respective running averages with window size of $100$ of those rewards.

\begin{figure*}[!htbp]
    \centering
    \begin{subfigure}[GD-MADDPG]{ \includegraphics[width=.4\textwidth]{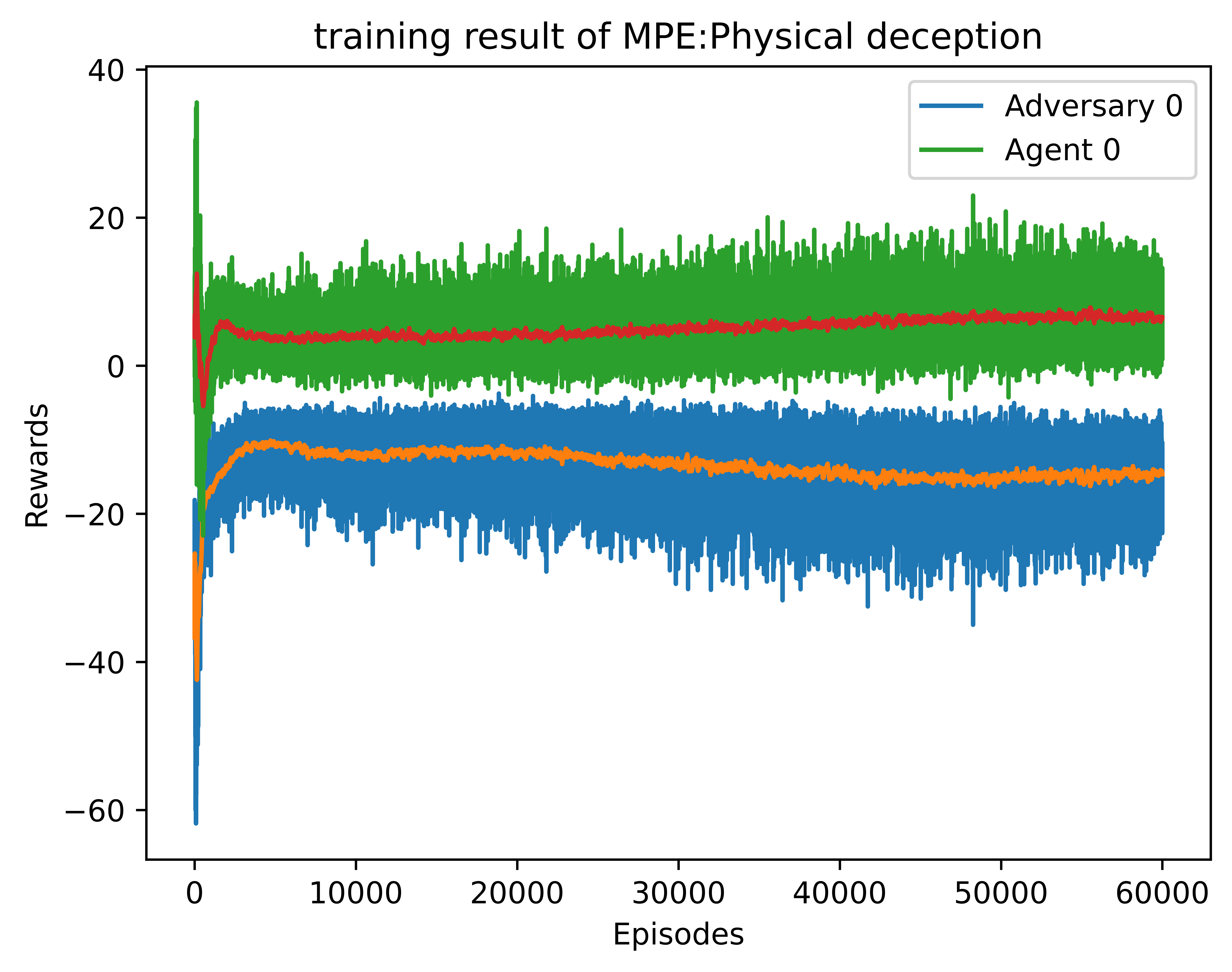}\label{fig:adam_pd}}
    \end{subfigure}
    \begin{subfigure}[LA-MADDPG]{\includegraphics[width=.4\textwidth]{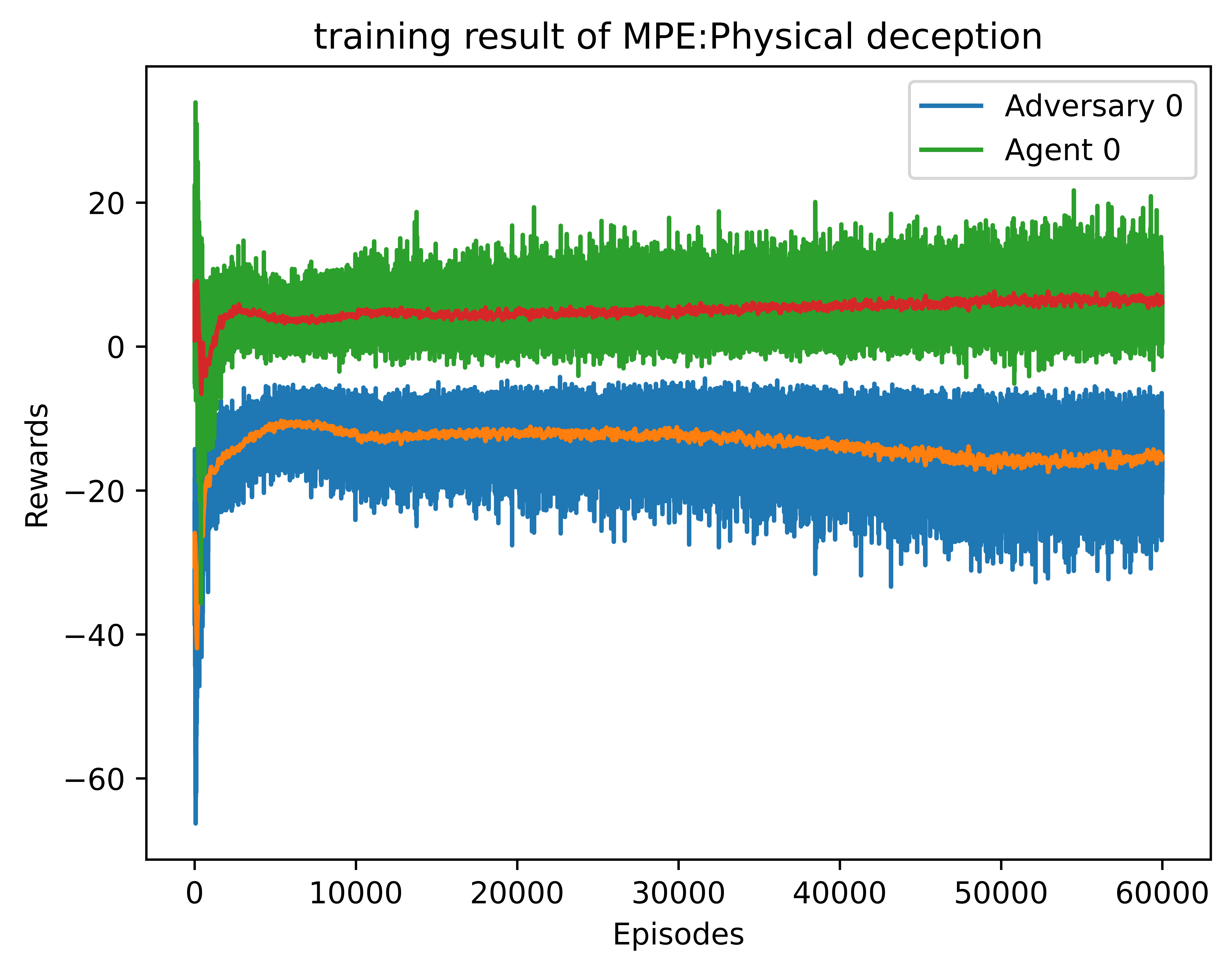}
        \label{fig:la_pd}}
    \end{subfigure}
\caption{\textbf{The figure shows the learning curves during training of GD-MADDPG and LA-MADDPG for MPE: Physical deception.} 
$x$-axis: training episodes. 
$y$-axis: agents rewards and their moving average with a window size of 100, calculated over $5$-seeds over $5$ seeds.}
\end{figure*}

Figures \ref{fig:adam_pd} and \ref{fig:la_pd} demonstrate same results but for MPE: Physical deception. In this game, We have two good agents, 'Agent $0$ and $1$' but since they are both receive same rewards, we only show agent $0$. 

\subsection{On the Rewards as Convergence Metric}\label{app:convergence}

Based on our experiments and findings from the multi-agent literature \citep{NIPS2004_88fee042}, we observe that average rewards offer a weaker measure of convergence compared to policy convergence in multi-agent games. This implies that rewards can reach a target value even when the underlying policy is suboptimal. For example, in the Rock--paper--scissors game, the Nash equilibrium policy leads to nearly equal wins for both players, resulting in a total reward of zero. However, this same reward can also be achieved if one player always wins while the other consistently loses, or if both players repeatedly select the same action, leading to a tie. As such, relying solely on rewards during training can be misleading.

\begin{figure}
  \centering
  \begin{subfigure}{\includegraphics[width=.45\linewidth]{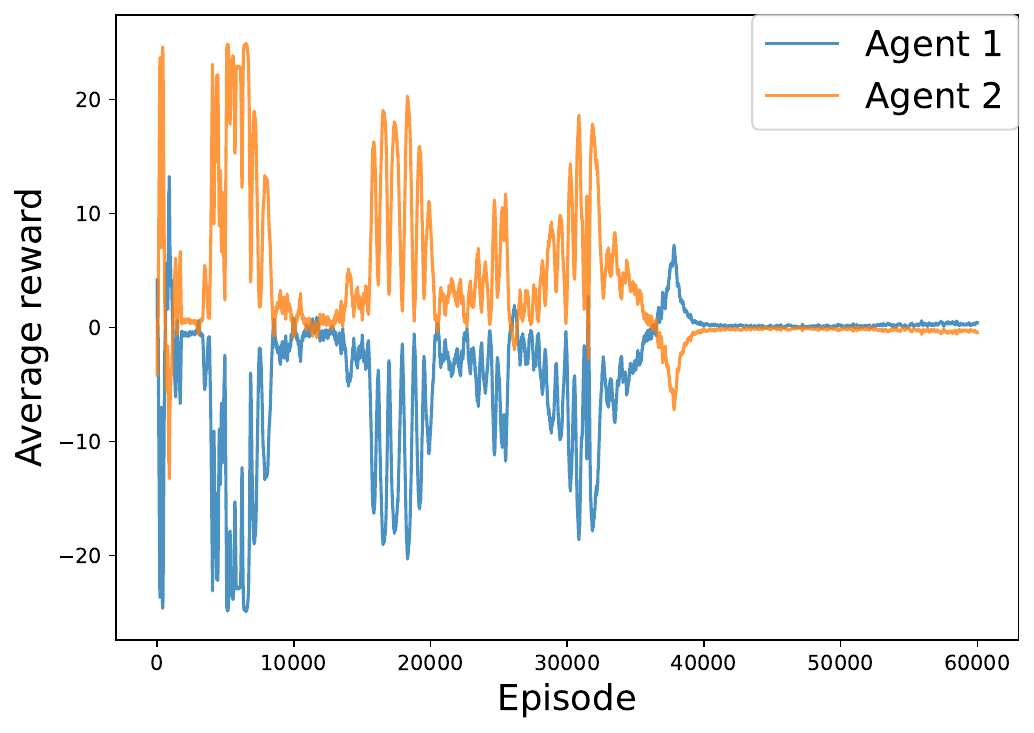}\label{fig:adam_rewards}}
  \end{subfigure}
  \begin{subfigure}{\raisebox{1em}{\includegraphics[width=.1\linewidth]{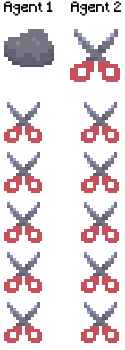}} \hspace{.7em}
  \raisebox{1em}{\includegraphics[width=.1\linewidth]{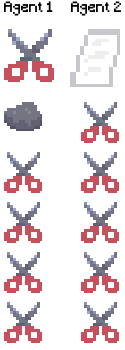}}\hspace{.7em}
  \raisebox{1em}{\includegraphics[width=.1\linewidth]{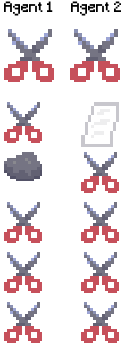}}}
  \end{subfigure}
  \begin{subfigure}
  {\includegraphics[width=.45\linewidth]{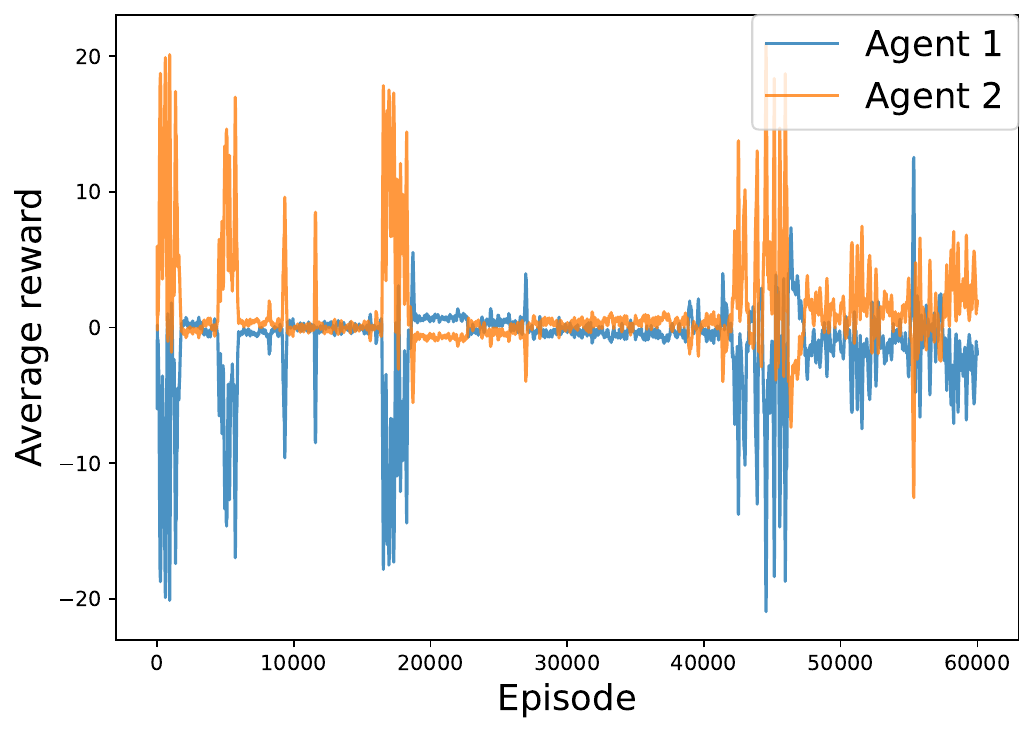}\label{fig:la_rewards}}
  \end{subfigure}
  \begin{subfigure}{\raisebox{1em}{\includegraphics[width=.1\linewidth]{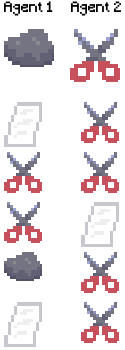}}\hspace{.7em}
  \raisebox{1em}{\includegraphics[width=.1\linewidth]{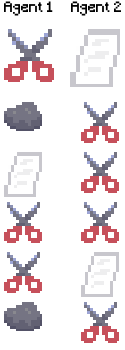}}\hspace{.7em}
  \raisebox{1em}{\includegraphics[width=.1\linewidth]{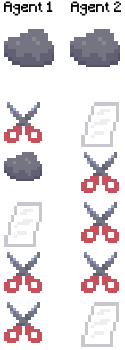}}}
  \end{subfigure}
  \vspace{-.2cm}
 \caption{\textbf{ A more detailed version of Figure~\ref{fig:snapshots_rps}.}
 Saturating rewards (left) versus actions of the learned policies at the end (right) in the Rock--paper--scissors game.
 \textbf{Top row:} \emph{GD-MADDPG}; \textbf{bottom row:} \emph{LA-MADDPG}.
 In the left column, blue and orange show the running average of rewards through a window of $100$ episodes.
 In the right column, we depict actions from the respective learned policies evaluation after training is completed, where each row represents what actions players have chosen in one step of the episode.
 Saturating rewards do not imply good performance, as evidenced by the top row; refer to Section~\ref{sec:results} for discussion.
 }\label{fig:snapshots_rps_more_details}
\end{figure}

Figure \ref{fig:snapshots_rps_more_details} (top row) depicts a case with the baseline where, despite rewards converging during training, the agents ultimately learned to play the same action repeatedly, resulting in ties. Although this matched the expected reward, it falls far short of equilibrium and leaves the agents vulnerable to exploitation by more skilled opponents. In contrast, the same figure shows results from LA-MADDPG under the same experimental conditions. Notably, while the rewards did not fully converge, the agents learned a near-optimal policy during evaluation, alternating between all three actions as expected. These results also align with the findings shown in Figure \ref{fig:adamvslavseg_rps}.

\begin{figure}
    \centering
    \includegraphics[width=\columnwidth]{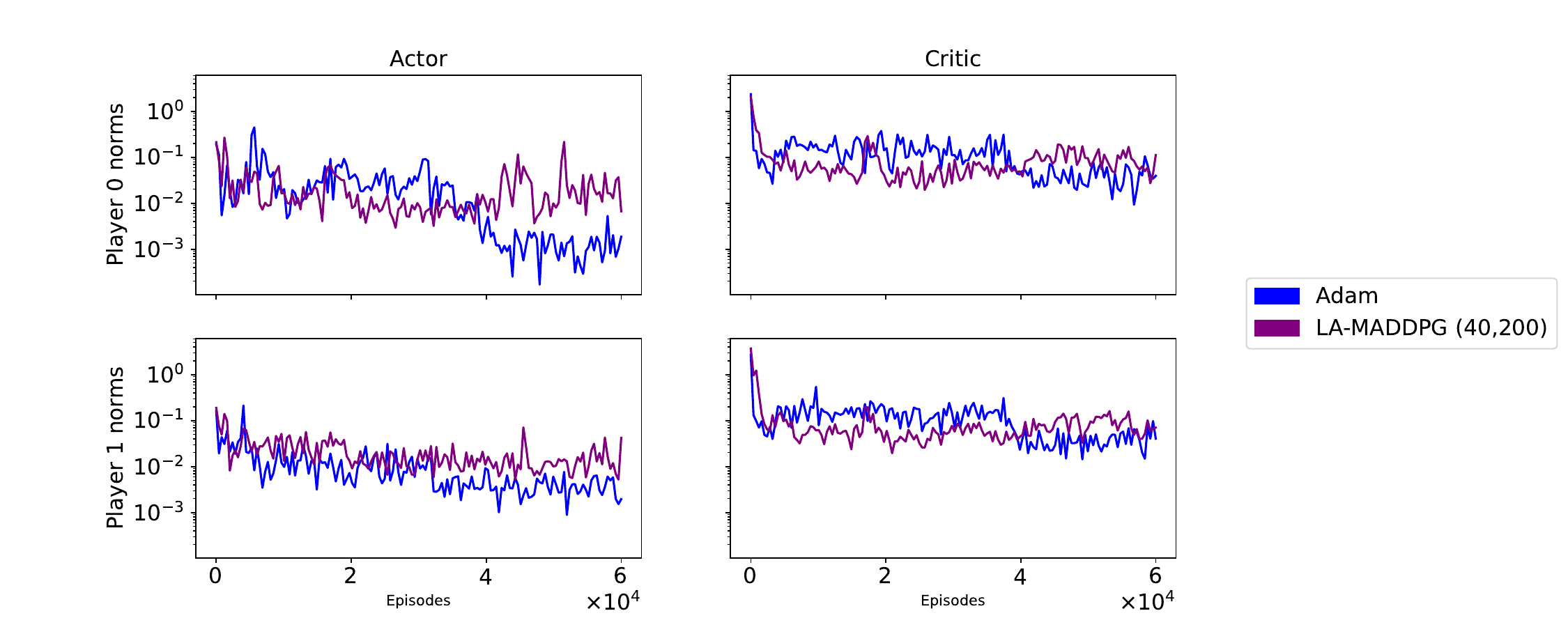}
    \caption{\textbf{Gradient norms across training in the \emph{Rock--paper--scissors} game.}
    }
    \label{fig:norms_rps}
\end{figure}

We explored the use of gradient norms as a potential metric in these scenarios but found them to be of limited utility, as they provided no clear indication of convergence for either method.  We include those results in Figure \ref{fig:norms_rps}, where we compare the gradient norms of Adam and LA across the networks of different players.

This work highlights the need for more robust evaluation metrics in multi-agent reinforcement learning, a point also emphasized in \citep{lanctot2023populationbasedevaluationrepeatedrockpaperscissors}, as reward-based metrics alone may be inadequate, particularly in situations where the true equilibrium is unknown.

\end{document}